\documentclass[]{youtu} 

\usepackage{mathpazo}
\usepackage{graphicx}
\usepackage[numbers]{natbib}
\usepackage{multirow}
\usepackage{xcolor}         
\usepackage{colortbl}
\usepackage{caption}
\usepackage{subcaption}
\usepackage{algorithmic}
\usepackage{algorithm}
\usepackage{soul}
\usepackage{xspace}
\definecolor{c1}{RGB}{249,242,234}
\definecolor{c2}{RGB}{228,246,246}
\definecolor{c3}{RGB}{223,243,230}
\definecolor{c4}{RGB}{224,222,241}

\definecolor{blue(ryb)}{rgb}{0.01, 0.28, 1.0} 
\definecolor{azure(colorwheel)}{rgb}{0.0, 0.5, 1.0}
\definecolor{turquoise}{rgb}{0.19, 0.84, 0.78}

\newcommand{\method}{\textsc{SPEAR}\xspace}

\title{Learn the Ropes, Then Trust the Wins: Self-imitation with Progressive Exploration for Agentic Reinforcement Learning}
\author{Youtu-Agent Team$^*$}

\date{September 22, 2025}
\correspondence{\{yuleiqin, arthurtan\}@tencent.com}
\sourcecode{https://github.com/TencentYoutuResearch/SPEAR}
\model{https://huggingface.co/collections/yolay/spear-68da1c8b75098b1868db59c8}

\begin{document}

\abstract{
Reinforcement learning (RL) is the dominant paradigm for sharpening strategic tool use capabilities of LLMs on long-horizon, sparsely-rewarded agent tasks, yet it faces a fundamental challenge of exploration-exploitation trade-off.
Existing studies stimulate exploration through the lens of policy entropy, but such mechanical entropy maximization is prone to RL instability due to the multi-turn distribution shifting.
In this paper, we target the progressive exploration-exploitation balance under the guidance of the agent's own experiences without succumbing to either entropy collapsing or runaway divergence.
We propose \method\raisebox{-0.1em}{\includegraphics[scale=0.12]{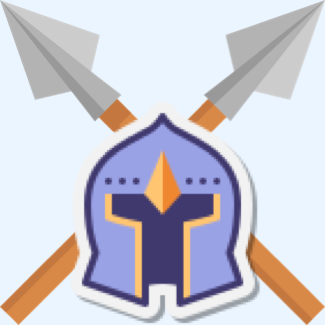}}, a self-imitation learning (SIL) recipe for training agentic LLMs.
It extends the vanilla SIL, where a replay buffer stores
good experience for off-policy update,
by gradually steering the policy 
entropy across stages.
Specifically,
the proposed curriculum scheduling harmonizes intrinsic reward shaping and self-imitation to 1) expedite exploration via frequent tool interactions at the beginning,
and 2) strengthen exploitation of successful tactics upon convergence towards familiarity with the environment.
We also combine bag-of-tricks of industrial RL optimizations
for a strong baseline Dr.BoT to demonstrate our effectiveness.
In ALFWorld and WebShop,
\method increases the success rates of GRPO/GiGPO/Dr.BoT by up to 16.1\%/5.1\%/8.6\% and 20.7\%/11.8\%/13.9\%, respectively.
In AIME24 and AIME25,
\method boosts Dr.BoT by up to 3.8\% and 6.1\%, respectively.
Such gains incur only 10\%–25\% extra theoretical complexity and negligible runtime overhead in practice, demonstrating the plug-and-play scalability of \method.
}

\maketitle

\renewcommand{\thefootnote}{*}
\footnotetext{Full author list in contributions.}
\renewcommand{\thefootnote}{\arabic{footnote}}

\section{Introduction}

\begin{figure}[htbp]
\begin{center}
\includegraphics[width=0.99\textwidth]{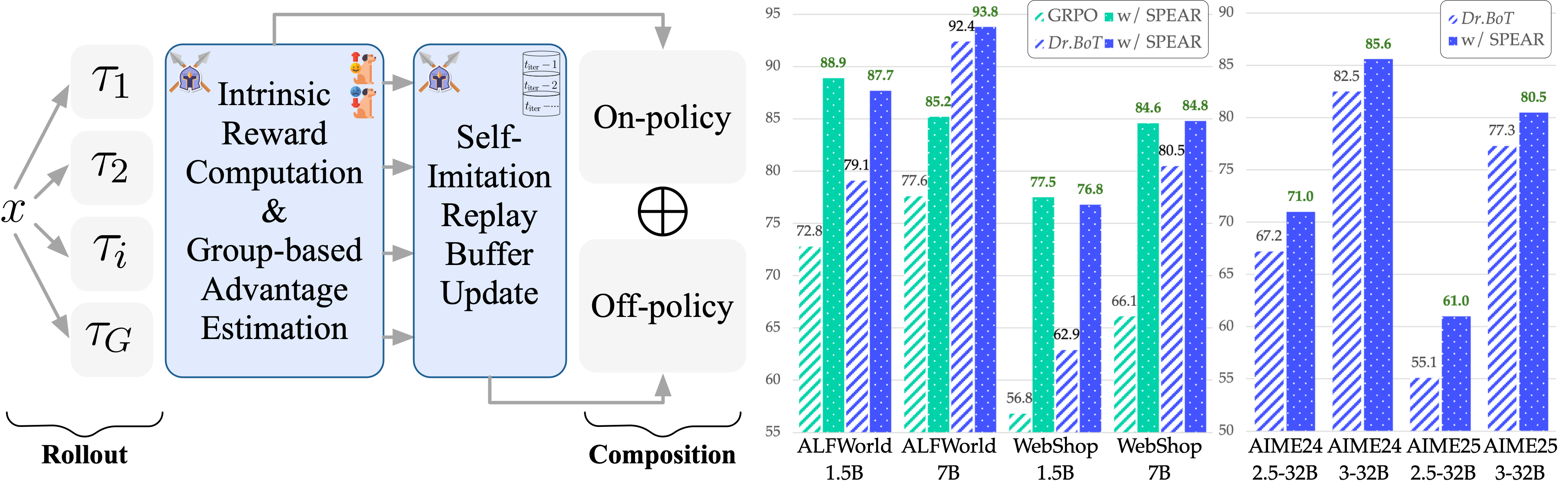}
\end{center}
\caption{{
Our \method harmonizes the curriculum-scheduled self-imitation learning with intrinsic reward shaping for progressive exploration,
improving policy performance across agentic tasks.}
}
\label{fig:concept}
\end{figure}

Reinforcement Learning (RL)~\cite{lambert2024tulu,guo2025deepseek,qin2025incentivizing} has driven the development of reasoning capabilities of Large Language Models (LLMs).
Built upon the reason-and-act (ReAct) paradigm~\cite{yao2023react}, LLMs have powered various agentic applications such as simulated robot navigation~\cite{shridhar2020alfworld,li2024embodied}, mobile assistant~\cite{wang2024mobile,li2025mobileuse}, web navigator~\cite{furuta2023multimodal,he2024webvoyager}, deep searcher~\cite{jin2025search,li2025webthinker,tao2025webshaper}, and GUI master~\cite{qin2025ui,hong2024cogagent}.
{
A fundamental challenge in applying RL to LLM agents
is to manage the balance between exploration and exploitation.
The LLM agent needs to \textit{exploit} both pretrained knowledge and past interactions to formalize experience that 
maximize rewards. 
At the same time, it must \textit{explore} novel behaviors through tool-integrated reasoning and reflection.}
The interweaving between exploration and exploitation determines the emerging agent's competence upon convergence.

Existing studies often quantify the exploration potential through entropy~\cite{sutton1988learning,williams1991function,cui2025entropy,xue2025simpletir}, where the decline of policy entropy indicates 
{over-confidence
with insufficient exploration.}
In this case, a series of regularization techniques~\cite{ziebart2008maximum,schulman2017proximal,haarnoja2018soft} have been proposed to maximize entropy
~\cite{haarnoja2017reinforcement,zhao2019maximum,xin2020exploration,zhang2021exploration,seo2021state,mehr2023maximum,kim2023adaptive,hao2023exploration}.
However, when it comes to LLM-driven agents, entropy-based control is fragile: the accumulation of low-probability tokens from the environment feedback induces severe distribution shifting, often leading to mode collapse~\cite {xue2025simpletir,dong2025agentic}.
{Agent} models may experience sustained entropy growth due to 
uncertainty about multi-turn interactions
and training instability becomes frequent~\cite{mai2025agent,yao2025offpolicy,wang2025ragen}.
Recent approaches attempt to mitigate this issue by 
cold-start supervised fine-tuning (SFT)~\cite{tao2025webshaper,qin2025ui,feng2025retool,qin2025unleashing} or 
hybrid schemes that combine RL with SFT~\cite{zhang2025policy}.
Although these methods improve stability, 
they compromise {policy's discovery of} strategies beyond those present in the SFT corpus.
This limitation highlights the need for adaptive training frameworks that can \textit{dynamically schedule LLM-driven agents to decide when to explore and when to exploit}.

In this paper, we are trying to answer the following core research question:
\textbf{Can we schedule a smooth transition between \textit{exploration} and \textit{exploitation} guided by the policy's own experience without going to extremes of either entropy collapsing or runaway divergence?}
{
We hypothesize that 
the agent should maintain} its policy entropy within a dynamic but controlled range that evolves over time:
1) \textbf{At the early stages, \textit{increasing entropy} is beneficial for broad \textit{skill-level} exploration}. The agent is expected to rapidly develop tool-use capabilities, encounter unfamiliar observations, and engage in trial-and-errors.
2) \textbf{As training advances, however, a shift toward \textit{converging entropy} is required}. This enables the agent to consolidate problem-solving heuristics and emphasize \textit{action-level} exploration.
The agent exploits reward signals to choose comparatively more effective actions and adapts to changing distributions for stabilizing its evolutionary path.

\begin{figure}[htbp]
\begin{center}
\includegraphics[width=0.99\textwidth]{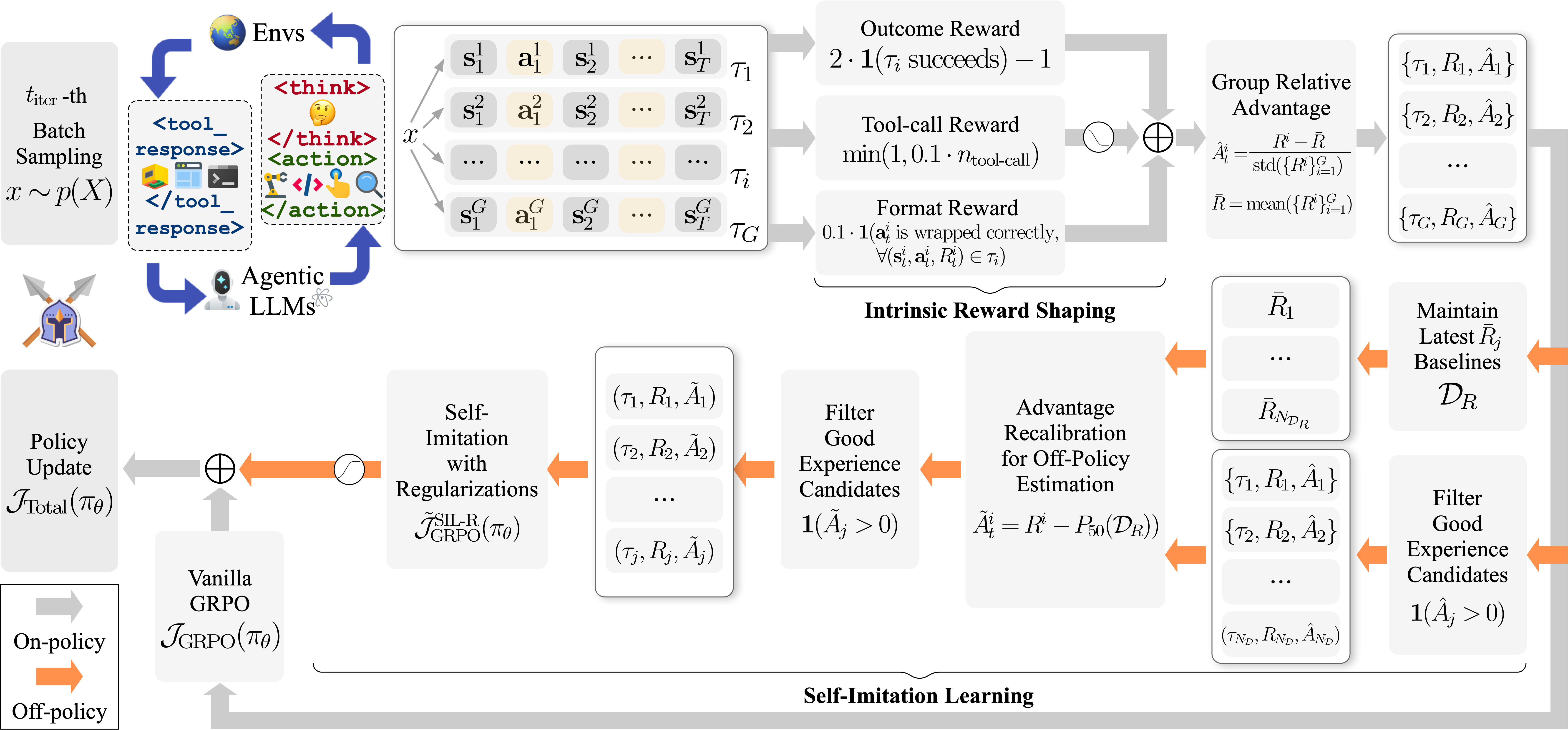}
\end{center}
\caption{
{
Overview of \method.
First, the agent interacts with the environment for a set of trajectories,
which flow through intrinsic reward shaping and advantage estimation with on-policy updates.
Second, they are selected and stored in a replay buffer, enabling off-policy updates via the proposed self-imitation scheme.
This dual integration allows the maximal utility of past experiences,
thereby expanding the effective exploration space, while simultaneously mitigating persistent uncertainty.
}
}
\label{fig:overview}
\end{figure}

To address this,
we propose the 
\textbf{S}elf-imitation with \textbf{P}rogressive \textbf{E}xploration for \textbf{A}gentic \textbf{R}einforcement learning (\textbf{\method}\raisebox{-0.1em}{\includegraphics[scale=0.12]{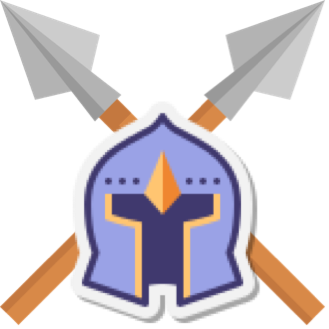}}),
a curriculum-based RL recipe for improving the \textit{exploration-exploitation} balance with \textit{self-imitation} and \textit{intrinsic reward}.
As shown in Figure~\ref{fig:concept}, the core principle follows the vanilla Self-Imitation Learning (SIL)~\cite{oh2018self,ferret2020self} where an independent replay buffer is prepared to store the state-action pairs only when their returns in the past episodes exceed the baselines.
Such a replay buffer is exploited to encourage actions with good returns and improve hard exploration based on these successful trajectories under the sparse-reward, long-horizon agent tasks. Specifically, we introduce three modifications to SIL tailored to the dynamics of policy entropy in agentic tasks.
First, we incorporate a curriculum to integrate both \textit{skill-level} and \textit{action-level} exploration by adjusting reward shaping and self-imitation across stages.
Second, we tackle the off-policy nature of the update with experiences in the buffer and avoid advantage recomputation by advantage recalibration.
Third, we regularize policy updates to stabilize entropy and mitigate reward hacking.
Finally, inspired by existing industrial bag-of-tricks, we present a strong baseline $\textit{Dr.BoT}$ { for agentic RL training.}
Our \method brings considerable performance gains to
{\color{turquoise}{GRPO}}/{\color{azure(colorwheel)}{GiGPO}}~\cite{feng2025group}/{\color{blue(ryb)}{$\textit{Dr.BoT}$}} respectively up to {\color{turquoise}{16.1\%}}/{\color{azure(colorwheel)}{5.1\%}}/{\color{blue(ryb)}{8.6\%}} on ALFWorld~\cite{shridhar2020alfworld} and {\color{turquoise}{20.7\%}}/{\color{azure(colorwheel)}{11.8\%}}/{\color{blue(ryb)}{13.9\%}} on WebShop~\cite{yao2022webshop}.
It boosts our {\color{blue(ryb)}{$\textit{Dr.BoT}$}} respectively up to {\color{blue(ryb)}{3.8\%}} on AIME24 and {\color{blue(ryb)}{6.1\%}} on AIME25~\cite{aime}.
These gains come with around $10\%\sim25\%$ computation overhead in theoretical complexity, but end up with quite comparable runtime per iteration in practice.
Such compatibility and scalability enable \method a plug-and-play algorithm for training versatile agents.
In summary, our contributions are:

{1)} We propose \method, a generalization of the SIL for training LLM agents.
It bypasses the costly expert imitation and allows exploration under the guidance of one's own rewarded experience.

{2)} We bring in curriculum scheduling to harmonize SIL with intrinsic reward shaping for policy entropy management and progressive transition from \textit{skill-based} to \textit{action-based} exploration.

{3)} We propose a strong baseline, $\textit{Dr.BoT}$, which combines established RL techniques validated in industrial practice, confirming its effectiveness and superiority over existing baselines.


\section{Related Work}
\label{sec:related}

\subsection{Reinforcement Learning Algorithms for LLMs}

With the advent of large-scale reasoning models~\cite{jaech2024openai}, Reinforcement Learning (RL)~\cite{ouyang2022training} has been adopted more broadly.
Proximal Policy Optimization (PPO)~\cite{schulman2017proximal}
leverages an actor–critic architecture together with the clipped surrogate objective and a Kullback–Leibler (KL) divergence penalty to constrain policy update.
Group Relative Policy Optimization (GRPO)~\cite{guo2025deepseek,shao2024deepseekmath} simplifies this setup by replacing the critic with a group-wise baseline.
Building on GRPO, 
DAPO~\cite{yu2025dapo} uses dynamic sampling and "clip higher" to encourage exploration and stabilize training.
Dr.GRPO~\cite{liu2025understanding} addresses length bias and the difficulty bias.
Existing methods have greatly advanced RL for LLMs.
However, naively combining them can lead to conflicts or tight couplings among techniques.
To this end, we harmonize the strengths of DAPO, Dr.GRPO, and other agent studies from research and industrial practice to establish a strong baseline, \textit{Dr.BoT}, as detailed in Section~\ref{sec:strong-baseline}.

\subsection{Optimization of LLM Agents}

Recent researches investigate how to endow models with better tool-use capabilities~\cite{feng2025retool, li2025torl, xue2025simpletir}.
LLMs are optimized to strengthen information seeking from open web~\cite{jin2025search,tao2025webshaper,gao2025beyond}.
RAGEN~\cite{wang2025ragen} improves the stability of multi-turn RL through instance filtering and gradient shaping.
GiGPO~\cite{feng2025group} augments group-level advantages with additional step-level advantage estimates.
ARPO~\cite{dong2025agentic} monitors entropy dynamics during rollouts to branch trajectories adaptively.
In this work, we address the exploration–exploitation dilemma under multi-turn tool-use settings.
We introduce a curriculum–regulated RL regime that gradually shifts skill-based exploration towards action-based exploration.
We integrate self-imitation and intrinsic reward to consolidate successful behaviors (Section~\ref{sec:self-imitation-learning}).
Our \method can work with existing algorithms in a plug-and-play manner,
exhibiting a high level of compatibility and generalization.

\subsection{Exploration in Reinforcement Learning}

Curiosity-driven methods~\cite{pathak2017curiosity,houthooft2016vime} grant intrinsic rewards for prediction error or novelty to actively seek unfamiliar states.
Count-based algorithms~\cite{bellemare2016unifying,tang2017exploration} introduce pseudo-counts derived from a density model to assign count-based bonuses.
Skill acquisition methods~\cite{gregor2016variational,eysenbach2018diversity} discover distinct options by maximizing the mutual information.
Entropy-regularization methods~\cite{haarnoja2018soft,cui2025entropy} maximize the expected reward and entropy of the policy.
However,
traditional exploration techniques can lead to divergence of agent LLMs as the multi-turn interactions already result in the increased uncertainty on unfamiliar observations.
Under such circumstance,
we propose the curriculum-guided self-imitation to leverage the agent's own experience for balancing exploration and exploitation.
It avoids handcrafted heuristic techniques in previous studies and instead fully relies on the agent itself to reinforce successful and valid patterns.

\subsection{Experience Replay in Reinforcement Learning}

Self-Imitation Learning (SIL)~\cite{oh2018self} takes advantage of past successful experience to drive its future learning~\cite{schaul2015prioritized,horgan2018distributed,gangwani2018learning,pan2022understanding,saglam2023actor}.
SAIL~\cite{ferret2020self} extends SIL to off-policy, action value-based RL methods.
\cite{tang2020self} proves that SIL's return-based update provides a bias–variance trade-off that speeds up learning.
SILfD~\cite{pshikhachev2022self} extends SIL to leverage both external demonstrations and the agent's own experience.
GSIL~\cite{xiao2024leverage} proposes an offline alignment framework that uses self-imitation on demonstration data.
While SIL benefits long-horizon problems,
its induces entropy collapsing to agent RL.
To mitigate this, we harmonize both self-imitation and intrinsic reward with curriculum scheduling for progressive exploration.

\section{Preliminaries}

\subsection{Problem Definition}
\label{sec:problemdefinition}

Given a task $x\sim p(X)$ where $p(X)$ represents data distribution,
an LLM agent parameterized by $\theta$ interacts with the environment $E$ until it completes the task or exceeds the max number of turns $T$.
It can be modeled by Markov Decision Process (MDP) where $\mathbf{s}_t$, $\mathbf{a}_t$, and $R_t$ respectively denote the state, action, and reward at time $t$.
Given a full episode $\tau=\{(\mathbf{s}_1, \mathbf{a}_1, R_1), (\mathbf{s}_2, \mathbf{a}_2, R_2),...\}$,
we aim to optimize the agent policy $\pi_{\theta}$.
\label{sec:actionspace}
Following previous studies~\cite{dong2025tool,feng2025group,feng2025retool,dong2025agentic}, 
we define three distinct types of actions
(see Appendix~\ref{sec:actionspacedetails}).

\subsection{Policy Optimization}
\label{sec:policyoptimization}

We adopt the GRPO~\cite{shao2024deepseekmath} which stems from PPO~\cite{schulman2017equivalence,schulman2017proximal} but 
replaces the model-based advantages ${A}$~\cite{schulman2015high}
with the group-based $\hat{A}$ (Appendix~\ref{sec:ppogrpo}).

\begin{figure}[htbp]
\begin{center}
\begin{subfigure}{.49\textwidth}
  \centering
  \includegraphics[width=\textwidth]{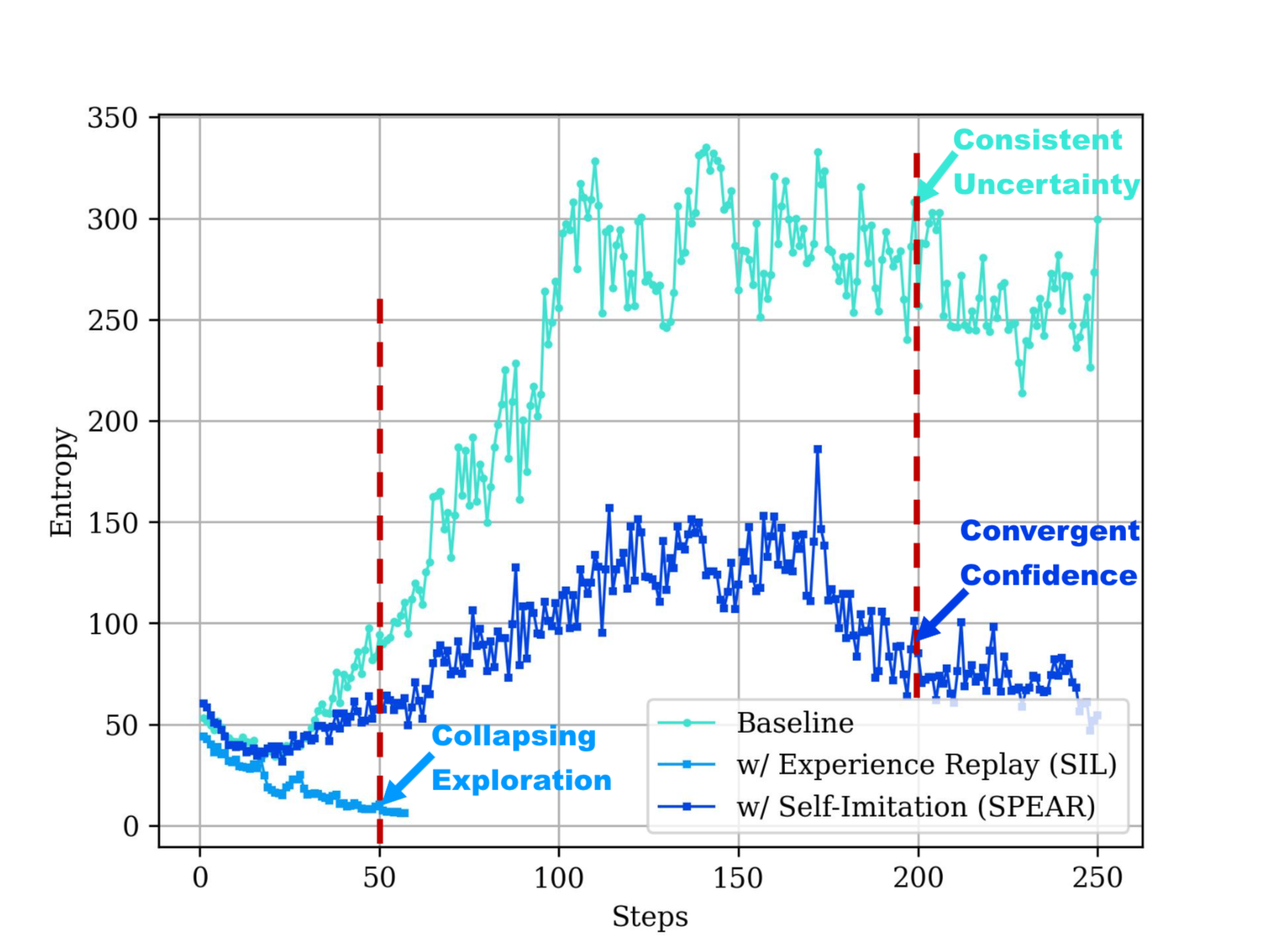}
  \caption{Entropy (\texttt{seq-mean-token-sum-norm}).}
\end{subfigure}%
\begin{subfigure}{.49\textwidth}
  \centering
  \includegraphics[width=\textwidth]{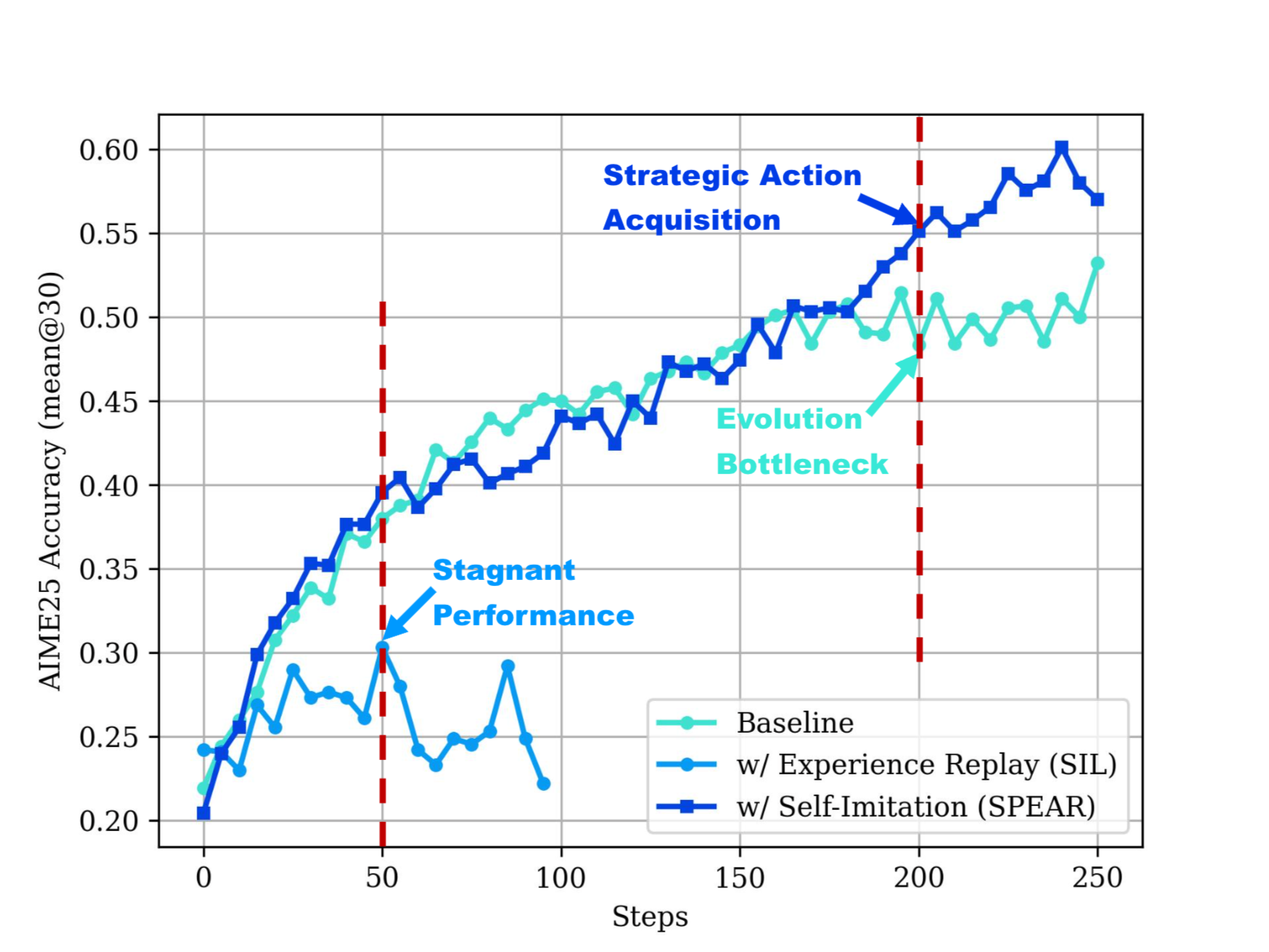}
  \caption{Accuracy on AIME 2025.}
\end{subfigure}%
\end{center}
\caption{Effect of our self-imitation on action-level strategy exploration (Qwen2.5-32B with code interpreter).
The vanilla experience replay technique~\cite{oh2018self} that enforces early overfitting of the few available trajectories in the buffer causes entropy collapsing and exploration shrinkage.
At the beginning,
the LLM agent struggles at tool-calling skills and fails to cultivate the transition of distribution towards frequent tool utilization and tool-integrated reasoning.
The naive replay limits the transformation of reasoning paradigm.
In contrast,
our \method introduces both curriculum- and covariance- based regularization into self-imitation.
Its curriculum schedule with an increasing emphasis on the replay data allows easy acquisition of tool-use skills at first,
and stimulates strategic action plans later.
The covariance clipping removes over-confident tokens, whose log probabilities are highly associated with their advantage gains,
out of optimization.
Our self-imitation gives promises to exploring novel strategies and achieves steady growth on AIME 2025.
}
\label{fig:entropy}
\end{figure}

\section{Training Agentic LLMs with \method\raisebox{-0.1em}{\includegraphics[scale=0.15]{figures/spear-logo.png}}}
\label{sec:method}

\subsection{Preliminary Findings}

\begin{figure}[htbp]
\begin{center}
\begin{subfigure}{.49\textwidth}
  \centering
  \includegraphics[width=\textwidth]{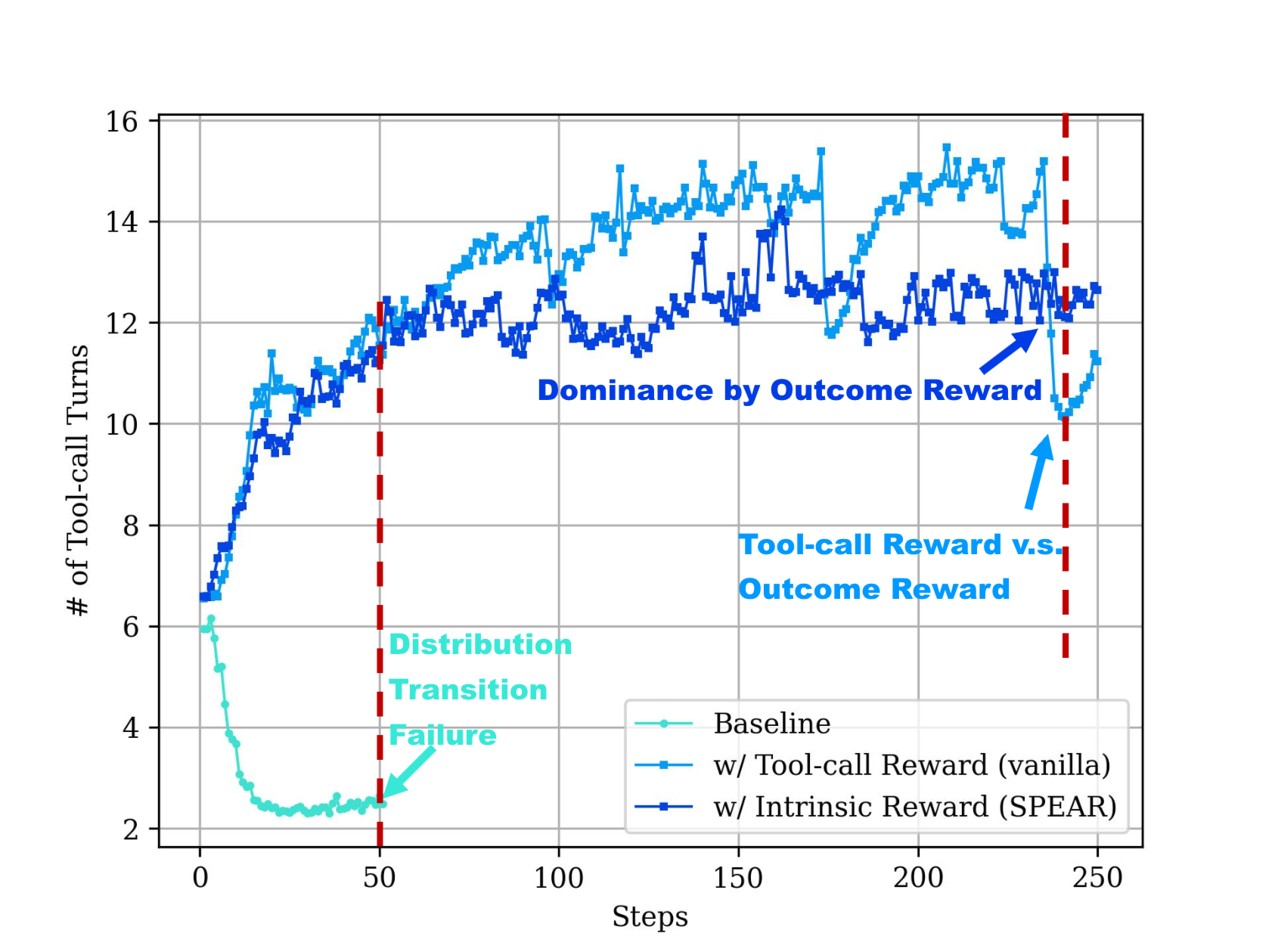}
  \caption{Number of tool-call turns.}
\end{subfigure}%
\begin{subfigure}{.49\textwidth}
  \centering
  \includegraphics[width=\textwidth]{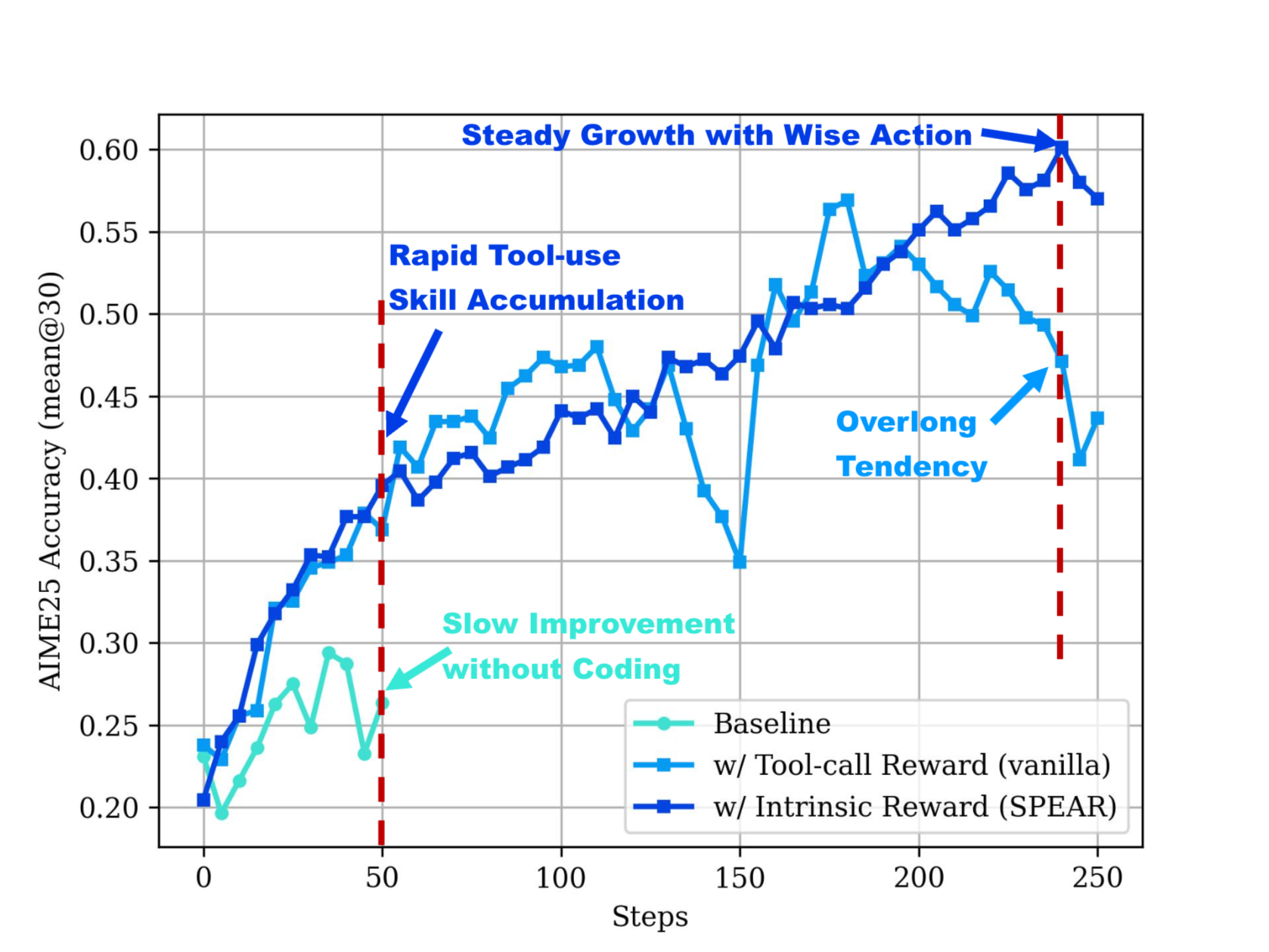}
  \caption{Accuracy on AIME 2025.}
\end{subfigure}%
\end{center}
\caption{Effect of our intrinsic reward on skill-level strategy exploration (Qwen2.5-32B with code interpreter).
The baseline does not consider tool-calling as a rewarded behavior and its number of interaction with the environment drops quickly due to the negative feedback of bad codes.
In this case,
the LLM gives up coding and degrades to text-based reasoning.
The vanilla tool-call reward,
despite being effective in learning tool-call skills at first,
causes competition with the outcome reward later.
Due to the limited context length,
the excessive tool-call turns prevents submission of the final answer and thereafter the accuracy declines immediately.
We propose the curriculum schedule as an intrinsic reward design where its strength decays over step to allow the agent to merely focus on the accuracy with wiser actions.
It prevents reward hacking for unnecessarily long interactions.
}
\label{fig:toolcall}
\end{figure}

The extension of SIL to LLM-driven agents faces entropy collapse.
Figure~\ref{fig:entropy} illustrates that the overfitting of the few available successful experience causes irreversible stagnation of exploration.
In addition,
{we} demonstrate that the inclusion of the tool-call reward is a double-edged sword (Figure~\ref{fig:toolcall}),
where the competition between reward terms causes the oscillations to converge.
To address these challenges, we introduce \method for progressive exploration with self-imitation (Algorithm~\ref{alg:sal}).

\subsection{Self-Imitation Learning}
\label{sec:self-imitation-learning}

{
We resort to self-imitation to unearth past successful experience for effective \textit{action-level} exploration,
where the agent learns novel strategies along the promising decision path instead of random walk and bifurcation.
We prevent policy entropy divergence by replaying rewarded trajectories.
}

\paragraph{Prioritized Experience Replay in Self-Imitation.}
A replay buffer is maintained to store previous trajectories, their rewards and advantages $\mathcal{D}=\{(\tau_{j}, R_{j}, \hat{A}_{j})\}, j=1,2,...,N_{\mathcal{D}}$ where $N_{\mathcal{D}}$ denotes the buffer size.
To exploit only good trajectories,
we keep those with positive advantages:
\begin{equation}
\begin{aligned}
\mathcal{J}_{\text{GRPO}}^{\text{SIL}}(\pi_\theta)&=
     \mathbb{E}_{\{\tau_{j}\}_{j=1}^{N_{\mathcal{D}}}\sim \{\pi_{\theta_{\text{old}}}(\cdot|x),\ x\sim p(X)\}}\sum_{j=1}^{N_{\mathcal{D}}}\mathcal{J}_{\text{GRPO}}^{j}\cdot\mathbf{1}(\hat{A}_{j}>0),\\
\end{aligned}
\end{equation}
where the indicator $\mathbf{1}(\cdot)$ equals to 1 when the condition satisfied and 0 otherwise.
The past trajectories not only come from the last policy $\pi_{\theta_{old}}$ but also the policies $\{\pi_{\theta_{\text{old}}}\}$ of few steps earlier.


\paragraph{Advantage Recalibration for Off-Policy Estimation.}
{We propose to recalibrate the advantage of trajectories in the buffer to address the underlying off-policy challenge.}
That is to say, the observed return of a trajectory from the past policy becomes increasingly different from the current one, under the assumption that the policy keeps improving during iterations~\cite{ferret2020self,luo2021self}. 
Under this assumption, vanilla SIL computes the advantage with a pointwise max with the per-state empirical return as a baseline, which can be seen as a proxy for the upper-envelope projection of the value function onto empirical returns.
GRPO removes the learned value baseline by estimating the state-dependent baseline performance through its reliance on intra-group reward averaging, but this still depends on the target policy and requires extra computation resources for sampling.
{Dynamic adjustment on the baseline performance is performed to calibrate relative gains without introducing additional computing.}
Specifically,
we maintain a First-In-First-Out (FIFO) buffer of intra-group baselines for the latest $N_{\mathcal{D}_{R}}$ trajectories $\mathcal{D}_R=\{\bar{R}_{j}\}_{j=1}^{\mathcal{D}_{R}}$ where $N_{\mathcal{D}_{R}}$ denotes the size of the baseline buffer.
As training progresses, due to the high variance nature of agentic RL, we utilize the 50-th percentile $P_{50}(\mathcal{D}_R)$ as a conservative but robust estimation of the policy baseline with either upward or downward trends.
To bypass the inaccurate estimation of intra-group standard deviation, we follow~\cite{liu2025understanding} to simply remove such a term in advantage computation:
\begin{equation}\label{eq:recalibrateadv}
    \tilde{A}^{i}_{t}=R^{i}-P_{50}(\mathcal{D}_R).
\end{equation}
Such recalibrated advantage enjoys three benefits:
1) the baseline performance correlates with the policy change;
2) the outdated experiences 
can be filtered out with both $\hat{A}_{j}>0$ and $\tilde{A}_{j}>0$;
3) the difficulty bias by group normalization
can be mitigated.
The updated off-policy 
SIL objective is:
\begin{equation}\label{eq:silvanilla}
    \tilde{\mathcal{J}}_{\text{GRPO}}^{\text{SIL}}(\pi_\theta)=
     \mathbb{E}_{\{\tau_{j}\}_{j=1}^{N_{\mathcal{D}}}\sim \{\pi_{\theta_{\text{old}}}(\cdot|x),\ x\sim p(X)\}}\sum_{j=1}^{N_{\mathcal{D}}}\tilde{\mathcal{J}}_{\text{GRPO}}^{j}\cdot\mathbf{1}(\hat{A}_{j}>0\ \&\ \tilde{A}_{j}>0),
\end{equation}
\begin{equation}\label{eq:silvanilladetail}
    \tilde{\mathcal{J}}_{\text{GRPO}}^{i}=\Biggr[
    \frac{1}{T}\sum_{t=1}^{T}(\min(r^{i}_{t}(\theta)\tilde{A}^{i}_{t}, \text{clip}(r^{i}_{t}(\theta), 1-\epsilon, 1+\epsilon)\tilde{A}^{i}_{t})-
    \beta D_{\text{KL}}^{i}(\pi_{\theta}||\pi_{\text{ref}})\Biggr].
\end{equation}

\paragraph{Progressive Experience Utilization with Curriculum Schedule.}
\label{sec:curriculumsil}
{We perform scheduling to
1) restrict mechanical imitation of probable-yet-immature experience at an early stage,
and 2) prevent consistent uncertainty about the environment states and policy actions at later stage.}
We apply a warm-up $\gamma$ on the SIL term under the assumption that initially the transition of distribution towards diverse actions outweighs the imitation of limited solution patterns (see Equation~\ref{eq:gamma} and Figure~\ref{fig:gammacurve}).

\begin{equation}\label{eq:grpo_sal}
\mathcal{J}_{\text{Total}}(\pi_\theta)=\mathcal{J}_{\text{GRPO}}(\pi_\theta) + \gamma\cdot\tilde{\mathcal{J}}_{\text{GRPO}}^{\text{SIL}}(\pi_\theta).
\end{equation}

\subsection{Intrinsic Reward Shaping}

{
We resort to intrinsic reward for \textit{skill-level} exploration where the agent is guided by a tool-call reward to broadly investigate tool usage.
Such design not only benefits tool learning but more importantly stimulates interactions that familiarize the agent with the environment for experience accumulation.
}

\paragraph{Reward Composition.}
\label{sec:rewardcomposition}
A compound reward $R^{i}$ of each trajectory $\tau_i$ not only considers the 
final outcome but also the 
behaviors that are promising to achieve the goal:
an 
outcome accuracy reward $R^i_{\text{outcome}}$, a 
continuous tool-call reward $R^i_{\text{tool-call}}$, and a
format reward $R^i_{\text{format}}$ (see Appendix~\ref{sec:rewarddef}).

\paragraph{Progressive Reward Modulation with Curriculum Schedule.}
\label{sec:curriculumoutcomereward}

{
We regulate the contribution of tool call reward to:
1) accelerate the mastering of tool usage for quick distribution shifting towards new task settings at an early stage, and 2) prevent optimization oscillation and competition at a later stage.}
Although previous studies~\cite{qian2025toolrl,li2023tool,da2025agent,xia2025agentrm,singh2025agentic,wei2025autotir,gou2023tora,lin2025understanding} experimented with various auxiliary rewards,
we show that the addition of tool-call reward is a double-edged sword.
The agent trained without the tool-call reward fails to develop tool-integrated reasoning (Figure~\ref{fig:toolcall}) due to negative tool response:
1) missing \texttt{import} of modules;
2) reference to \texttt{undefined} variables;
3) unexpected \texttt{indentation} error;
and 4) forgetting to \texttt{print} results.
The agent quickly gives up coding to run away from errors and turns to pure textual reasoning.
On the other hand,
the enforcement of tool-call reward stimulates an increasing number of interaction turns,
leading to over-long responses that cause oscillation to outcome accuracy.
We alleviate the competition between reward terms by scheduling the tool-call reward with $\mu$ (Equation~\ref{eq:mudecay} and Figure~\ref{fig:reward}):
\begin{equation}\label{eq:rewardtotal}
    R^{i}=R^i_{\text{outcome}}+\mu\cdot R^i_{\text{tool-call}}+R^i_{\text{format}}.
\end{equation}

\subsection{\textit{Dr.BoT} as A Strong Baseline}
\label{sec:strong-baseline}

To provide a strong baseline,
we refer to the existing studies~\cite{liu2025part,sun2025stabilizing,bai2025towards,cui2025entropy} for diverse exploration,
stable convergence,
and effective training.
Our baseline,
\textit{Dr.BoT},
consists of bag-of-tricks modifications to the GRPO (see Appendix~\ref{sec:botdetails}).

\begin{table}[htbp]
\centering
\caption{Results on ALFWorld \& WebShop (\%).
PT \& FW
stand for prompting \& framework.
}
\label{tab:ALFWorldwebshop}
\resizebox{0.67\linewidth}{!}{
\setlength{\tabcolsep}{1mm}{
\fontsize{9pt}{10pt}\selectfont{
\begin{tabular}{lllllllllll}
\toprule
\multirow{2}{*}{\textbf{Type}} & \multirow{2}{*}{\textbf{Method}} & \multicolumn{7}{c}{\textbf{ALFWorld}} & \multicolumn{2}{c}{\textbf{WebShop}} \\
 &  & \multicolumn{1}{l}{Pick} & \multicolumn{1}{l}{Look} & \multicolumn{1}{l}{Clean} & \multicolumn{1}{l}{Heat} & \multicolumn{1}{l}{Cool} & \multicolumn{1}{l}{Pick2} & \multicolumn{1}{l}{All} & \multicolumn{1}{l}{Score} & \multicolumn{1}{l}{SR} \\
 \midrule
\multicolumn{11}{c}{\textit{Qwen2.5-1.5B-Instruct}} \\
PT & I/O & 5.9 & 5.5 & 3.3 & 9.7 & 4.2 & 0.0 & 4.1 & 23.1 & 5.2 \\
FW & ReAct & 17.4 & 20.5 & 15.7 & 6.2 & 7.7 & 2.0 & 12.8 & 40.1 & 11.3 \\
FW & Reflexion & 35.3 & 22.2 & 21.7 & 13.6 & 19.4 & 3.7 & 21.8 & 55.8 & 21.9 \\
RL & PPO & 64.8 & 40.5 & 57.1 & 60.6 & 46.4 & 47.4 & 54.4 & 73.8 & 51.5 \\
RL & RLOO & 88.3 & 52.8 & 71.0 & 62.8 & 66.4 & 56.9 & 69.7 & 73.9 & 52.1 \\
\rowcolor[gray]{.95}
RL & GRPO & 85.3 & 53.7 & 84.5 & 78.2 & 59.7 & 53.5 & 72.8 & 75.8 & 56.8 \\
\rowcolor{c2}
RL & + \method (ours) & 93.9 & 80.9 & 96.4 & 87.4 & 88.3 & 79.1 & 88.9{\tiny \textbf{\textcolor{teal}{(+16.1\%)}}} & 90.0 & 77.5{\tiny \textbf{\textcolor{teal}{(+20.7\%)}}} \\
\rowcolor[gray]{.95}
RL & \textit{Dr.BoT} (GRPO) & 92.2 & 75.8 & 81.0 & 81.8 & 72.8 & 61.9 & 79.1 & 78.7 & 62.9 \\
\rowcolor{c2}
RL & + \method (ours) & 91.2 & 72.2 & 94.1 & 95.1 & 88.3 & 74.4 & 87.7{\tiny \textbf{\textcolor{teal}{(+8.6\%)}}} & 88.4  & 76.8{\tiny \textbf{\textcolor{teal}{(+13.9\%)}}} \\
RL & GiGPO w/std & 94.4 & 67.5 & 94.8 & 94.4 & 79.8 & 76.4 & 86.7 & 83.1 & 65.0 \\
\rowcolor[gray]{.95}
RL & GiGPO w/o std & 96.0 & 76.5 & 91.8 & 91.3 & 71.7 & 79.5 & 86.1 & 83.5 & 67.4 \\
\rowcolor{c2}
RL & + \method (ours) & 95.2 & 79.2 & 89.1 & 94.0 & 88.8 & 95.5 & 91.2{\tiny \textbf{\textcolor{teal}{(+5.1\%)}}} & 90.7 & 79.3{\tiny \textbf{\textcolor{teal}{(+11.8\%)}}} \\
\rowcolor[gray]{.95}
RL & \textit{Dr.BoT} (GiGPO) & 98.6 & 91.4 & 93.7 & 93.8 & 85.4 & 78.4 & 90.6 & 84.1 & 68.8  \\
\rowcolor{c2}
RL & + \method (ours) & 96.4 & 86.5 & 96.1 & 99.0 & 87.6 & 91.6 & 93.2{\tiny \textbf{\textcolor{teal}{(+2.6\%)}}} & 90.9 & 81.1{\tiny \textbf{\textcolor{teal}{(+12.2\%)}}} \\
\midrule
\multicolumn{11}{c}{\textit{Qwen2.5-7B-Instruct}} \\
PT & I/O & 33.4 & 21.6 & 19.3 & 6.9 & 2.8 & 3.2 & 14.8 & 26.4 & 7.8 \\
FW & ReAct & 48.5 & 35.4 & 34.3 & 13.2 & 18.2 & 17.6 & 31.2 & 46.2 & 19.5 \\
FW & Reflexion & 62.0 & 41.6 & 44.9 & 30.9 & 36.3 & 23.8 & 42.7 & 58.1 & 28.8 \\
RL & PPO & 92.3 & 64.0 & 92.5 & 89.5 & 80.3 & 68.8 & 80.4 & 81.4 & 68.7 \\
RL & RLOO & 87.6 & 78.2 & 87.3 & 81.3 & 71.9 & 48.9 & 75.5 & 80.3 & 65.7 \\
\rowcolor[gray]{.95}
RL & GRPO & 90.8 & 66.1 & 89.3 & 74.7 & 72.5 & 64.7 & 77.6 & 79.3 & 66.1 \\
\rowcolor{c2}
RL & + \method (ours) & 93.7 & 62.4  &  97.2  & 78.0  & 83.1 & 75.5 & 85.2{\tiny \textbf{\textcolor{teal}{(+7.6\%)}}} & 92.4 & 84.6{\tiny \textbf{\textcolor{teal}{(+18.5\%)}}} \\
\rowcolor[gray]{.95}
RL & \textit{Dr.BoT} (GRPO) & 99.9 & 95.8 & 93.8 & 92.8 & 90.4 & 80.6 & 92.4  & 90.4  & 80.5 \\
\rowcolor{c2}
RL & + \method (ours) & 98.8 & 97.9 & 97.1 & 88.5 & 89.2 & 87.2 & 93.8{\tiny \textbf{\textcolor{teal}{(+1.4\%)}}} & 91.4 & \underline{84.8}{\tiny \textbf{\textcolor{teal}{(+4.3\%)}}} \\
RL & GiGPO w/std & 97.7 & 82.7 & 98.8 & 83.7 & 89.3 & 79.2 & 90.8 & 84.4 & 72.8 \\
\rowcolor[gray]{.95}
RL & GiGPO w/o std & 91.8 & 88.6 & 95.9 & 90.2 & 86.5 & 85.2 & 90.2 & 86.2 & 75.2 \\
\rowcolor{c2}
RL & + \method (ours) & 99.9 & 82.4 & 98.0 & 92.8 & 92.6 & 86.6 & \underline{94.1}{\tiny \textbf{\textcolor{teal}{(+3.9\%)}}} & 92.7 & 83.8{\tiny \textbf{\textcolor{teal}{(+8.6\%)}}} \\
\rowcolor[gray]{.95}
RL & \textit{Dr.BoT} (GiGPO) & 98.3 & 99.9 & 96.9 & 92.8 & 91.8 & 88.3 & 94.0  &  90.7  &  81.8  \\
\rowcolor{c2}
RL & + \method (ours) & 99.9 & 85.1 & 95.6  & 96.4  & 89.9  &  95.1  &  \textbf{94.7}{\tiny \textbf{\textcolor{gray}{(+0.7\%)}}}  & 92.5 & \textbf{85.7}{\tiny \textbf{\textcolor{teal}{(+3.9\%)}}} \\
\bottomrule
\end{tabular}
}}}
\end{table}

\section{Experiments}

\subsection{Experimental Setup}

\label{sec:dataset}
Three benchmarks are used:
ALFWorld~\cite{shridhar2020alfworld},
WebShop~\cite{yao2022webshop},
and DAPO-MATH-17K~\cite{yu2025dapo} (Appendix~\ref{sec:datasetdetails}).
\label{sec:baseline}
According to these benchmarks,
we respectively follow~\cite{feng2025group} and~\cite{feng2025retool} to report a range of competitive baselines (Appendix~\ref{sec:detailsbaseline}).
\label{sec:implementation}
All the training settings and hyper-parameters are detailed in Appendix~\ref{sec:implementation_detail_appdx}.

\subsection{Performance}
\label{sec:performance}

Table~\ref{tab:ALFWorldwebshop} demonstrates our effectiveness on ALFWorld and WebShop.
It is compatible with GRPO~\cite{shao2024deepseekmath},
GiGPO~\cite{feng2025group},
and our \textit{Dr.BoT}.
\method brings consistent gains across 1.5B and 7B models up to 20\%.
Such generalization benefits from the collection of successful trajectories,
which acts as a walkthrough guide to the agent.
Especially for tasks where the success rate is fairly low at the beginning,
the agent has to figure out the underlying interaction logics and summarize action plans tailored specific to each task.
The experience replay expedites the accumulation of tactics and thereafter reduces blind trials and errors.
Furthermore,
our \textit{Dr.BoT} boosts GRPO and GiGPO up to 15\%,
showcasing the validity of mixture of tricks.

\begin{table}[htbp]
\centering
\caption{Results (mean@30) on AIME 2024 \& 2025 (\%).
$^\dagger$: Official implementation
already utilizes DAPO tricks.
$^\ddagger$: Official results reported by Qwen~\cite{yang2025qwen3}.
PT stands for prompting.
}
\label{tab:aime2425}
\resizebox{0.67\linewidth}{!}{
\setlength{\tabcolsep}{1mm}{
\fontsize{9pt}{10pt}\selectfont{
\begin{tabular}{llllllll}
\toprule
\multirow{2}{*}{\textbf{Type}} & \multirow{2}{*}{\textbf{Method}} & \multirow{2}{*}{\textbf{Model}} & \multirow{2}{*}{\textbf{Tool}} & \multicolumn{2}{c}{\textbf{Context}}  & \multirow{2}{*}{\textbf{AIME24}} & \multirow{2}{*}{\textbf{AIME25}} \\
 &  &  &  & \textbf{Train} & \textbf{Test} &  &  \\
\midrule
PT & I/O & \textit{Qwen2.5-32B-Instruct} & --  & -- & 16K & 13.4  & 12.9  \\
PT & I/O & \textit{Qwen2.5-32B-Instruct} & CI & -- & 16K & 29.6 & 23.1 \\
RL & PPO$^\dagger$ & \textit{Qwen2.5-32B-Instruct} & CI & 16K & 16K  & -- & 55.0 \\
RL & GRPO$^\dagger$ & \textit{Qwen2.5-32B-Instruct} & CI & 16K & 16K & -- & 60.0 \\
RL & ReTool &\textit{Qwen2.5-32B-Instruct}& CI & 16K & 16K & 67.0 & 49.3 \\
RL & SimpleTIR &\textit{Qwen2.5-32B-Instruct}& CI & 12K & 12K & 59.9 & 49.2 \\
RL & ZeroTIR &\textit{Qwen2.5-32B-Instruct}& CI & 8K & 8K & 56.7 &  33.3 \\
RL & AFM &\textit{Qwen2.5-32B-Instruct}& CI & 32K & 32K & 66.7 & 59.8  \\
\rowcolor[gray]{.95}
RL & \textit{Dr.BoT} (GRPO) & \textit{Qwen2.5-32B-Instruct} & CI & 16K & 16K  & 64.7  & 54.0  \\
\rowcolor{c2}
RL  & + \method (ours) & \textit{Qwen2.5-32B-Instruct} & CI & 16K & 16K  &  66.3{\tiny \textbf{\textcolor{teal}{(+1.6\%)}}}  & 60.1{\tiny \textbf{\textcolor{teal}{(+6.1\%)}}} \\
\rowcolor[gray]{.95}
RL & \textit{Dr.BoT} (GRPO) & \textit{Qwen2.5-32B-Instruct} & CI &  32K & 32K  & 67.2  &  55.1  \\
\rowcolor{c2}
RL & + \method (ours) & \textit{Qwen2.5-32B-Instruct} & CI & 32K & 32K & 71.0{\tiny \textbf{\textcolor{teal}{(+3.8\%)}}} & 61.0{\tiny \textbf{\textcolor{teal}{(+5.9\%)}}} \\
\midrule
PT & I/O & \textit{Qwen3-32B-Instruct} & -- & -- & 16K & 68.5  & 53.5  \\
PT & I/O$^\ddagger$ & \textit{Qwen3-32B-Instruct} & -- & -- & 38K & 81.4 & 72.9 \\
PT & I/O & \textit{Qwen3-32B-Instruct} & CI  & --  &  16K  & 31.1 & 24.4 \\
\rowcolor[gray]{.95}
RL & \textit{Dr.BoT} (GRPO) & \textit{Qwen3-32B-Instruct} & CI &  16K  &  16K & 81.3 & 74.1 \\
\rowcolor{c2}
RL & + \method (ours) & \textit{Qwen3-32B-Instruct} & CI & 16K  & 16K  &   81.8{\tiny \textbf{\textcolor{gray}{(+0.5\%)}}}  &  \underline{78.8}{\tiny \textbf{\textcolor{teal}{(+4.7\%)}}} \\
\rowcolor[gray]{.95}
RL & \textit{Dr.BoT} (GRPO) & \textit{Qwen3-32B-Instruct} & CI & 32K & 32K & \underline{82.5} & 77.3 \\
\rowcolor{c2}
RL & + \method (ours) & \textit{Qwen3-32B-Instruct} & CI &  32K  &  32K  &  \textbf{85.6}{\tiny \textbf{\textcolor{teal}{(+3.1\%)}}}  & \textbf{80.5}{\tiny \textbf{\textcolor{teal}{(+3.2\%)}}}  \\
\bottomrule
\end{tabular}
}}}
\end{table}

Table~\ref{tab:aime2425} reports the performance 
of CI-integrated reasoning on AIME24 and AIME25.
$\textit{Dr.BoT}$ indeed outperforms recent RL baselines.
The reduced context length of Qwen3 impedes complete reasoning and answer parsing.
The agent learns to exploit the CI feedback for double-check and self-reflection.
\method achieves comparable performance with Qwen3 but using a much smaller token budget.
When the context is relaxed to 32K,
an improvement is observed on both Qwen2.5 and Qwen3,
confirming our generalization
with more interactions turns and reasoning tokens.

\subsection{Ablation Study}
\label{sec:ablationstudy}

\begin{table}[htbp]
\centering
\caption{Ablation on ALFWorld \& WebShop.
SI \& IR stand for Self-Imitation \& Intrinsic Reward.
}
\label{ablation:ALFWorldwebshop}
\resizebox{0.7\linewidth}{!}{
\setlength{\tabcolsep}{1mm}{
\fontsize{9pt}{10pt}\selectfont{
\begin{tabular}{lllllllllll}
\toprule
\multirow{2}{*}{\textbf{Type}} & \multirow{2}{*}{\textbf{Method}} & \multicolumn{7}{c}{\textbf{ALFWorld}} & \multicolumn{2}{c}{\textbf{WebShop}} \\
 &  & \multicolumn{1}{l}{Pick} & \multicolumn{1}{l}{Look} & \multicolumn{1}{l}{Clean} & \multicolumn{1}{l}{Heat} & \multicolumn{1}{l}{Cool} & \multicolumn{1}{l}{Pick2} & \multicolumn{1}{l}{All} & \multicolumn{1}{l}{Score} & \multicolumn{1}{l}{SR} \\
 \midrule
\multicolumn{11}{c}{\textit{Qwen2.5-1.5B-Instruct}} \\
\rowcolor[gray]{.95}
RL & GRPO & 85.3 & 53.7 & 84.5 & 78.2 & 59.7 & 53.5 & 72.8 & 75.8 & 56.8 \\
RL & + SI & 86.8 & 61.0 & 87.4 & 87.7 & 71.1 & 56.6 & 77.3{\tiny \textbf{\textcolor{teal}{(+4.5\%)}}} & 85.1 & 74.2{\tiny \textbf{\textcolor{teal}{(+17.4\%)}}} \\
\rowcolor{c2}
RL & + SI + IR (\method)  & 93.9 & 80.9 & 96.4 & 87.4 & 88.3 & 79.1 & 88.9{\tiny \textbf{\textcolor{teal}{(+16.1\%)}}} & 90.0 & 77.5{\tiny \textbf{\textcolor{teal}{(+20.7\%)}}} \\
\rowcolor[gray]{.95}
RL & GiGPO w/o std & 96.0 & 76.5 & 91.8 & 91.3 & 71.7 & 79.5 & 86.1 & 83.5 & 67.4 \\
RL & + SI & 93.2 & 82.5 & 96.3 & 87.4 & 92.7 & 87.5 & 90.6{\tiny \textbf{\textcolor{teal}{(+4.5\%)}}} & 89.4 & 79.0{\tiny \textbf{\textcolor{teal}{(+11.6\%)}}} \\
\rowcolor{c2}
RL & + SI + IR (\method) & 95.2 & 79.2 & 89.1 & 94.0 & 88.8 & 95.5 & 91.2{\tiny \textbf{\textcolor{teal}{(+5.1\%)}}} & 90.7 & 79.3{\tiny \textbf{\textcolor{teal}{(+11.8\%)}}} \\
\midrule
\multicolumn{11}{c}{\textit{Qwen2.5-7B-Instruct}} \\
\rowcolor[gray]{.95}
RL & GRPO & 90.8 & 66.1 & 89.3 & 74.7 & 72.5 & 64.7 & 77.6 & 79.3 & 66.1 \\
RL & + SI & 93.2 & 82.5 & 96.3 & 87.4 & 92.7 & 87.5  & 90.6{\tiny \textbf{\textcolor{teal}{(+13.0\%)}}} & 90.4 & 83.4{\tiny \textbf{\textcolor{teal}{(+17.3\%)}}} \\
\rowcolor{c2}
RL & + SI + IR (\method)  & 93.7 & 62.4  &  97.2  & 78.0  & 83.1 & 75.5 & 85.2{\tiny \textbf{\textcolor{teal}{(+7.6\%)}}} & 92.4 & \underline{84.6}{\tiny \textbf{\textcolor{teal}{(+18.5\%)}}} \\
\rowcolor[gray]{.95}
RL & GiGPO w/o std & 91.8 & 88.6 & 95.9 & 90.2 & 86.5 & 85.2 & 90.2 & 86.2 & 75.2 \\
RL & + SI & 96.1 & 81.9 & 98.4 & 95.3 & 94.5 & 83.9 & \underline{93.6}{\tiny \textbf{\textcolor{teal}{(+3.4\%)}}}  &  94.6  &  \textbf{87.5}{\tiny \textbf{\textcolor{teal}{(+12.3\%)}}}  \\
\rowcolor{c2}
RL & + SI + IR (\method) & 99.9 & 82.4 & 98.0 & 92.8 & 92.6 & 86.6 & \textbf{94.1}{\tiny \textbf{\textcolor{teal}{(+3.9\%)}}} & 92.7 & 83.8{\tiny \textbf{\textcolor{teal}{(+8.6\%)}}} \\
\bottomrule
\end{tabular}
}}}
\end{table}

\begin{table}[htbp]
\centering
\caption{Ablation on AIME 2024 \& 2025 (\%).
SI \& IR stand for Self-Imitation \& Intrinsic Reward.
}
\label{ablation:aime2425}
\resizebox{0.7\linewidth}{!}{
\setlength{\tabcolsep}{1mm}{
\fontsize{9pt}{10pt}\selectfont{
\begin{tabular}{llllllll}
\toprule
\multirow{2}{*}{\textbf{Type}} & \multirow{2}{*}{\textbf{Method}} & \multirow{2}{*}{\textbf{Model}} & \multirow{2}{*}{\textbf{Tool}} & \multicolumn{2}{c}{\textbf{Context}}  & \multirow{2}{*}{\textbf{AIME24}} & \multirow{2}{*}{\textbf{AIME25}} \\
 &  &  &  & \textbf{Train} & \textbf{Test} &  &  \\
\midrule
\rowcolor[gray]{.95}
RL  & \textit{Dr.BoT} (GRPO) & \textit{Qwen2.5-32B-Instruct} & CI & 16K & 16K & 64.7 & 54.0 \\
RL & + SI & \textit{Qwen2.5-32B-Instruct} & CI & 16K & 16K & 63.8{\tiny \textbf{\textcolor{gray}{(-0.9\%)}}} & 56.9{\tiny \textbf{\textcolor{teal}{(+2.9\%)}}} \\
\rowcolor{c2}
RL  & + SI + IR (\method) & \textit{Qwen2.5-32B-Instruct} & CI & 16K & 16K & 66.3{\tiny \textbf{\textcolor{teal}{(+1.6\%)}}}  & 60.1{\tiny \textbf{\textcolor{teal}{(+6.1\%)}}}   \\
\rowcolor[gray]{.95}
RL  & \textit{Dr.BoT} (GRPO)& \textit{Qwen2.5-32B-Instruct} & CI & 32K & 32K & 67.2  & 55.1 \\
RL  & + SI & \textit{Qwen2.5-32B-Instruct} & CI & 32K & 32K & 66.0{\tiny \textbf{\textcolor{purple}{(-1.2\%)}}} & 60.5{\tiny \textbf{\textcolor{teal}{(+5.4\%)}}} \\
\rowcolor{c2}
RL  & + SI + IR (\method) & \textit{Qwen2.5-32B-Instruct} & CI & 32K & 32K & 71.0{\tiny \textbf{\textcolor{teal}{(+3.8\%)}}} & 61.0{\tiny \textbf{\textcolor{teal}{(+5.9\%)}}} \\
\midrule
\rowcolor[gray]{.95}
RL  & \textit{Dr.BoT} (GRPO) & \textit{Qwen3-32B-Instruct} & CI & 16K & 16K & 81.3 & 74.1 \\
RL & + SI & \textit{Qwen3-32B-Instruct} & CI & 16K & 16K & 81.2{\tiny \textbf{\textcolor{gray}{(-0.1\%)}}}  &  75.8{\tiny \textbf{\textcolor{teal}{(+1.70\%)}}} \\
\rowcolor{c2}
RL  & + SI + IR (\method) & \textit{Qwen3-32B-Instruct} & CI & 16K & 16K & 81.8{\tiny \textbf{\textcolor{gray}{(+0.5\%)}}}  &  \underline{78.8}{\tiny \textbf{\textcolor{teal}{(+4.70\%)}}} \\
\rowcolor[gray]{.95}
RL  & \textit{Dr.BoT} (GRPO) & \textit{Qwen3-32B-Instruct} & CI & 32K & 32K & \underline{82.5} & 77.3 \\
RL  & + SI & \textit{Qwen3-32B-Instruct} & CI & 32K & 32K & 81.8{\tiny \textbf{\textcolor{gray}{(-0.7\%)}}}  &  78.2{\tiny \textbf{\textcolor{gray}{(+0.9\%)}}} \\
\rowcolor{c2}
RL  & + SI + IR (\method) & \textit{Qwen3-32B-Instruct} & CI & 32K & 32K & \textbf{85.6}{\tiny \textbf{\textcolor{teal}{(+3.1\%)}}}  &  \textbf{80.5}{\tiny \textbf{\textcolor{teal}{(+3.2\%)}}}  \\
\bottomrule
\end{tabular}
}}}
\end{table}

\paragraph{Self-Imitation.}
The SIL improves baselines consistently across model scales (Table~\ref{ablation:ALFWorldwebshop}).
Since either 1.5B or 7B models perform poorly at the early stage (i.e., success rate $<15\%$),
past experiences are quite beneficial to explore promising strategies.
The re-use of trajectories facilitates convergence and prevents mechanical trials especially for small agents.
Table~\ref{ablation:aime2425} shows that AIME24 dropped a bit by self-imitation but AIME25 still gets improved.
Such fluctuation is related to the phenomenon (Figure~\ref{fig:toolcall}) where the imitation of samples with multiple tool calls leads to rapid increase of interaction turns and thereafter causes training instability.
The competition between different reward terms affects the robust selection of good experience, ultimately degrading the effectiveness of SIL.

\paragraph{Intrinsic Reward.}
The rewarding of interaction turns benefit 1.5B models consistently (Table~\ref{ablation:ALFWorldwebshop}).
Two 7B outliers are found where the self-imitation alone brings the most performance gains.
Such exception might be related to both the task definition and the RL algorithm.
One should experiment with different combinations in practice.
Table~\ref{ablation:aime2425} shows that the intrinsic reward is indispensable for both Qwen2.5 and 3 because it encourages transformation from text-based reasoning to tool-integrated reasoning.
It promotes frequent tool calling and such rich observation signals motivate the agent to correct coding errors,
check the validity of the answer,
and reflect on alternative solutions.

\subsection{Generalization on Vision-Language Agents}
\label{sec:generalvlagent}

\begin{table}[htbp]
\centering
\vspace{-1em}
\begin{minipage}{0.33\linewidth}
\caption{Success rate (\%) of the visual agent for playing Sokoban.}
\label{tab:sokoban}
\centering
\small
\resizebox{\textwidth}{!}{
\begin{tabular}{lll}
\toprule
\textbf{Type} & \textbf{Method} & \textbf{Sokoban} \\
\midrule
\multicolumn{3}{c}{\textit{Qwen2.5-VL-3B-Instruct}} \\
PT & I/O &  11.7  \\
\rowcolor[gray]{.95}
RL & GRPO &  67.1  \\
\rowcolor{c2}
RL & + \method (ours) & 86.7{\tiny \textbf{\textcolor{teal}{(+19.6\%)}}} \\
\rowcolor[gray]{.95}
RL & \textit{Dr.BoT} (GRPO) & 76.0 \\
\rowcolor{c2}
RL & + \method (ours) & 85.4{\tiny \textbf{\textcolor{teal}{(+9.4\%)}}} \\
RL & GiGPO w/ std &  76.9  \\
\rowcolor[gray]{.95}
RL & GiGPO w/o std &  81.0  \\
\rowcolor{c2}
RL & + \method (ours) & 87.7{\tiny \textbf{\textcolor{teal}{(+6.7\%)}}} \\
\rowcolor[gray]{.95}
RL & \textit{Dr.BoT} (GiGPO) & 81.3 \\
\rowcolor{c2}
RL & + \method (ours) & 87.9{\tiny \textbf{\textcolor{teal}{(+6.6\%)}}} \\
\bottomrule
\end{tabular}
}
\end{minipage}
\hspace{0.01\linewidth}
\begin{minipage}{0.49\linewidth}  
\vskip 0.1in
\begin{center}
\begin{subfigure}{.4\textwidth}
  \centering
  \includegraphics[width=\textwidth]{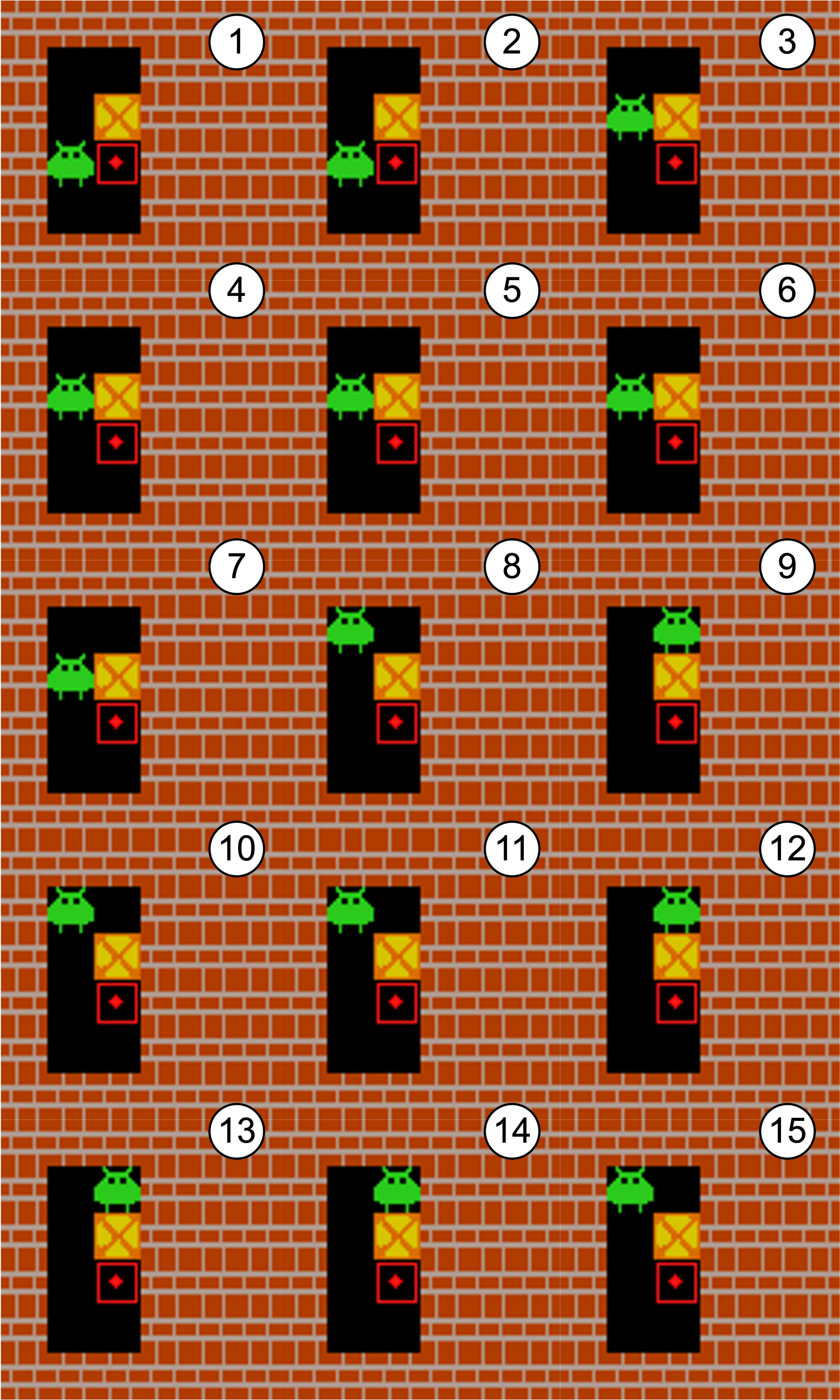}
  \caption{Before (step 15).}
\end{subfigure}
\hspace{0.1\linewidth}
\begin{subfigure}{.4\textwidth}
  \centering
  \includegraphics[width=\textwidth]{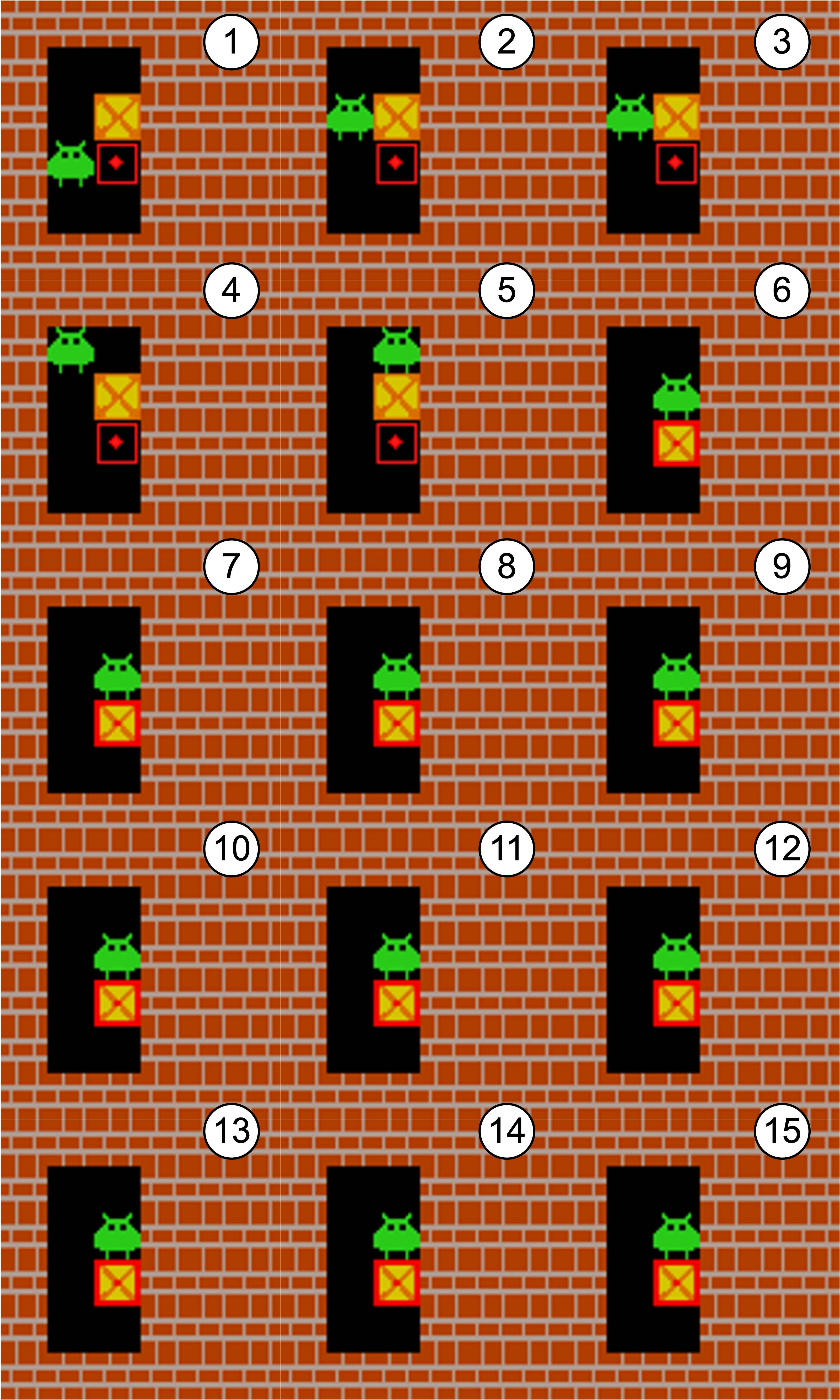}
  \caption{After (step 125).}
\end{subfigure}%
\end{center}
\vspace{-1em}
\captionof{figure}{The agent learns to push the box.
}
\label{fig:sokoban}
\end{minipage}\vspace{-1em}
\end{table}

To test whether the proposed \method is still complimentary to existing GRPO-like algorithms on training visual agents,
we follow~\cite{feng2025group} to conduct experiments on the popular visual game Sokoban~\cite{SchraderSokoban2018}.
In this setting,
the Qwen2.5-VL-3B-Instruct~\cite{bai2025qwen2} is adopted as the agentic LLM to solve the puzzle game where the player must push the boxes along the grid towards target positions without hitting the walls.
It challenges the agent on spatial comprehension and long-term planning capabilities.
The grid size is of $6\times6$ and the visual agent receives both the visual (RGB arrays) and textual inputs as states.
As shown in Table~\ref{tab:sokoban},
the proposed method generally improves the performance on Sokoban with either GRPO, GiGPO, and the proposed \textit{Dr.BoT} baselines.
At first,
the visual agent is unaware of the winning logic behind the game and wanders around for "aimlessly" exploration (see Figure~\ref{fig:sokoban}).
After optimization,
it not only comprehends the spatial relationship to control the box but also learns to stop moving when the task is completed.

\subsection{Generalization on Search-augmented QA Tasks}
\label{sec:searchaugqa}
{

\begin{table}[htbp]
\centering
\caption{Results on search-augmented QA Tasks.
}
\label{ablation:seachaugment}
\resizebox{0.9\linewidth}{!}{
\setlength{\tabcolsep}{1mm}{
\fontsize{9pt}{10pt}\selectfont{
\begin{tabular}{llllllllll}
\toprule
\multirow{2}{*}{\textbf{Type}} & \multirow{2}{*}{\textbf{Method}} & \multicolumn{3}{c}{\textbf{Single-Hop QA}} & \multicolumn{4}{c}{\textbf{Multi-Hop QA}} & \multirow{2}{*}{\textbf{Avg.}} \\
 &  & NQ & TriviaQA & PopQA & HotpotQA & 2Wiki & MuSiQue & Bamboogle &  \\
\midrule
\multicolumn{10}{c}{\textit{Qwen2.5-7B-Instruct}} \\
\rowcolor[gray]{.95}
RL & Search-R1 (num\_steps=500) & 39.3  &  61.0  &  39.7  & 37.0  &  40.1  &  14.6  & 36.8  &  38.5  \\
RL & \method (num\_steps=300) & 35.7  & 62.7  & 34.5 & 46.9 & 43.4 & 17.2 & 44.8 & 40.7 \\
\rowcolor{c2}
RL & \method (num\_steps=550) &  43.7  &  66.8  &  42.8 &  48.1 &  45.1  &  21.1 & 50.4  & 45.4  \\
\multicolumn{10}{c}{\textit{Qwen2.5-14B-Instruct}} \\
\rowcolor[gray]{.95}
RL & Search-R1 (num\_steps=500) & 48.8 & 67.7 & 48.2 & 45.5 & 47.0 & 21.1 & 51.6  &  49.1  \\
RL & \method (num\_steps=300) & 47.6 & 69.3 & 47.8 & 48.5 & 48.8 & 26.7 & 56.6 & 49.3 \\
\rowcolor{c2}
RL & \method (num\_steps=550) & 48.8 & 70.6 & 48.0 & 50.2 & 48.8 & 27.5  & 53.6 & 49.6 \\ 
\bottomrule
\end{tabular}
}}}
\end{table}

To evaluate the performance of \method on knowledge-intensive reasoning tasks,
we conduct experiments on search-augmented QA tasks,
including the single-hop QA datasets (NQ~\cite{kwiatkowski2019natural},
TriviaQA~\cite{joshi2017triviaqa},
and PopQA~\cite{mallen2023not}) and multi-hop QA datasets (HotpotQA~\cite{yang2018hotpotqa},
2Wiki~\cite{ho2020constructing},
MuSiQue~\cite{trivedi2022musique},
and Bamboogle~\cite{press2023measuring}).
We follow the experimental settings of SearchR1~\cite{jin2025search,gao2025beyond} to launch the local wiki-18 retrieval service.
We adopt the Hierarchical Navigable Small World (HNSW) CPU indexing as approximation of nearest neighbor retrieval.
Our \method with GRPO improves over the Search-R1 baseline on average,
especially on the multi-hop QA benchmarks.
Such multi-hop QA datasets require reasoning for problem decomposition and several turns of information seeking.
In this case, our intrinsic reward that encourages multiple tool uses for broad exploration prevents arbitrary conclusions with only one or two searches.
Our \method respectively requires $\sim14.48$ and $\sim14.42$ calls for 7B and 14B models, respectively.
Such behavior is expected due to the stimulation of exploration at the beginning.
Despite the QA tasks are relatively short-horizon,
the agent still benefits from the detailed decomposition of the complex queries with cross-validation via step-by-step searching.
Note that our retrieval service adopts the HNSW of E5 embedding for efficient training,
which slightly impedes performance~\cite{jin2025empirical}.
Furthermore, we also notice that the number of training steps exerts an important effect on cultivation of searching capability,
highlighting that the scaling of training steps is critical to the RL performance.
Compared with the Search-R1,
our \method achieves better performance under comparable training steps,
demonstrating its effectiveness.
}

\subsection{More Discussions}
\label{sec:discussionintotal}

Due to the page limit,
discussions on 
theoretical analysis on convergence~\ref{sec:theoretical},
hyper-parameters~\ref{sec:discusshyper},
qualitative analysis~\ref{sec:qualitative},
training cost and complexity~\ref{sec:trainingcost},
and future research directions~\ref{sec:future} are presented in the appendix.
One could easily adapt \method to training any (M)LLM-driven agents robustly without binding to a specific optimization algorithm.

\section{Conclusions and Limitations}

In this paper,
we target the pivotal challenge of balancing exploration and exploitation in RL training of LLM agents.
Our proposed solution, \method\raisebox{-0.1em}{\includegraphics[scale=0.12]{figures/spear-logo.png}}, extends the vanilla SIL by
advantage recalibration, scheduled entropy control, and intrinsic rewards.
These components work in a curriculum manner to prevent policy collapse and excessive uncertainty, progressively guiding the policy through a smooth transition between exploration and exploitation. 
In addition, we propose a strong baseline \textit{Dr.BoT} tailored for agentic RL with existing bag-of-tricks verified from numerical industrial practices.
Empirical results across tasks and models showcase \method's superiority over existing methods, with performance boosts and acceptable computational overhead.
The effectiveness of our \method underscores the value of learning from past experiences while managing policy entropy, offering a robust framework for training LLMs with strong reasoning and tool integration skills.


There exist two potential limitations:
1) The vague definition of good experiences under highly complex, stochastic environments with unreliable tools. In such cases, observations can be noisy and severely degrade the feasibility of the task. The sparse outcome reward cannot distinguish between good and bad experiences and thereafter the relative advantages might be simply attributed to randomness instead of the agent's behavior. We suggest a possible solution that more fine-grained, stepwise supervision should be enforced.
For example,
a step-wise process reward that evaluates the logical consistency~\cite{zhang2025rlvmr} of the agent's thought and action might be helpful.
2) The rigidity of entropy control which relies on prior-based scheduling and covariance-based clipping. In the present study, 
the proposed scheduling and clipping designs might not be optimal for all kinds of agentic tasks. A more adaptive solution lies in the policy's self-confidence on decisions under each observation.
One might use the token-level dynamic reweighting for SIL~\cite{wu2025generalization} which avoids over-concentration on certain low-probability reference tokens in the replay buffer. Similarly, the clipping could depend on token probability instead of the bounded random sampling.
We leave the exploration mentioned above as a promising direction for improvement in the future.

\clearpage
\newpage
\section*{Contributions}
\paragraph{Authors}
Yulei Qin\textsuperscript{\rm 1*}\quad Xiaoyu Tan\textsuperscript{\rm 1*}\quad Zhengbao He\textsuperscript{\rm 1,2*}\quad Gang Li\textsuperscript{\rm 1}\quad Haojia Lin\textsuperscript{\rm 1}\quad Zongyi Li\textsuperscript{\rm 1}\quad Zihan Xu\textsuperscript{\rm 1}\quad Yuchen Shi\textsuperscript{\rm 1}\quad Siqi Cai\textsuperscript{\rm 1}\quad Renting Rui\textsuperscript{\rm 1,2}\quad Shaofei Cai\textsuperscript{\rm 1,3}\quad Yuzheng Cai\textsuperscript{\rm 1,4}\quad Xuan Zhang\textsuperscript{\rm 1,4}\quad Sheng Ye\textsuperscript{\rm 1,5}\quad Ke Li\textsuperscript{\rm 1}\quad Xing Sun\textsuperscript{\rm 1}

\paragraph{Affiliations}
\textsuperscript{\rm 1}Tencent Youtu Lab\quad \textsuperscript{\rm 2}Shanghai Jiao Tong University\quad \textsuperscript{\rm 3}Peking University\\\quad\textsuperscript{\rm 4}Fudan University\quad\textsuperscript{\rm 5}Xiamen University

\paragraph{$^*$Equal Contributions}
Yulei Qin\quad Xiaoyu Tan\quad Zhengbao He

\paragraph{Acknowledgments}
We greatly thank the VeRL~\cite{sheng2024hybridflow} and the VeRL-agent~\cite{feng2025group} communities for their implementation of various RL training and inference frameworks for multi-turn agent development.

\clearpage
\newpage
\appendix
\section{Appendix}

\subsection{Summary of the Appendix}
\label{sec:actionappdx}
{
In the appendix, we provide detailed explanations on the following.
\begin{itemize}
    \item Descriptions about the Action Space
    \item Brief Introduction to the PPO and GRPO
    \item PseudoCode of the SPEAR
    \item Visualization of the Curriculum Schedule
    \item Definition of the Reward Function
    \item Descriptions about RL Bag-of-Tricks
    \item Theoretical Analysis on Convergence
    \item Descriptions of the Data and Environment
    \item Choice of Baselines
    \item Implementation Details
    \item Discussions and Guidelines on Hyper-parameters
    \item Qualitative Analysis
    \item Training Cost and Complexity
    \item Future Research Directions
\end{itemize}
}

\subsection{Detailed Action Space}
\label{sec:actionspacedetails}

The following contents correspond to Section~\ref{sec:actionspace} in the main text.

\paragraph{TextWorld Embodied Tool.} The embodied actions follows ALFWorld~\cite{shridhar2020alfworld} where a language-driven agent interacts with the TextWorld~\cite{cote2018textworld}. It allows the agent to take one of the following high-level actions: \texttt{goto \{recep\}}, \texttt{take \{obj\} from \{recep\}}, \texttt{put \{obj\} in/on \{recep\}}, \texttt{open \{recep\}}, \texttt{close \{recep\}}, \texttt{toggle \{obj\}\{recep\}}, \texttt{clean \{obj\} with \{recep\}}, \texttt{heat \{obj\} with \{recep\}}, and \texttt{cool \{obj\} with \{recep\}}, where \texttt{\{obj\}} and \texttt{\{recep\}} denote \texttt{objects} and \texttt{receptacles}, respectively.

\paragraph{Web Browsing Tool.} The definition of web browsing follows WebShop~\cite{yao2022webshop} where only two actions are allowed: \texttt{search[query]} and \texttt{choose[button]} where \texttt{query} and \texttt{button} respectively stand for searching \texttt{query} and clickable elements such as \texttt{back to search}, \texttt{prev/next page}, \texttt{\{product title\}}, \texttt{\{option\}}, \texttt{\{desc/overview\}}, \texttt{previous}, and \texttt{buy}.

\paragraph{Code Interpreter Tool.} The code interpreter executes the code generated by the language model and return both the \texttt{stdout} and \texttt{stderr}. If the code runs correctly, the \texttt{stdout} contains the output. On the other hand, the compiler error messages are provided for the next-round correction. We follow~\cite{feng2025retool} to deploy a SandBox~\cite{bytedanceseedfoundationcodeteam2025fullstackbenchevaluatingllms} service that receives execution requests from the interpreter tool. In addition, we add a reminder in the \texttt{stdout} for empty output when the LLM forgets to \texttt{print} computation results: \textit{Empty stdout! You might forget to print the answer}. For non-empty \texttt{stderr}, we also add an instruction as hint: \textit{Errors occurred! Check your code}.



\subsection{Detailed Policy Optimization Algorithms}
\label{sec:ppogrpo}

The following contents correspond to Section~\ref{sec:policyoptimization} in the main text.

\paragraph{Proximal Policy Optimization (PPO).}
Typically,
PPO optimizes the following:
\begin{equation}
    \mathcal{J}({\pi_{\theta}})=\mathbb{E}_{x\sim p(X), \mathbf{a}\sim\pi_{\theta}(\cdot|x,\mathbf{s})}\Biggr[\mathcal{R}(x, \mathbf{s}, \mathbf{a})-\beta D_{\text{KL}}[\pi_\theta(\cdot|x,\mathbf{s})||\pi_{\text{ref}}(\cdot|x,\mathbf{s})]\Biggr],\\
\end{equation}
where $\mathcal{R}(x, \mathbf{s}, \mathbf{a})=\sum_{t=1}^{T}r_t(x, \mathbf{s}_t, \mathbf{a}_t)$ is the return~\cite{sutton1998reinforcement} for the trajectory and $\pi_{\text{ref}}$ is the reference policy model.
The KL divergence proposed~\cite{christiano2017deep} to prevent the policy $\pi_{\theta}$ from deviating greatly from the reference $\pi_{\text{ref}}$ ($\beta>0$).
In consideration of the simplicity, we follow TULU 3~\cite{lambert2024tulu} to adopt RL with the verifiable reward where the rule-based verifiers are designed to provide the outcome reward signal $r$ instead of the reward model $r_\theta$.
In addition, we follow~\cite{liu2025understanding} to drop the KL term by setting $\beta=0$, which not only emphasizes agent performance but also saves memory and computation during training.

\paragraph{Group Relative Policy Optimization (GRPO).}
Specifically,
the policy model $\pi_{\theta_{\text{old}}}$ from the previous iteration generates a group of $G$ individual trajectories $\{\tau_{i}\}_{i=1}^{G}$.
GRPO updates the policy $\pi_{\theta}$ by maximizing the objective below.
\begin{equation}\label{eq:grpo_all}
\begin{aligned}
    \mathcal{J}_{\text{GRPO}}(\pi_\theta)&=
     \mathbb{E}_{x\sim p(X), \{\tau_{i}\}_{i=1}^{G}\sim \pi_{\theta_{old}}(\cdot|x)}\frac{1}{G}\sum_{i=1}^{G}\mathcal{J}_{\text{GRPO}}^{i},\\
     \tau_{i}&=\{(\mathbf{s}_1^{i}, \mathbf{a}_1^{i}, R_1^{i}), (\mathbf{s}_2^{i}, \mathbf{a}_2^{i}, R_2^{i}),...,(\mathbf{s}_{T}^{i}, \mathbf{a}_{T}^{i}, R_{T}^{i})\},\\
\end{aligned}
\end{equation}
\begin{equation}\label{eq:grpo_vanilla}
    \mathcal{J}_{\text{GRPO}}^{i}=
    \frac{1}{T}\sum_{t=1}^{T}\min\Biggr[r^{i}_{t}(\theta)\hat{A}^{i}_{t}, \text{clip}[r^{i}_{t}(\theta), 1-\epsilon, 1+\epsilon]\hat{A}^{i}_{t}\Biggr]-
    \beta D_{\text{KL}}^{i}(\pi_{\theta}||\pi_{\text{ref}}),
\end{equation}
\begin{equation}\label{eq:adv_vanilla}
    r^{i}_{t}=\frac{\pi_{\theta}(\mathbf{a}^{i}_{t}|x,\mathbf{s}^{i}_{t})}{\pi_{\theta_{\text{old}}}(\mathbf{a}^{i}_{t}|x,\mathbf{s}^{i}_{t})},
    \hat{A}^{i}_{t}=\frac{R^{i}-\bar{R}}{\text{std}(\{R^{i}\}_{i=1}^{G})},\bar{R}=\text{mean}(\{R^{i}\}_{i=1}^{G}),
\end{equation}
\begin{equation}\label{eq:kl}
    D_{\text{KL}}^{i}(\pi_{\theta}||\pi_{\text{ref}})=\frac{\pi_{\text{ref}}(\mathbf{a}^{i}_{t}|x,\mathbf{s}^{i}_{t}))}{\pi_{\theta}(\mathbf{a}^{i}_{t}|x,\mathbf{s}^{i}_{t})}-\log\frac{\pi_{\text{ref}}(\mathbf{a}^{i}_{t}|x,\mathbf{s}^{i}_{t}))}{\pi_{\theta}(\mathbf{a}^{i}_{t}|x,\mathbf{s}^{i}_{t}))}-1.
\end{equation}

\subsection{Pseudo Code}
\label{sec:pseudocode}

The following contents correspond to Section~\ref{sec:method} in the main text.

Algorithm~\ref{alg:sal} summarizes the full training procedure of the proposed \method.
It is noted that our \method is compatible with various baselines such as GRPO~\cite{shao2024deepseekmath} and GiGPO~\cite{feng2025group},
enjoying a high-level of generalization.
Specifically,
the algorithm is featured by:
1) Maintenance of a replay buffer and a baseline buffer that respectively stores the trajectories for good experience replay and estimates the current policy's average performance;
2) Recalibration of the previous advantages for off-policy update;
3) Regularization against the pre-mature entropy collapsing;
4) Shaping of the composite intrisic rewards for dominance of the outcome reward.

Compared with the vanilla GRPO-like training,
the proposed method only introduced:
1) Additional policy update iterations positively associated with the number of $N_{\mathcal{D}}$ in terms of computational complexity;
2) A replay buffer of the size $N_{\mathcal{D}}$ and a baseline performance buffer of the size $N_{\mathcal{D}_R}$ in terms of space complexity.

Since we re-utilize previous trajectories without completely re-computing the rollout generation, log-probability estimation, and the advantages,
such operations are light-weight and incur minimal computation overhead.
In the present study,
we empirically set $N_{\mathcal{D}}=2048$ without meticulous hyper-parameter tuning.
For both ALFWorld, WebShop, and Sokoban,
the number of trajectories per data batch is the product of train\_batch\_size$\times$n\_samples\_per\_prompt=256 and there exist around 4K turn-level training samples under the VeRL-agent~\cite{feng2025group} framework.
For the DAPO-MATH-17K,
the number of trajectories per data batch is 2048 and there exist exactly 2048 trajectory-level training samples under the VeRL~\cite{sheng2024hybridflow} framework.
In this case,
our replay buffer reaches its full capacity around every two or three training steps on average for all experiments.
For each policy update by self-imitation,
the number of iterations is comparable to that of the vanilla policy update by GRPO under the present settings.
The detailed analyses on the training cost and complexity can also be found in Section~\ref{sec:trainingcost}.

\definecolor{darkgreen}{cmyk}{0.8,0.0,1.0,0.5}
\begin{algorithm}[htbp]
\caption{Training Agentic LLMs with \method}
\label{alg:sal}
\begin{algorithmic}[1]
\REQUIRE Initial policy $\pi_{\theta_{\text{old}}}$, data distribution $p(X)$, clipping bounds $\epsilon_{\text{lb}}$, $\epsilon_{\text{ub}}$, KL penalty $\beta$ ($\beta=0$), replay buffer $\mathcal{D}$ with buffer size $N_{\mathcal{D}}$, intra-group baseline buffer $\mathcal{D}_{R}$ with buffer size $N_{\mathcal{D}_{R}}$, the warm-up factor $\gamma$ with the number of warm-up steps $T_{\text{warm-up}}$, covariance clipping bounds $\omega_{\text{lb}}$, $\omega_{\text{ub}}$, the covariance-based clipping ratio $\lambda$ ($\lambda=0.02$), the decay factor $\mu$ with the number of decay steps$T_{\text{decay}}$, the group size $G$, the maximum allowed interaction turns $T$.
\ENSURE Updated policy $\pi_{\theta}$
\STATE Initialze $\mathcal{D}=\emptyset$ and $\mathcal{D}_{R}=\emptyset$
\FOR{each training step $t_{\text{iter}}$}
    \STATE Update the old policy model: $\theta_{\text{old}}\leftarrow\theta$
    \STATE {\color{darkgreen}{\# Repeat batch sampling and rollout generation for trajectories}}
    \STATE Sample data batch with each unique sample $x\sim p(X)$
    \STATE {\color{darkgreen}{\# Sample $G$ trajectories $\{\tau_i\}^{G}_{i=1}$ for each $x$ }}
    \FOR{$i=1$ to $G$}
        \STATE Initialize environment states $\mathbf{s}_{1}^{i}$
        \STATE {\color{darkgreen}{\# Sample at most $T$ actions}}
        \FOR{$t=1$ to $T$}
            \STATE Sample action $\mathbf{a}^{i}_{t}\sim\pi_{\theta}(\cdot|x,\mathbf{s}^{i}_{t})$
            \STATE Execute actions, receive rewards $R^{i}_{t}$, observe the new states $\mathbf{s}^{i}_{t+1}$
        \ENDFOR
        \STATE Organize the trajectory $\tau_{i}=\{(\mathbf{s}_1^{i}, \mathbf{a}_1^{i}, R_1^{i}), (\mathbf{s}_2^{i}, \mathbf{a}_2^{i}, R_2^{i}),...,(\mathbf{s}_{T}^{i}, \mathbf{a}_{T}^{i}, R_{T}^{i})\}$
    \ENDFOR
    \STATE {\color{darkgreen}{\# Apply intrinsic reward shaping for advantage estimation}}
    \STATE Compute the vanilla objective $ \mathcal{J}_{\text{GRPO}}(\pi_\theta)$ via Equation~\ref{eq:grpo_all} with the decay-scheduled $R^{i}$ via Equation~\ref{eq:rewardtotal}
    \STATE {\color{darkgreen}{\# Maintain the replay buffer and the baseline buffer}}
    \STATE $\mathcal{D}_{R}\leftarrow \mathcal{D}_{R}\cup \{\bar{R}\}, \bar{R}=\text{mean}(\{R^{i}\}_{i=1}^{G})$
    \WHILE{$|\mathcal{D}_{R}| > N_{\mathcal{\mathcal{D}_{R}}}$}
        \STATE Pop the oldest baseline $\mathcal{D}_{R}\leftarrow \mathcal{D}_{R} \setminus \{ \bar{R}_0 \}$
    \ENDWHILE
    \IF{$|\mathcal{D}| < N_{\mathcal{D}}$}
        \STATE {\color{darkgreen}{\# Add trajectories into the buffer only when their advantages are positive}}
        \STATE $\mathcal{D}\leftarrow \mathcal{D}\cup \{\tau_{i}|\hat{A}_{i}>0\}$
        \STATE {\color{darkgreen}{\# Apply on-Policy update with the vanilla GRPO}}
        \STATE Update policy by maximizing objective $\mathcal{J}_{\text{GRPO}}(\pi_\theta)$
    \ELSE
        \STATE {\color{darkgreen}{\# Recalibrate the advantage}}
        \STATE Compute the newly estimated advantage $\tilde{A}_j$ for all $\tau_j\in\mathcal{D}$ via Equation~\ref{eq:recalibrateadv}
        \STATE Only keep $\tau_j$ with positive $\tilde{A}_j$ as $\mathcal{D}\leftarrow \{\tau_j|\tilde{A}_j>0,\forall\tau_j\in\mathcal{D}\}$
        \STATE {\color{darkgreen}{\# Apply regularization on self-imitation learning}}
        \STATE Compute the self-imitation objective $\tilde{\mathcal{J}}_{\text{GRPO}}^{\text{SIL-R}}(\pi_\theta)$ via Equation~\ref{eq:grpo_sal_reg}  with covariance-based clipping via Equation~\ref{eq:regularizationmask}
        \STATE Apply the warm-up schedule for the total objective $\mathcal{J}_{\text{Total}}(\pi_\theta)$ via Equation~\ref{eq:grpo_sal}
        \STATE {\color{darkgreen}{\# Apply both the on-policy and the off-policy update for self-imitation}}
        \STATE Update policy by maximizing objective $\mathcal{J}_{\text{Total}}(\pi_\theta)$
        \STATE Reset the replay buffer $\mathcal{D}\leftarrow\emptyset$
    \ENDIF
\ENDFOR
\RETURN $\pi_{\theta}$
\end{algorithmic}
\end{algorithm}

\clearpage
\newpage

\subsection{Policy Entropy}
\label{sec:policyentropydef}

The following contents are mentioned in Section~\ref{sec:curriculumsil} in the main text.

The policy entropy quantifies the confidence inherent in the actions triggered off by the LLM.
Under the context of agent tasks,
we measure the average entropy of the entire trajectory $\tau$ for the policy model via sequence-mean-token-sum in accordance with the Dr.GRPO technique~\cite{liu2025understanding}.
Given the training data batch $\mathcal{D}_B$,
the entropy is defined as:
\begin{equation}\label{eq:entropysum}
    \mathcal{H}(\pi_\theta, \mathcal{D}_B)=-\mathbb{E}_{\mathcal{D}_B, \pi_\theta}[\log \pi_\theta(\tau|x)]=-\frac{1}{|\mathcal{D}_B|}\sum_{x\in\mathcal{D}_B, x\sim p(X)}\sum_{(\mathbf{s}_{t},\mathbf{a}_{t})\in\tau}\mathbb{E}_{\mathbf{a}_{t}\sim \pi_\theta}[\log \pi_\theta(\mathbf{a}_{t}|x,\mathbf{s}_t)]
\end{equation}

\subsection{Curriculum Schedule}
\label{sec:curriculumschedule}

\begin{figure}[htbp]
\begin{center}
\begin{subfigure}{.4\textwidth}
  \centering
  \includegraphics[width=\textwidth]{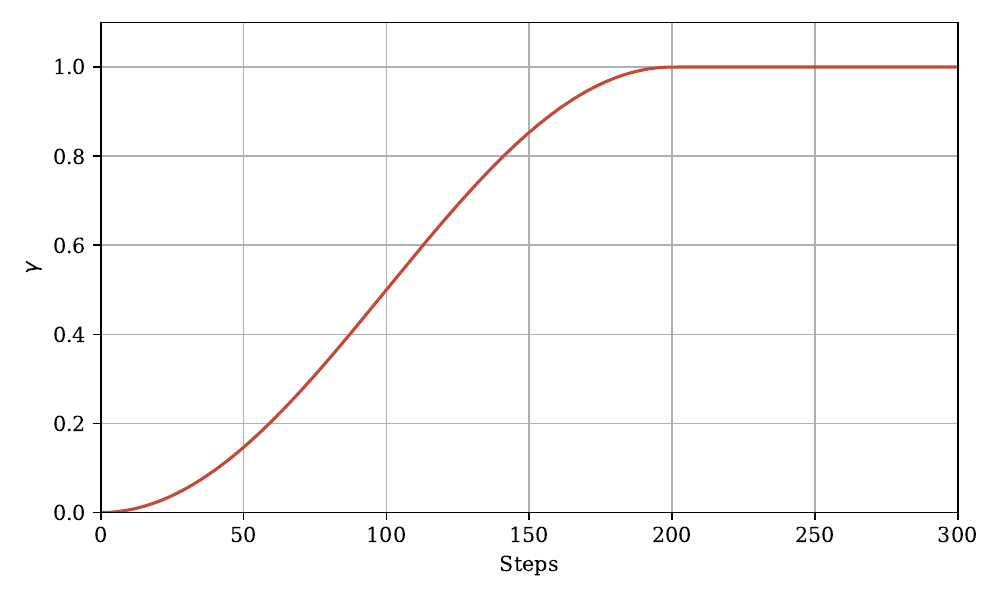}
  \caption{Visualization of $\gamma$ for the SIL term with $T_{\text{warm-up}}=200$.
The weight of SIL loss gradually increases from 0 to 1 in the first $T_{\text{warm-up}}$ steps.}
  \label{fig:gammacurve}
\end{subfigure}
\hspace{1em}
\begin{subfigure}{.4\textwidth}
  \centering
  \includegraphics[width=\textwidth]{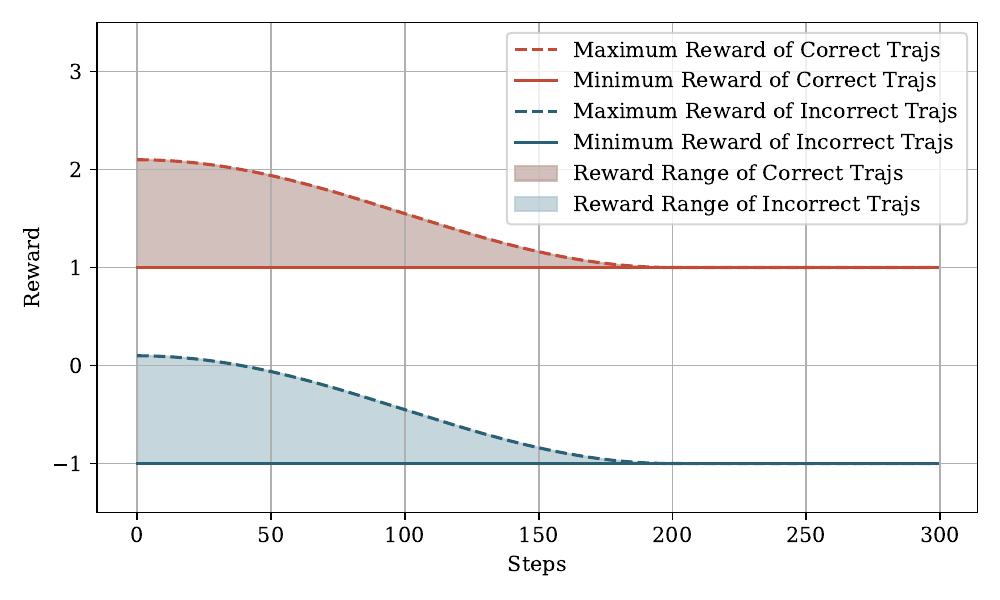}
  \caption{Visualization of the composite intrinsic reward ($T_{\text{decay}}=200$).
The tool-call reward gradually decays from 1 to 0 in the first 200 training steps.}
  \label{fig:reward}
\end{subfigure}%
\end{center}
\caption{Visualization of the curriculum for progressive exploration.
}
\label{fig:gammarewardcurve}
\end{figure}

\paragraph{Self-Imitation.}
The following contents are mentioned in Section~\ref{sec:curriculumsil} in the main text.

The schedule for strengthening SI is defined as below:
\begin{equation}\label{eq:gamma}
\gamma=
\begin{cases}
\frac{1}{2}(1 - \cos(\pi\frac{t_{\text{iter}} }{ T_{\text{warm-up}} })), & t_{\text{iter}}\leq T_{\text{warm-up}},\\
1, & t_{\text{iter}}> T_{\text{warm-up}},
\end{cases}
\end{equation}
\noindent where $t_{\text{iter}}$ and $T_{\text{warm-up}}$ respectively denote the training iteration step and the total warm-up steps.

\paragraph{Intrinsic Reward.}
The following contents are mentioned in Section~\ref{sec:curriculumoutcomereward} in the main text.

The schedule for decaying IR is defined as below:
\begin{equation}\label{eq:mudecay}
    \mu=
\begin{cases}
\frac{1}{2}(\cos(\pi\frac{t_{\text{iter}} }{ T_{\text{decay}} })+1), & t_{\text{iter}}\leq T_{\text{decay}},\\
0, & t_{\text{iter}}> T_{\text{decay}},
\end{cases}
\end{equation}
\noindent where $T_{\text{decay}}$ denotes the number of decaying steps.


\subsection{Reward Definition}
\label{sec:rewarddef}

The following contents correspond to Section~\ref{sec:rewardcomposition} in the main text.

\paragraph{Outcome Reward}
A binary signal is assigned at the end of a episode according to the pre-defined verification rules.
\begin{equation}\label{eq:outcomereward}
    R^i_{\text{outcome}}=\begin{cases}
        1, & \tau_i\ \text{succeeds},\\
        -1, & \text{otherwise}.
    \end{cases}
\end{equation}

\paragraph{Tool-call Reward.}
To incentivize multi-turn interactions,
an action-based reward that is proportional to the number of tool call turns is added.
To avoid reward hacking where the LLM repeats meaningless tool calling,
the action reward is confined smaller than the outcome reward.
\begin{equation}\label{eq:actionreward}
    R^i_{\text{tool-call}}=\min(1, 0.1\cdot n_{\text{tool-call}}), n_{\text{tool-call}}\geq0,
\end{equation}
where $n_{\text{tool-call}}$ denotes the number of valid tool call turns in the trajectory $\tau_i$.

\paragraph{Format Reward.}
A negligible reward is assigned to the trajectory if the model's output contains valid wrapping format given the task descriptions (e.g., \texttt{<think>...</think><action>...</action>}).

\begin{equation}\label{eq:formatreward}
    R^i_{\text{format}}=\begin{cases}
        0.1, & \text{if}\ \mathbf{a}_{t}^{i}\ \text{is wrapped correctly}, \forall (\mathbf{s}_{t}^{i}, \mathbf{a}_{t}^{i}, R_{t}^{i})\in\tau_i \\
        0, & \text{otherwise}.
    \end{cases}
\end{equation}

\subsection{Bag-of-Tricks for \textit{Dr.BoT}}
\label{sec:botdetails}

The following contents correspond to Section~\ref{sec:strong-baseline} in the main text.

\paragraph{Removal of KL Divergence.}
We follow~\cite{yu2025dapo,liu2025understanding} to simply remove the KL divergence by setting $\beta=0$.
This allows the distribution of the LLM to diverge from the initial policy $\pi_0$ for adaptation to tool-integrated reasoning under the agent tasks.

\paragraph{Clip-Higher.}
We follow~\cite{yu2025dapo} to raise the upper clip bound $\epsilon_{\text{ub}}=0.28$ and keep the lower bound $\epsilon_{\text{lb}}=0.2$ as default.
The decoupled lower and higher clipping range leaves more space for the increase of low-probability tokens.
It relaxes the exploration of the policy which benefits premature entropy collapsing.

\paragraph{Removal of Intra-group Normalization.}
We follow~\cite{liu2025understanding} to drop the advantage normalization term where the standard deviations lead to a difficulty bias in optimization.
It has two benefits:
1) The samples with smaller intra-group standard deviations contribute more to the policy update and the removal of normalization allows balancing between samples of various difficulty;
2) The estimation of standard deviations are inaccurate for the off-policy advantage recalibration of replay samples.
It is challenging to measure the sampling diversity of a specific group.

\paragraph{Removal of Length Normalization.}
We follow~\cite{liu2025understanding} to drop the length normalization terms.
We choose the token-level sum and sequence-level normalization as the aggregation approach for both loss computation and the entropy monitoring.

\paragraph{Filtering of Over-long and Void-turn Samples.}
We follow~\cite{bespoke_improving_multi_turn_tool_use,yu2025dapo} to mask out the loss for rollout samples that exceed the predefined maximum response length.
The improper reward shaping for overlong samples introduces noise into training,
which causes instability of training.
Besides,
it prevents from test-time scaling when the context length of evaluation is longer than that of training.
In addition,
we mask out all the trajectories with void turns~\cite{xue2025simpletir},
where the LLM fails to call any tools in the response.
Such void turns are often accompanied with the occurrence of repetitive reasoning contents,
wrong chat-template formatting,
and nonsensical tokens.
The filtering of these void-turn samples prevents mode-collapsing where their distribution deviate severely from the initial policy.

\paragraph{Filtering of Low-variance Groups.}
We follow~\cite{wang2025ragen} to only keep groups with high intra-group variance for each batch of training samples.
The bottom $25\%$ samples with small intra-group reward standard deviations are removed to keep the policy update informative.
High intra-group variance indicates diverse agent behaviors and the contrast between different actions is beneficial to exploitation.

\paragraph{Regularization with Covariance-based Clipping}
We introduce the covariance-based clipping~\cite{cui2025entropy} to the trajectory-level entropy control.
The changes of output logits that are highly associated with advantage gains greatly decrease the entropy.
We remove tokens with high covariances~\cite{cui2025entropy,wang2025beyond} out of loss contribution for $\tilde{\mathcal{J}}_{\text{GRPO}}^{\text{SIL-R}}(\pi_\theta)$,
preventing aggressive changes of log probability for advantage acquisition.
\begin{equation}\label{eq:grpo_sal_reg}
\tilde{\mathcal{J}}_{\text{GRPO}}^{\text{SIL-R}}(\pi_\theta)=
     \mathbb{E}_{\{\tau_{j}\}_{j=1}^{N_{\mathcal{D}}}\sim \{\pi_{\theta_{\text{old}}}(\cdot|x),\ x\sim p(X)\}}\sum_{j=1}^{N_{\mathcal{D}}}\tilde{\mathcal{J}}_{\text{GRPO}}^{j}\cdot\mathbf{1}(\hat{A}_{j}>0\ \&\ \tilde{A}_{j}>0)\cdot M^{j},
\end{equation}

\begin{equation}\label{eq:regularizationmask}
     M^j_{t}=
     \begin{cases}
         0, &t\in I_{\text{clip}}^{j},\\
         1, &t\notin I_{\text{clip}}^{j},
     \end{cases}
\end{equation}

\begin{equation}\label{eq:clipuniform}
    I_{\text{clip}}^{i}=Ind\sim\operatorname{Uniform}(t|\omega_{\text{lb}}\leq\operatorname{Cov}(\log\pi_{\theta}(\mathbf{a}_t^{i}|x,\mathbf{s}_t^{i}), \tilde{A}_{t}^{i})\leq\omega_{\text{ub}}, N^{i}_{\text{clip}}),
\end{equation}

\begin{equation}\label{eq:covariance}
\operatorname{Cov}(\log\pi_{\theta}(\mathbf{a}_t^{i}|x,\mathbf{s}_t^{i}), \tilde{A}_{t}^{i})=(\log\pi_\theta(\mathbf{a}_t^{i}|x,\mathbf{s}_t^{i})-\frac{1}{G}\sum_{j=1}^{G}\log\pi_\theta(\mathbf{a}_t^{j}|x,\mathbf{s}_t^{j}))\cdot(\tilde{A}_{t}^{i}-\frac{1}{G}\sum_{j=1}^{G}\tilde{A}_{t}^{j}),
\end{equation}
\noindent where the lower bound and upper bound for determining the range of high-covariance tokens are respectively represented as $\omega_{\text{lb}}$ and $\omega_{\text{ub}}$.
The operation $\operatorname{Uniform}(t|\cdot, N_{\text{clip}})$ refers to the uniform sampling of tokens $t$ with high covariance until a budget of $N_{\text{clip}}$ tokens.
The indices of the selected tokens for loss masking are represented as $Ind$.
It is noted that such masking introduces randomness which benefits the convergence of RL.
The detailed settings of $\omega_{\text{lb}}$, $\omega_{\text{ub}}$, and $N_{\text{clip}}$ are subject to both the LLM and the task.
We empirically set the rounded integers of the mean covariance in the range of top $20\%$ and top $0.02\%$ respectively for $\omega_{\text{lb}}$ and $\omega_{\text{ub}}$,
and set $N^{i}_{\text{clip}}=\lambda N^{i}$ with $N^{i}$ being the total number of learnable tokens of $\tau_i$ and $\lambda$ denoting the clipping ratio.

{
\subsection{Theoretical Justification}
\label{sec:theoretical}

\paragraph{Claim 1.}\textit{The self-imitation, with a warm-up schedule coefficient $\gamma(t_{\text{iter}})$ that increases from 0 to 1 (Eq.~\ref{eq:grpo_sal}), implements a constrained projection onto the distribution of good responses, ensuring monotonic improvement of the surrogate objective.}

\paragraph{Theorem 1 (Surrogate Objective Improvement Bound).} Let $\pi_{\theta_{t_{\text{iter}}}}$ be the policy at iteration $t_{\text{iter}}$, $\gamma(t_{\text{iter}}) \in [0,1]$ the warm-up coefficient, and $r(\mathbf{a}) = \frac{\pi_{\theta_{t_{\text{iter}}+1}}(\mathbf{a})}{\pi_{\theta_{t_{\text{iter}}}}(\mathbf{a})}$ the importance weight ratio with its clipped surrogate $\tilde{r}(\mathbf{a}) = \text{clip}(r(\mathbf{a}), 1-\epsilon, 1+\epsilon)$. We define the good experiences for group sample $j$ as $I_j=\mathbf{1}(\hat{A}_j > 0\ \&\ \tilde{A}_j > 0)$, where $\hat{A}_j$ and $\tilde{A}_j$ are the estimated and baseline-corrected advantages. Under the assumptions that: (1) the policy change is bounded by the clipping range, and (2) the advantage estimates are unbiased, the surrogate objective improvement satisfies:
\begin{equation}
    \mathcal{J}(\pi_{\theta_{t_{\text{iter}}+1}}) - \mathcal{J}(\pi_{\theta_{t_{\text{iter}}}}) \geq \underbrace{\mathbb{E}_{\mathbf{a} \sim \pi_{\theta_{t_{\text{iter}}}}}\left[\tilde{r}(\mathbf{a}) \cdot A_{\pi_{\theta_{t_{\text{iter}}}}}(\mathbf{a})\right]}_{\text{GRPO improvement}} + \gamma(t_{\text{iter}}) \cdot \underbrace{\mathbb{E}_{j \sim \mathcal{D}}\left[I_j \cdot \log r(\mathbf{a}_j)\right]}_{\text{SIL improvement}}-\epsilon R_{\max},
\end{equation}
where $R_{\max}$ is the maximum per-token reward, and $\mathcal{J}$ denotes the surrogate objective function.

\paragraph{Proof 1.} Consider the combined objective (Eq.~\ref{eq:grpo_sal}), we can decompose the total improvement by linearity:
\begin{equation}
    \Delta \mathcal{J}_{\text{total}} = \mathcal{J}(\pi_{\theta_{t_{\text{iter}}+1}}) - \mathcal{J}(\pi_{\theta_{t_{\text{iter}}}}) = \Delta \mathcal{J}_{\text{GRPO}} + \gamma(t_{\text{iter}}) \cdot \Delta \tilde{\mathcal{J}}_{\text{GRPO}}^{\text{SIL-R}}.
\end{equation}

The GRPO component has a lower bound from the clipped surrogate theorem~\cite{schulman2017proximal}:
\begin{equation}
    \Delta \mathcal{J}_{\text{GRPO}} \geq \mathbb{E}_{\mathbf{a} \sim \pi_{\theta_{t_{\text{iter}}}}}\left[\tilde{r}(\mathbf{a}) \cdot A_{\pi_{\theta_{t_{\text{iter}}}}}(\mathbf{a})\right] - \epsilon R_{\max}.
\end{equation}

For the self-imitation term, under the assumption of small policy changes ($\|\theta_{t+1} - \theta_t\|$ bounded), we approximate the finite difference via gradient integration:
\begin{equation}
    \nabla_\theta \tilde{\mathcal{J}}_{\text{SIL}}^{\text{GRPO}} = \mathbb{E}_{j \sim \mathcal{D}}\left[I_j \cdot \frac{\nabla_\theta \pi_{\theta}(\mathbf{a}_j)}{\pi_{\theta}(\mathbf{a}_j)}\right].
\end{equation}

Using the mean value theorem and assuming smoothness of the objective, we integrate from $\theta_{t_{\text{iter}}}$ to $\theta_{t_{\text{iter}}+1}$:
\begin{equation}
    \Delta \tilde{\mathcal{J}}_{\text{SIL}}^{\text{GRPO}} \approx \mathbb{E}_{j \sim \mathcal{D}}\left[I_j \cdot \log \frac{\pi_{\theta_{t_{\text{iter}}+1}}(\mathbf{a}_j)}{\pi_{\theta_{t_{\text{iter}}}}(\mathbf{a}_j)}\right] = \mathbb{E}_{j \sim \mathcal{D}}\left[I_j \cdot \log r(\mathbf{a}_j)\right].
\end{equation}

The coefficient $\gamma(t_{\text{iter}})$ scales the SIL contribution gradually. Combining terms yields the final bound:
\begin{equation}
    \Delta \mathcal{J}_{\text{total}} \geq \mathbb{E}_{\mathbf{a} \sim \pi_{\theta_{t_{\text{iter}}}}}\left[\tilde{r}(\mathbf{a}) \cdot A_{\pi_{\theta_{t_{\text{iter}}}}}(\mathbf{a})\right] + \gamma(t_{\text{iter}}) \cdot \mathbb{E}_{j \sim \mathcal{D}}\left[I_j \cdot \log r(\mathbf{a}_j)\right] - \epsilon R_{\max}.
\end{equation}

Under trust region constraints, improvements in the surrogate objective $\mathcal{J}$ translate to improvements in expected return $J$~\cite{schulman2017proximal}.

\paragraph{Claim 2.}\textit{The choice of median ($P_{50}$) as the baseline estimator is grounded in robust statistics and variance minimization in agentic RL with heavy-tailed return distributions.}

\paragraph{Theorem 2 (Robustness to Outliers).} Let $\mathcal{R} = \{R_1, R_2, ..., R_n\}$ be a set of returns in baseline buffer $\mathcal{D}_R$. The median $P_{50}$ minimizes the expected absolute deviation and has a bounded influence function, making it robust to outliers compared to the mean.

\paragraph{Proof 2.} For any estimator $b$, the loss minimization objectives are:
\begin{itemize}
    \item Mean: $\underset{b}{\operatorname{argmin}} \sum_{i=1}^n (R_i - b)^2 \implies b = \frac{1}{n}\sum_i R_i$
    \item Median: $\underset{b}{\operatorname{argmin}} \sum_{i=1}^n |R_i - b| \implies b = P_{50}(\mathcal{R})$
\end{itemize}

The influence functions characterize robustness~\cite{huber2011robust}:
\begin{itemize}
    \item Mean: $\text{IF}(R; \text{mean}) = R - \mathbb{E}[R]$ (unbounded)
    \item Median: $\text{IF}(R; \text{median}) = \frac{\operatorname{sgn}(R - P_{50})}{2f(P_{50})}$ (bounded when $f(P_{50}) > 0$)
\end{itemize}

Thus, the median is robust to outliers while the mean is sensitive. This property extends to advantage estimation since advantages are linear functions of returns.

\paragraph{Claim 3.}\textit{The $P_{50}$ achieves a balance between robustness and informativeness. Comparatively, the $P_{25}$ and $P_{75}$ percentiles are either overly conservative or aggressive during advantage-based replay filtering.}

\paragraph{Theorem 3 (Minimax Risk).} For the class $\mathcal{P}$ of symmetric unimodal distributions, the median minimizes the minimax risk for absolute error loss among translation-equivariant estimators:
\begin{equation}
    \inf_{\hat{b}} \sup_{p \in \mathcal{P}} \mathbb{E}[|\hat{b} - \mu(p)|] = \sup_{p \in \mathcal{P}} \mathbb{E}[|P_{50}(X) - \mu(p)|]
\end{equation}

\paragraph{Proof 3.} This is a standard result in robust statistics~\cite{law1986robust, huber2011robust}. For symmetric unimodal distributions, the median is minimax for absolute deviation loss among translation-equivariant estimators.

\paragraph{Claim 4.}\textit{The dual filtering mechanism using both historical advantage $\hat{A}_j$ and recalibrated advantage $\tilde{A}_j$ ensures robust policy updates and leads to better convergence properties.}

\paragraph{Theorem 4 (Dual Filtering).} The combined condition $\hat{A}_j > 0$ and $\tilde{A}_j > 0$ in the SIL objective reduces the variance of gradient estimates and promotes stable policy improvement.

\paragraph{Proof 4.} The dual filtering mechanism provides two benefits:

1. \textbf{Variance Reduction:} By filtering trajectories that were both historically good ($\hat{A}_j > 0$) and remain valuable under the current policy ($\tilde{A}_j > 0$), we focus on a higher-quality subset of experiences. This reduces the effective sample size but increases the signal-to-noise ratio, potentially lowering gradient variance.

2. \textbf{Stability:} The exponential decay in the probability of reusing old trajectories (Eq. 39) prevents over-reliance on outdated experiences. Under appropriate importance weighting and assuming the advantages are estimated correctly, the policy improvement follows the standard off-policy policy gradient theorem~\cite{degris2012off}.

The combined filtering ensures that policy updates are based on relevant, high-quality experiences, promoting monotonic improvement under trust region constraints.

}

\subsection{Detailed Datasets and Environments}
\label{sec:datasetdetails}

The following contents correspond to Section~\ref{sec:dataset} in the main text.

\textit{ALFWorld} is an interactive environment created to evaluate how well LLM agents can handle multi-step decision-making tasks. In each scenario, the agent is given a textual goal and must achieve it by engaging in multiple rounds of interaction with the environment. The platform offers 4,639 task examples spanning six typical household activity categories: Pick \& Place (Pick), Examine in Light (Look), Clean \& Place (Clean), Heat \& Place (Heat), Cool \& Place (Cool), and Pick Two \& Place (Pick2).

\textit{WebShop}, on the other hand, is a sophisticated web-based platform aimed at assessing LLM agents in authentic online shopping situations. Agents are required to interact with a simulated HTML shopping site to search for products, browse items, and purchase an appropriate product. WebShop supports a broad and varied action space, featuring more than 1.1 million products and 12K user instructions.

\textit{DAPO-MATH-17K} is a rigorously engineered, competition-grade benchmark designed to stress-test large-scale RL on LLM agents.
The agent must develop multi-step mathematical reasoning, perform strategic tool-calling for code verification, and reflect on feedback from the sandbox before submitting its final answer.
It contains 17K manually-curated prompts sourced from olympiad-level problems, each transformed so that every ground-truth label is an integer—eliminating symbolic-parsing noise and yielding a clean, deterministic reward signal.

For ALFWorld, we report the average success rate for each subtask as well as the overall results.
For WebShop, we report the average score and the success rate (SR).

\subsection{Detailed Baselines}
\label{sec:detailsbaseline}

The following contents correspond to Section~\ref{sec:baseline} in the main text.

\paragraph{ALFWorld and WebShop.}
We compare with baselines such as prompting-based method (i.e., direct I/O) for the proprietary models GPT-4o~\cite{achiam2023gpt} and Gemini~\cite{team2023gemini},
framework-based method such ReAct~\cite{yao2023react} and Reflexion~\cite{shinn2023reflexion},
RL methods including PPO~\cite{schulman2017proximal}, RLOO~\cite{kool2019buy,ahmadian2024back}, GRPO~\cite{shao2024deepseekmath,guo2025deepseek}, GiGPO~\cite{feng2025group}, and our proposed strong baseline Dr.BoT.

\paragraph{DAPO-MATH-17K.}
We compare with baselines including domain-specific experts (e.g., Qwen2.5-Math~\cite{yang2024qwen2}), existing reasoning models (e.g., Sky-T1~\cite{team2025sky}, o1~\cite{jaech2024openai}, DeepSeek-distilled Qwen 32B~\cite{guo2025deepseek}, QwQ~\cite{team2025qwq}, and s1~\cite{muennighoff2025s1}),
and the tool-integrated RL counterparts (e.g., ReTool~\cite{feng2025retool}, SimpleTIR~\cite{xue2025simpletir}, ZeroTIR~\cite{mai2025agent}, and AFM~\cite{li2025chain}).

\subsection{Implementation Details}
\label{sec:implementation_detail_appdx}

The following contents correspond to Section~\ref{sec:implementation} in the main text.

For ALFWorld and WebShop,
we follow~\cite{feng2025group} to use Qwen2.5-1.5B-Instruct and Qwen2.5-7B-Instruct~\cite{yang2024qwen2} as our base models.
For DAPO-MATH-17K,
we follow~\cite{feng2025retool} to use Qwen2.5-32B-Instruct~\cite{yang2024qwen2} for fair comparison.
In addition,
we use the latest Qwen3-32B-Instruct~\cite{yang2025qwen3} for generalization studies.

The implementation of the present study is based on VeRL~\cite{sheng2024hybridflow} and its extension VeRL-Agent~\cite{feng2025group}.
We use the vLLM~\cite{kwon2023efficient} as the inference engine during online rollout generation.

\begin{table}[htbp]
\centering
\caption{Descriptions of the hyper-parameters for training and inference.}
\label{tab:hyperparams_definition}
\resizebox{\linewidth}{!}{
\setlength{\tabcolsep}{1mm}{
\fontsize{9pt}{10pt}\selectfont{
\begin{tabular}{ll}
\toprule
\textbf{Config}  &  \multicolumn{1}{c}{\textbf{Explanation}} \\
\midrule
train\_batch\_size & The batch size for training \\
val\_data\_size & The batch size for validation \\
ppo\_mini\_batch\_size & The mini batch size for actor update iterations \\
ppo\_max\_token\_len\_per\_gpu & The maximum number of tokens on each GPU  for training \\
ppo\_micro\_batch\_size\_per\_gpu & The micro batch size on each GPU for training \\
log\_prob\_max\_token\_len\_per\_gpu & The maximum number of tokens on each GPU for log-probability  \\
log\_prob\_micro\_batch\_size\_per\_gpu &  The micro batch size on each GPU for log-probability \\
use\_dynamic\_bsz & Whether to use dynamic batch size for load balance  \\
ulysses\_sequence\_parallel\_size & The sequence parallel size for training efficiency \\
tensor\_model\_parallel\_size  & The tensor parallel size of model deployment for rollout generation \\
temperature &  The temperature for decoding in LLM generation \\
top\_p &  The top-p for decoding in LLM generation \\
n\_samples\_per\_prompt &  The number of generated samples per prompt \\
actor\_learning\_rate  &  The learning rate of the actor \\
max\_epochs & The maximum number of epochs \\
num\_steps  &  The number of steps \\
$T_{\text{warm-up}}$ &  The number of steps \\
$T_{\text{decay}}$ &  The number of steps \\
use\_kl\_in\_reward  &  Whether to use the KL term in reward \\
kl\_coef &  The coefficient for the KL divergence term \\
use\_kl\_loss & Whether to use the KL loss \\
$\beta$ &  The coefficient of the KL loss (i.e., kl\_loss\_coef) \\
max\_prompt\_length &  The maximum length of input prompt \\
max\_response\_length &  The maximum length of output generation \\
multi\_turn\_max\_turns & The maximum number of tool-call turns \\
$\epsilon_{\text{lb}}$  &  The lower bound of the policy ratio clipping (i.e., clip\_ratio\_low) \\
$\epsilon_{\text{ub}}$ & The upper bound of the policy ratio clipping (i.e., clip\_ratio\_high) \\
$N_{\mathcal{D}}$ & The replay buffer size for self-imitation learning \\
$N_{\mathcal{D}_R}$ & The baseline buffer size for storing the intra-group average performance \\
$C$ & The lower bound of the value for dual-clip PPO/GRPO (i.e., clip\_ratio\_c) \\
$\omega_{\text{lb}}$  & The lower bound of the covariance-based clipping \\
$\omega_{\text{ub}}$ & The upper bound of the covariance-based clipping \\
$\lambda$ & The ratio of the covariance-based clipping \\
rollout\_filter\_type & The type of filtering based on intra-group variance\\
rollout\_filter\_ratio & The ratio of filtered group \\
loss\_agg\_mode  & The aggregation technique for loss \\
norm\_adv\_by\_std\_in\_grpo & Whether to drop the advantage normalization  \\
training strategy  &  The strategy of training (e.g., FSDP, megatron) \\
\bottomrule
\end{tabular}
}}}
\end{table}

\begin{table}[htbp]
\centering
\caption{Hyper-parameters of ALFWorld, WebShop, DAPO-MATH, Sokoban, and SearchR1.}
\label{tab:hyperparams_value}
\resizebox{0.9\linewidth}{!}{
\setlength{\tabcolsep}{1mm}{
\fontsize{9pt}{10pt}\selectfont{
\begin{tabular}{lccccc}
\toprule
\textbf{Config} &  \textbf{ALFWorld}  & \textbf{WebShop}  & \textbf{DAPO-MATH} & \textbf{Sokoban} & {\textbf{SearchR1}}   \\
\midrule
train\_batch\_size & 32 & 32 & 128 & 32 & 128 \\
val\_data\_size  & \multicolumn{5}{c}{128}   \\
ppo\_mini\_batch\_size & 1024 & 256 & 32 & 64 & 32 \\
ppo\_max\_token\_len\_per\_gpu & -- & -- & 18432 & -- & -- \\
ppo\_micro\_batch\_size\_per\_gpu & 8 & 4 & -- & 8 & -- \\
log\_prob\_max\_token\_len\_per\_gpu & -- & -- & 73728 & -- & 73728 \\
log\_prob\_micro\_batch\_size\_per\_gpu & 8 & 4 & -- & 8 &  --  \\
use\_dynamic\_bsz & \texttt{False} & \texttt{False} & \texttt{True} & \texttt{False}  &  \texttt{True} \\
ulysses\_sequence\_parallel\_size & -- & -- & 8 & -- &  8  \\
tensor\_model\_parallel\_size  &  2 & 2 & 4  &  2  & 4  \\
temperature &  0.4 & 0.4 & 1.0 &  0.4 & 1.0 \\
top\_p &  1 & 1 & 0.6  & 1 & 0.6  \\
n\_samples\_per\_prompt & 8 & 8 & 16 & 8 & 16 \\
actor\_learning\_rate  &  \multicolumn{5}{c}{1e-6} \\
max\_epochs & 200  &  350  &  1  & 200  & 20 \\
num\_steps  & -- & -- & 300 & --  & 300  \\
$T_{\text{warm-up}}$ &  100 & 200 & 300 &  100 &  300  \\
$T_{\text{decay}}$ &  \multicolumn{5}{c}{200}   \\
use\_kl\_in\_reward  &  \multicolumn{5}{c}{\texttt{False}} \\
kl\_coef &  \multicolumn{5}{c}{0} \\
use\_kl\_loss & \multicolumn{5}{c}{\texttt{False}} \\
$\beta$ &  \multicolumn{5}{c}{0} \\
max\_prompt\_length &  2048  &  4096 & 2048   &  1024  & 2048 \\
max\_response\_length &  512 & 1024 &  16384/30000 & 1024 & 30000 \\
multi\_turn\_max\_turns &   50 & 15 & 8{/15} & 15  & 32 \\
$\epsilon_{\text{lb}}$  &  \multicolumn{5}{c}{0.2} \\
$\epsilon_{\text{ub}}$ &\multicolumn{5}{c}{0.28}  \\
$N_{\mathcal{D}}$ & \multicolumn{5}{c}{2048} \\
$N_{\mathcal{D}_R}$ & \multicolumn{5}{c}{10240} \\
$C$ & \multicolumn{5}{c}{10}  \\
$\omega_{\text{lb}}$  &  2 & 2 & 1  &  2 & 1  \\
$\omega_{\text{ub}}$ & 60 & 60 & 40 & 60  & 40  \\
$\lambda$ & \multicolumn{5}{c}{0.02} \\
rollout\_filter\_type &  \multicolumn{5}{c}{\texttt{std.}} \\
rollout\_filter\_ratio & \multicolumn{5}{c}{0.75} \\
loss\_agg\_mode  &  \multicolumn{5}{c}{seq-mean-token-sum-norm}  \\
norm\_adv\_by\_std\_in\_grpo &  \multicolumn{5}{c}{\texttt{False}}  \\
training strategy  &  \multicolumn{5}{c}{FSDP} \\
\bottomrule
\end{tabular}
}}}
\end{table}

\subsubsection{Hyper-parameters}

We present the details of the hyper-parameter settings in the present study.
Table~\ref{tab:hyperparams_definition} provides the definitions of the hyper-parameters used in the present study.
We follow~\cite{sheng2024hybridflow} to keep most of the default empirical settings unchanged for comparability.
For the covariance-based clipping,
we follow~\cite{cui2025entropy} to set the clipping bounds $\omega_{\text{lb}},\omega_{\text{ub}}$ respectively as the mean value of the top 0.02\% and top 2\% covariance.
It is noted that the token-level covariance differs from task to task.
Therefore,
we perform statistics analysis on the covariance between action probability and the advantage with the initial model at the first training step to determine the clipping bounds.

All the settings of their values can be found in Table~\ref{tab:hyperparams_value}.
Without loss of generalizability,
we do not perform meticulous fine-tuning of the hyper-parameters.
One would expect better performance with grid search for the optimal hyper-parameters.

\subsubsection{Computing Resources}
All experiments are performed on workstations with 380 CPU cores, 2.2TB memory, and 8 GPUs of 96GB memory.
For both 1.5B/7B LLMs and 3B VLMs,
the training is performed on four workstations with 32 GPUs in total.
For the 32B models,
the training is performed on sixteen workstations with 128 GPUs in total.

For ALFWorld, Webshop, and Sokoban,
it takes less than 60 hours for optimization of 1.5B and 7B models.
While for the DAPO-MATH-17K,
it takes around a week for training the 32B models.

\subsection{Discussions on Hyper-parameters}
\label{sec:discusshyper}

The following contents are mentioned in Section~\ref{sec:discussionintotal} in the main text.

\begin{figure}[htbp]
\begin{center}
\begin{subfigure}{.32\textwidth}
  \centering
  \includegraphics[width=\textwidth]{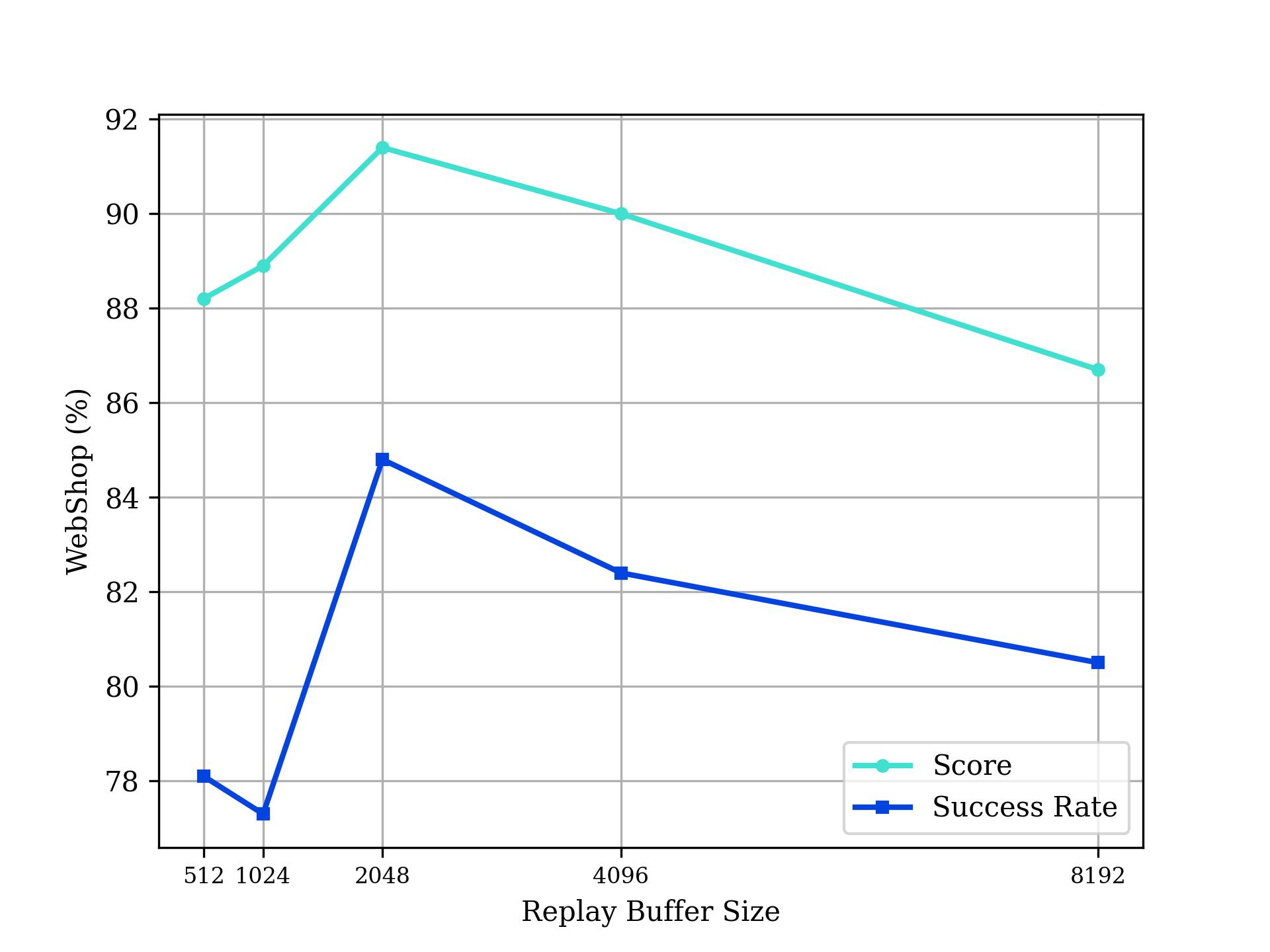}
  \caption{Replay Buffer Size $N_{\mathcal{D}}$.}
\end{subfigure}
\begin{subfigure}{.32\textwidth}
  \centering
  \includegraphics[width=\textwidth]{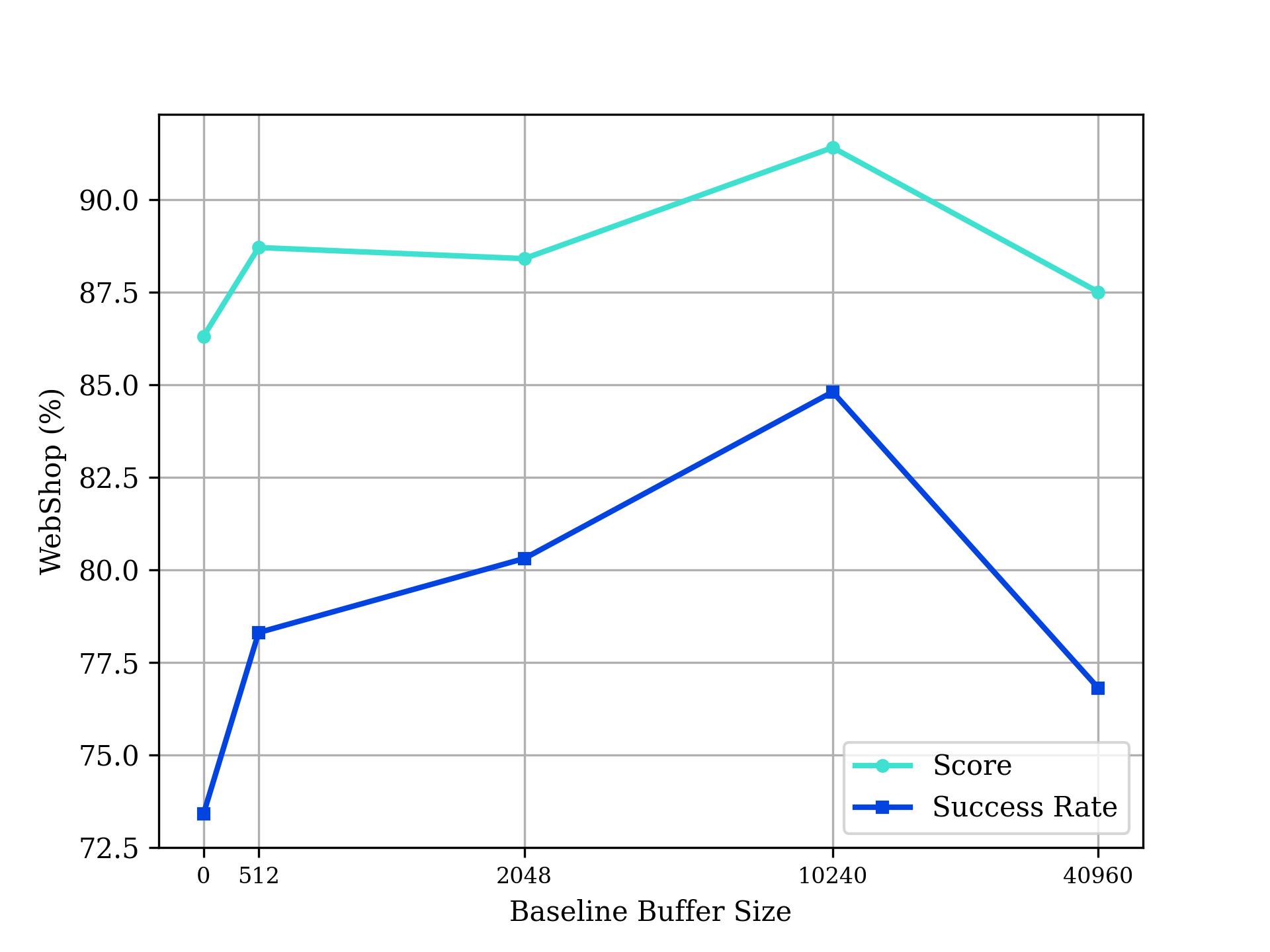}
  \caption{Baseline Buffer Size $N_{\mathcal{D}_R}$.}
\end{subfigure}
\begin{subfigure}{.32\textwidth}
  \centering
  \includegraphics[width=\textwidth]{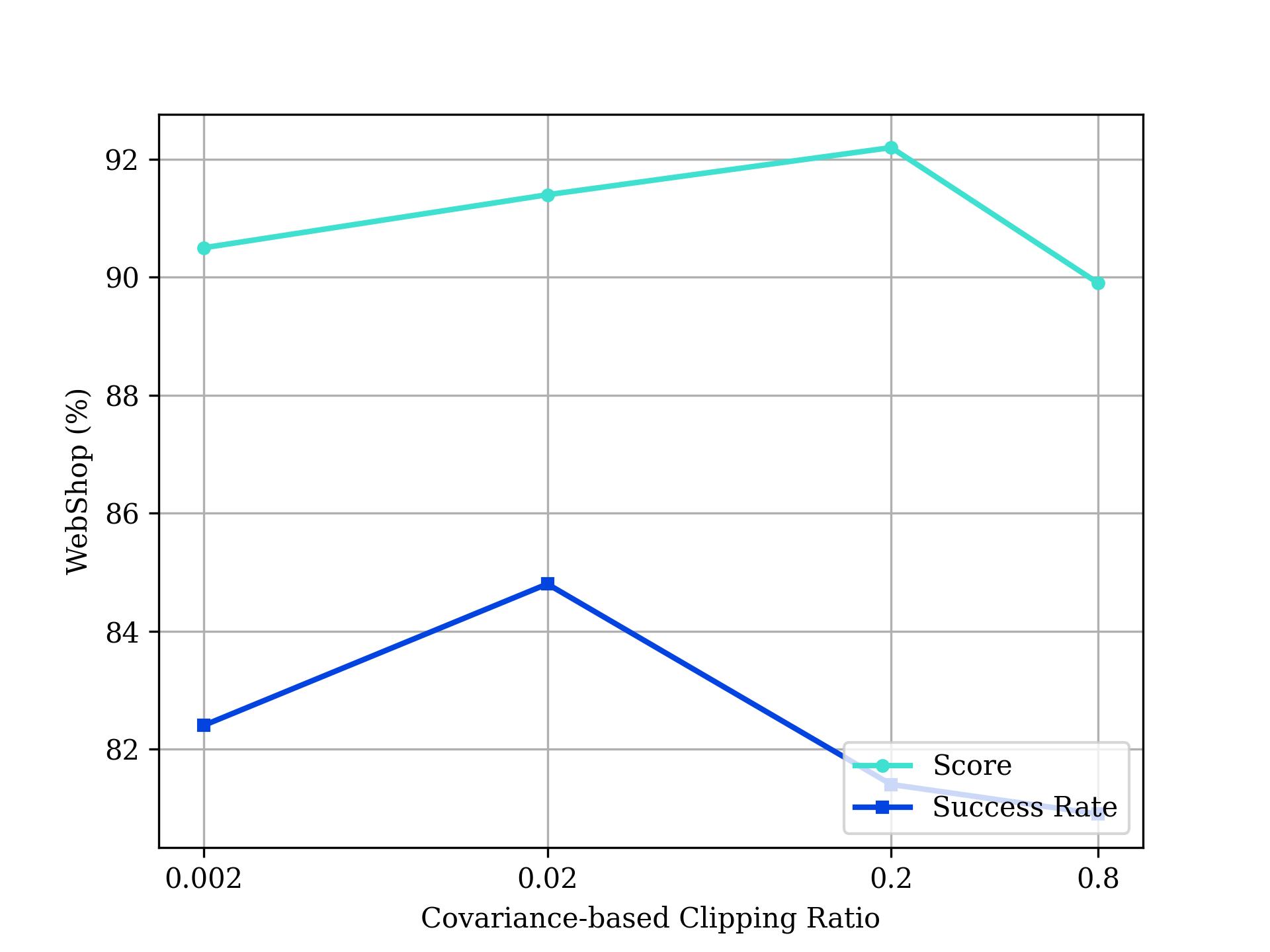}
  \caption{Clipping Ratio $\lambda$.}
\end{subfigure}%
\\
\begin{subfigure}{.32\textwidth}
  \centering
  \includegraphics[width=\textwidth]{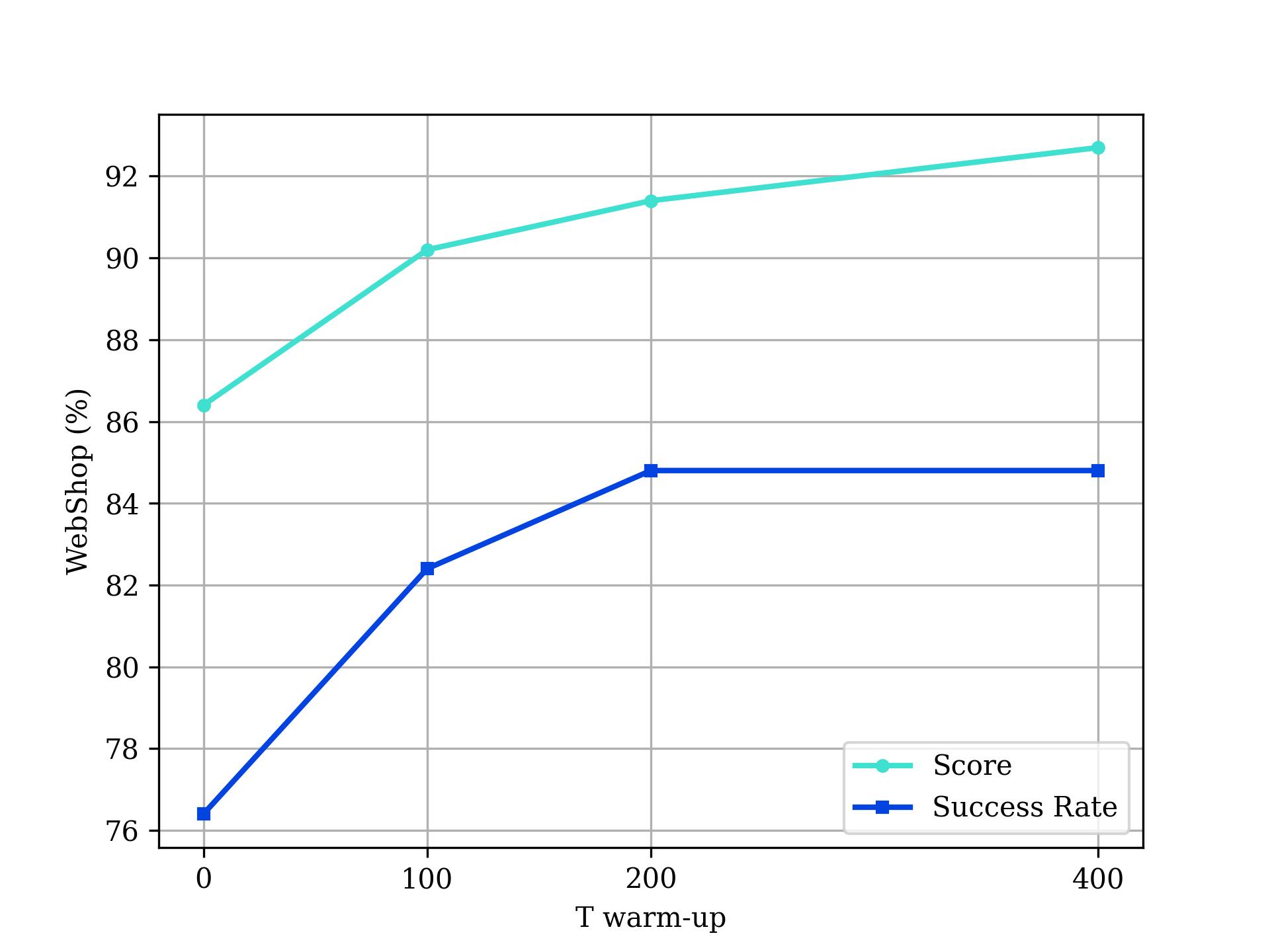}
  \caption{Warm-up Steps $T_{\text{warm-up}}$.}
\end{subfigure}%
\hspace{1em}
\begin{subfigure}{.32\textwidth}
  \centering
  \includegraphics[width=\textwidth]{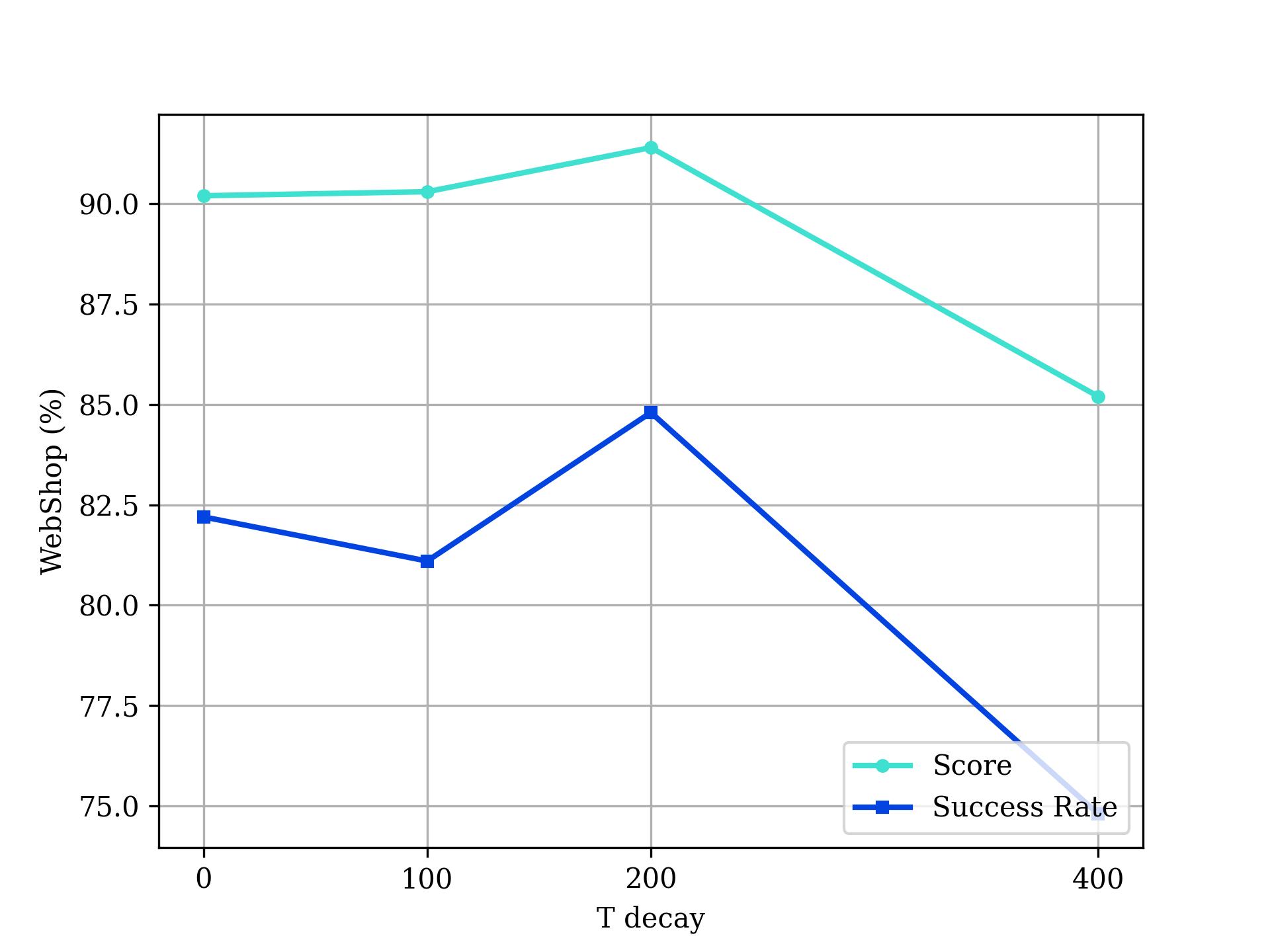}
  \caption{Decay Steps $T_{\text{decay}}$.}
\end{subfigure}%
\end{center}
\caption{Effect of hyper-parameters of \textit{Dr.BoT} (GRPO) with \method on WebShop (Qwen2.5-7B-Instruct).
}
\label{fig:hyperparams}
\end{figure}

{
\subsubsection{Effect of Hyper-parameters}
}

We investigate the following key hyper-parameters (see Figure~\ref{fig:hyperparams}) of \textit{Dr.BoT} (GRPO) with \method on WebShop (Qwen2.5-7B-Instruct) while keeping the value of others fixed (see Table~\ref{tab:hyperparams_value}).

\paragraph{Replay Buffer Size $N_{\mathcal{D}}$.}
As the buffer size increases,
the performance first improves due to the improved diversity and impact of the collected trajectories in the buffer.
However,
when the buffer continues to expand,
trajectories in the buffer might come from earlier training batches and thereafter causes more severe degree of off-policy.
The self-imitation of excessively outdated experiences becomes detrimental to the update of current policy.
In addition,
the large replay buffer takes more iterations to refill and thereafter the policy update frequency from self-imitation is lower than that of a smaller buffer,
further diminishing its intervention in agent exploration.

\paragraph{Baseline Buffer Size $N_{\mathcal{D}_R}$.}
When $N_{\mathcal{D}_R}=0$,
the original advantages are used without recalibration and filtering (see Equation~\ref{eq:silvanilla}).
It shows that the direct imitation of these experiences can be suboptimal where certain trajectories are outdated for the current policy.
By timely adjusting the advantages and removing inappropriate experiences ($\tilde{A}_{j}\leq0$),
we reduce the inaccurate estimation for off-policy update.
{
It is noted that using advantage rather than reward in the baseline buffer helps mitigate learning bias, as it allows for contributions from samples with negative rewards as long as there is variance within a group.
The removal of the standard deviation of outcome rewards is crucial for reducing difficulty bias.
Furthermore, our double-positive advantage gate for replay filtering is essential for off-policy learning.
}
We also find that $N_{\mathcal{D}_R}$ should not be set too large as such 50-th percentile reward deviates from the latest ones,
decreasing the effectiveness of recalibration.

\paragraph{Covariance-based Clipping Ratio $\lambda$.}
The clipping ratio can be viewed as the degree of regularization for policy entropy,
where a larger ratio causes more tokens to be ignored during policy update.
In this case,
the contribution of self-imitation gets weakened.
A modest range of clipping ratio (e.g., $0.0002\sim0.02$) not only suffices the entropy management but also allows proper exploitation of the collected experiences.

\paragraph{Warm-up Step $T_{\text{warm-up}}$.}
A smaller warm-up step implies earlier self-imitation of the premature, suboptimal experiences during RL.
Especially when the distribution of the task and environment differs greatly from the pre-trained knowledge,
the overfitting of the initial trajectories hinders exploration of low-probable solutions and leads to action-level local optima.
Intuitively,
$T_{\text{warm-up}}$ can be first set the same as the total number of training steps and then adjusted according to the task and the model for the improved performance.

\paragraph{Decay Step $T_{\text{decay}}$.}
A smaller decay step reduces the stimulation from the intrinsic reward for acquisition of tool-use skills.
If the LLM already excels at interacting with the environment (e.g., use of tools and comprehension of observations),
$T_{\text{decay}}$ can be set close to 0.
A large $T_{\text{decay}}$ is not encouraged as the interference with the outcome reward causes inconsistent policy optimization for convergence.

{
\subsubsection{Guidelines on Hyper-parameters Tuning}
In this section, we provide guidelines on the choice of these hyper-parameters for practical usage. It is noted that most of the hyper-parameters share the same value settings across benchmarks of various domains and tasks.
One would expect performance gains without meticulous fine-tuning.

\paragraph{Replay Buffer Size $N_{\mathcal{D}}$.}
It should not be set too large to avoid severe off-policy deviation. A modest size of $2K\sim 4K$ proportional to the training batch size of 128 and group size of 16 ($128\times16$) is expected to work well for frequent refilling and policy update.
In other word, $N_{\mathcal{D}}$ cam be set as 2x/4x of $\text{train\_batch\_size}\times\text{n\_samples\_per\_prompt}$.

\paragraph{Baseline Buffer Size $N_{\mathcal{D}_R}$.}
An appropriate setting between 2K and 10K prevents outdated and untimely estimation of current policy baseline performance.
In other word, $N_{\mathcal{D}_{R}}$ can be set as 1x/4x of $N_{\mathcal{D}}$.

\paragraph{Covariance-based Clipping Ratio $\lambda$.}
The percentage of clipped tokens should be controlled between 0.02\% and 2\%. A smaller percentage would reduce the effect of anti-overfitting while a larger percentage slows down the policy exploitation of experiences.

\paragraph{Warm-up Step $T_{\text{warm-up}}$.}
The self-imitation should be scheduled to reach its maximum after 200 steps. For difficult tasks, it should be increased to allow exploration of diverse trajectories without convergence to local sub-optimum.
One could first try $T_{\text{warm-up}}=\text{num\_steps}$.

\paragraph{Decay Step $T_{\text{decay}}$.}
A decay step between 100 and 200 would be sufficient. If the tool is hard to master (e.g., complex slot-filling), the decay step should be increased to allow more stimulation of tool-calling behaviors.
One could first try $T_{\text{decay}}=\text{num\_steps}$.
}

\subsection{Qualitative Analysis}
\label{sec:qualitative}

The following contents are mentioned in Section~\ref{sec:discussionintotal} in the main text.

\begin{figure}[htbp]
\begin{center}
\begin{subfigure}{.5\textwidth}
  \centering
  \includegraphics[width=\textwidth]{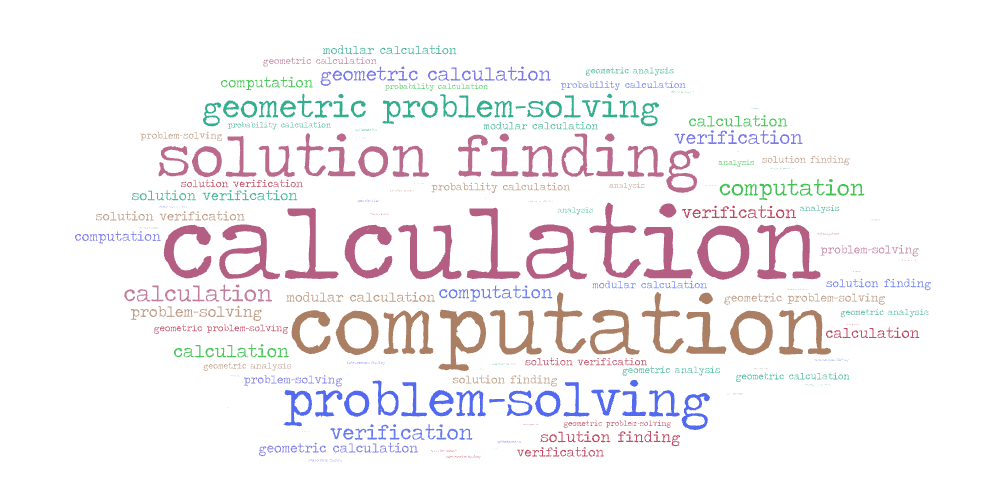}
  \caption{Before RL training.}
\end{subfigure}%
\begin{subfigure}{.5\textwidth}
  \centering
  \includegraphics[width=\textwidth]{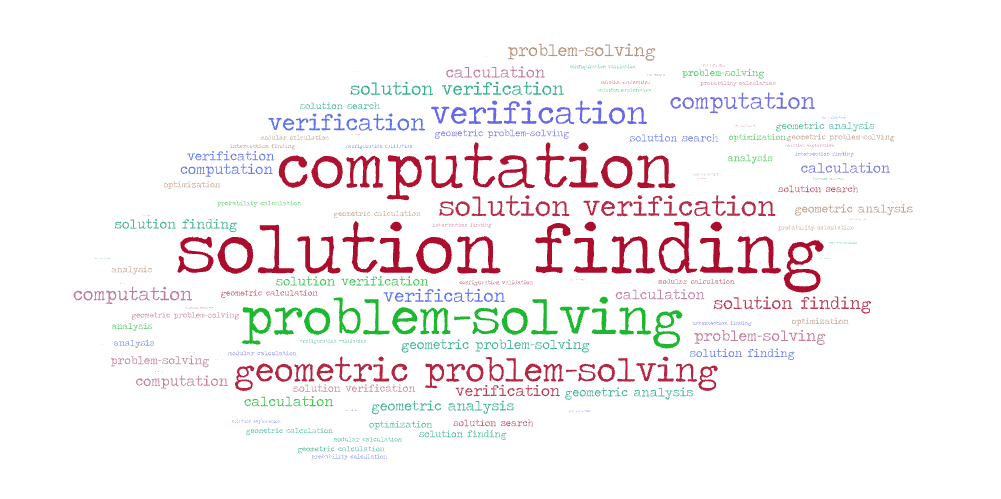}
  \caption{After RL training.}
\end{subfigure}%
\\
\begin{subfigure}{\textwidth}
  \centering
  \includegraphics[width=\textwidth]{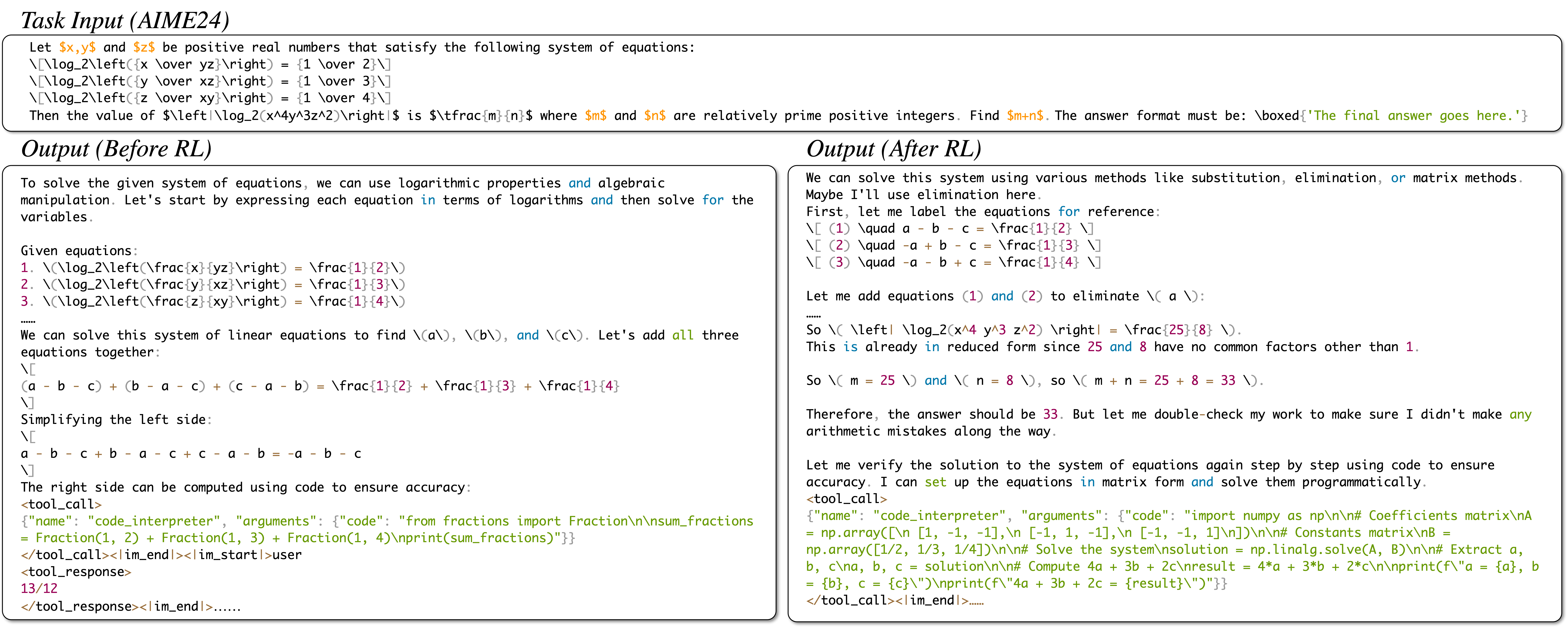}
  \caption{The evolution of efficient coding from the purpose of computation to verification (best viewed magnified).}
\end{subfigure}%
\end{center}
\caption{Development of the agent's coding skills.}
\label{fig:skilldev}
\end{figure}

\subsubsection{Tool-integrated Reasoning}
\paragraph{Skill Development.}
We follow~\cite{feng2025retool} to analyze the coding capabilities of the agent before and after RL by classifying the purpose of the code snippets.
Specifically,
we employ Hunyuan-Large~\cite{sun2024hunyuan} to interpret reasoning contexts before each tool-calling and judge the intention of the codes passed into the code interpreter on DAPO-MATH-17K dataset.
{
The external LLM first performs intent classification with open-ended categories in a free-form manner.
Then, we manually deduplicate these categories and only keep the top 20 frequent ones:
\textit{calculation, computation, solution finding, problem-solving, geometric problem-solving, verification, geometric calculation, solution verification, modular calculation, probability calculation, geometric analysis, analysis, optimization, intersection finding, solution search, function analysis, configuration validation, data computation, game analysis, data processing, game strategy analysis, solution exploration, data analysis, list validation}.
We further use the LLM to classify each code snippet into at most three categories.
The agreement between the LLM and manual classification is above 90\% on 50 randomly chosen samples.
}
The code purposes with their frequency over twice are kept and illustrated in Figure~\ref{fig:skilldev}.
We find that the after RL,
the agent becomes proficient in writing codes for solution finding, problem-solving, and solution verification,
which is quite advanced compared with the basic calculation and computation before RL.
Our case study shows that after training,
the agent learns to master matrix computation with \texttt{numpy} for higher efficiency,
confirming the improvement of coding skills.

\begin{figure}[htbp]
\begin{center}
\centering
\includegraphics[width=\textwidth]{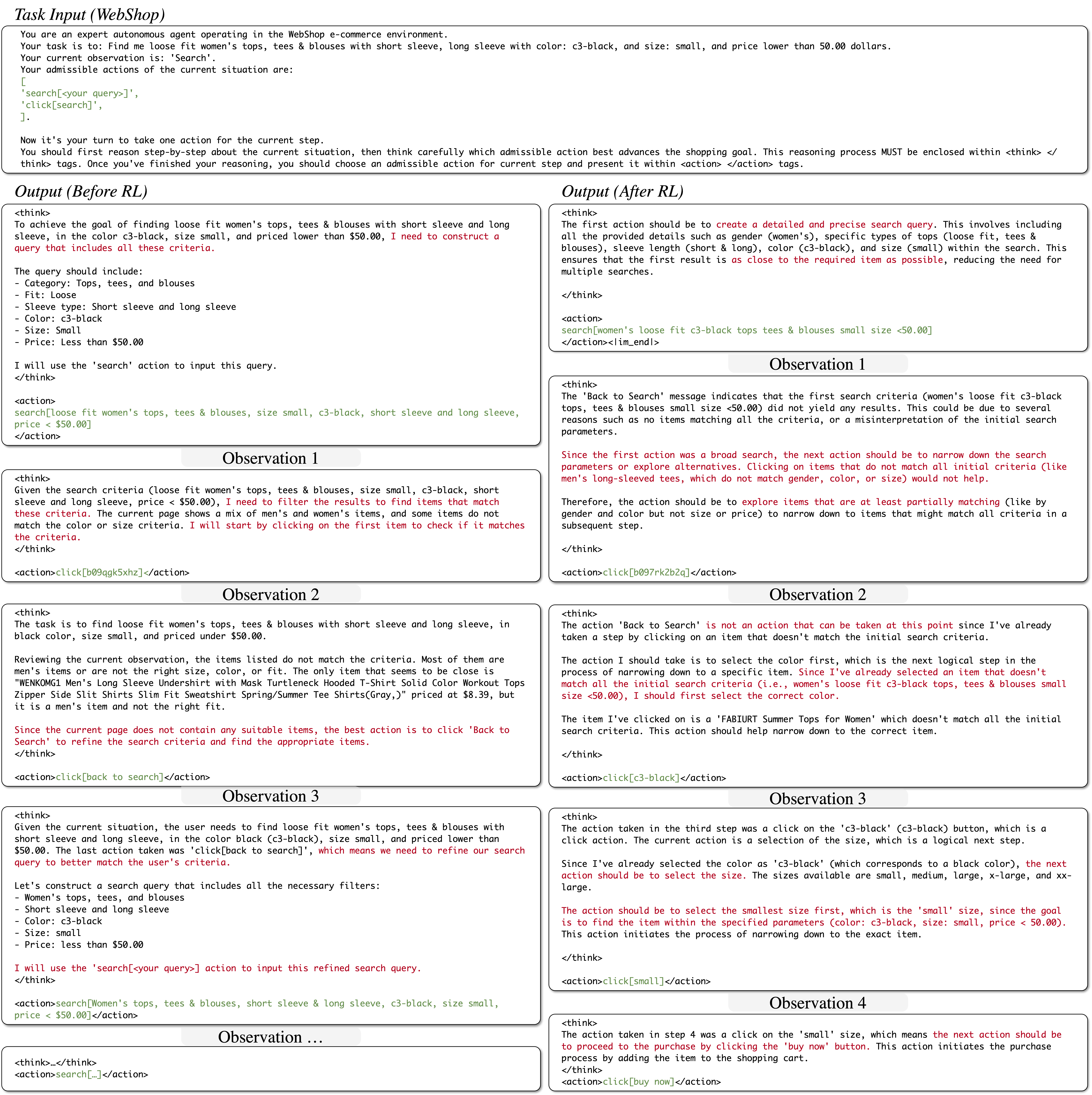}
\end{center}
\caption{The advancement of strategy from the search query perfectionism to goal-oriented active progression (best viewed magnified).}
\label{fig:actionref}
\end{figure}

\paragraph{Action Refinement.}
As shown in Figure~\ref{fig:actionref},
the agent initially aims at finding the target product that satisfies all the constraints simply by searching.
However,
such continuous choice of the action \texttt{search} is trapped by the unqualified retrieval results.
The attributes of product such as color and size should be determined only at the product page.
After RL,
the agent jumps out of the perfectionism for the search queries and tries to break the task step by step.
It learns to choose the action wisely for persistent focus on the task.

{
\subsubsection{Context Budget}

\begin{table}[htbp]
\caption{{The number of tool call turns and response length of SPEAR on Qwen2.5-32B and Qwen3-32B under 16K and 32K context budgets, respectively.}}
\centering
\label{tab:contextbudget}
\resizebox{0.8\linewidth}{!}{
\setlength{\tabcolsep}{1mm}{
\fontsize{9pt}{10pt}\selectfont{
\begin{tabular}{lllll}
\toprule
Model & \# Turns@16K & Responses Len@16K & \# Turns@32K & Responses Len@32K \\
\midrule
Qwen2.5-32B & 7.18 & 4855.48 & 7.13 & 7502.59 \\ 
Qwen3-32B & 3.23 & 10522.38 & 4.43 & 12371.95 \\ 
\bottomrule
\end{tabular}
}}}
\end{table}

In this section,
we provide more analysis on the differences of reasoning behaviors between 16K and 32K token contexts.
Table~\ref{tab:contextbudget} shows that for Qwen2.5 models, the number of tool call turns does not increase abruptly from 16K to 32K.
Two reasons are possible: 1) The tool call reward (Eq.~\ref{eq:actionreward}) allows a maximum of 1 score which corresponds to 10 tool calls. More tool calls (>10) will not be rewarded.
2) The intrinsic reward design is targeted at stimulating exploration at the beginning and the dominance of outcome reward is guaranteed via scheduling.
The mechanical increase of tool use for reward hacking will be penalized to promote reasoning for accuracy.
For Qwen3-32B, the number of tool calls increases a bit but still falls behind that of Qwen2.5-32B.
This is because the Qwen3 series are reasoning models and tend to develop sophisticated solution patterns via pure text.
In this case, the agent mainly uses the tool to double-check its previous textual reasoning and computation.
The context budget from 16K to 32K allows 2K more response tokens and accordingly follows one or two more rounds of tool calls for verification.

\begin{figure}[htbp]
\begin{center}
\centering
\includegraphics[width=\textwidth]{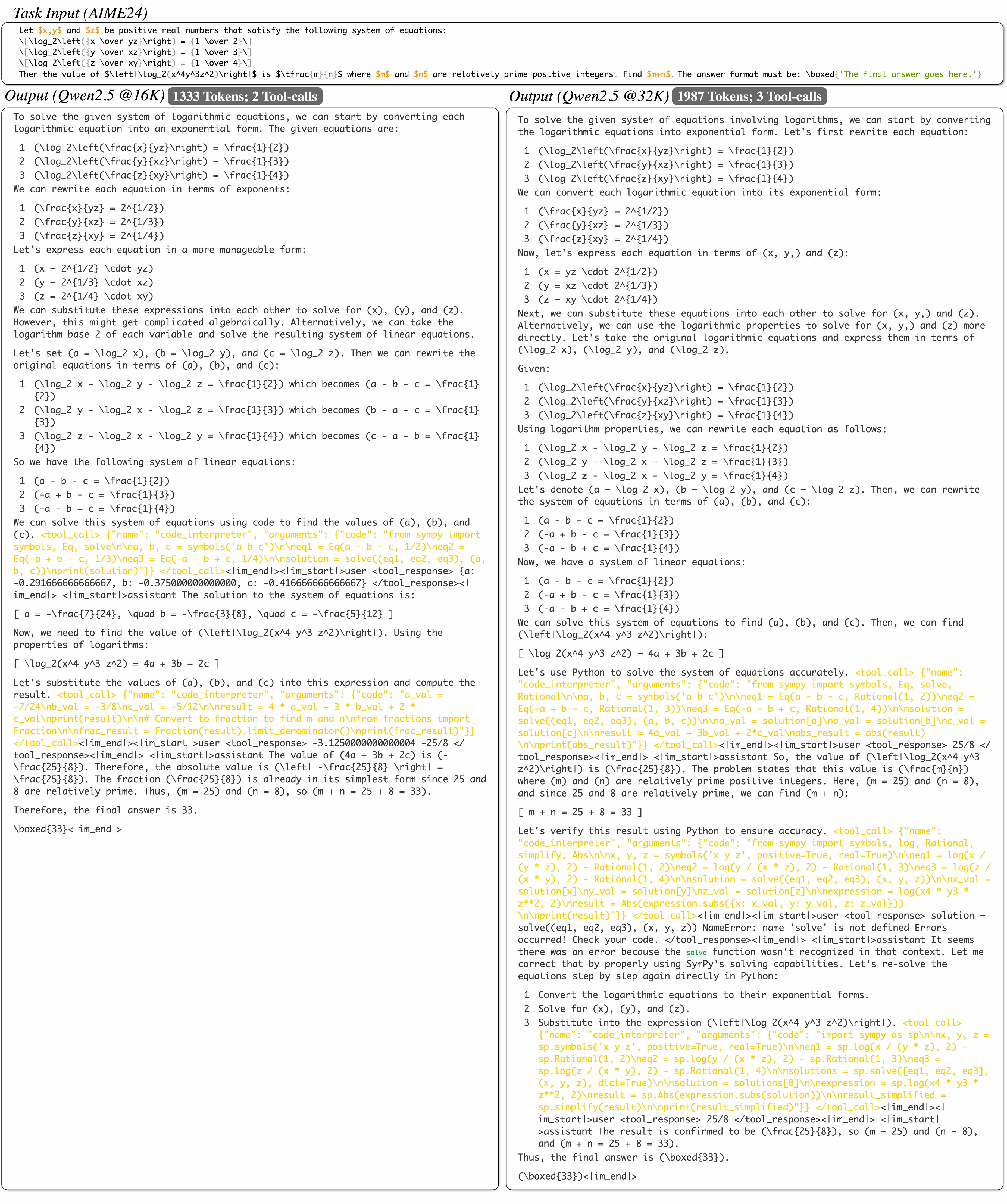}
\end{center}
\caption{{The qualitative results of Qwen2.5 reasoning between 16K and 32K contexts.}}
\label{fig:qualitativeqwen2}
\end{figure}

\begin{figure}[htbp]
\begin{center}
\centering
\includegraphics[width=\textwidth]{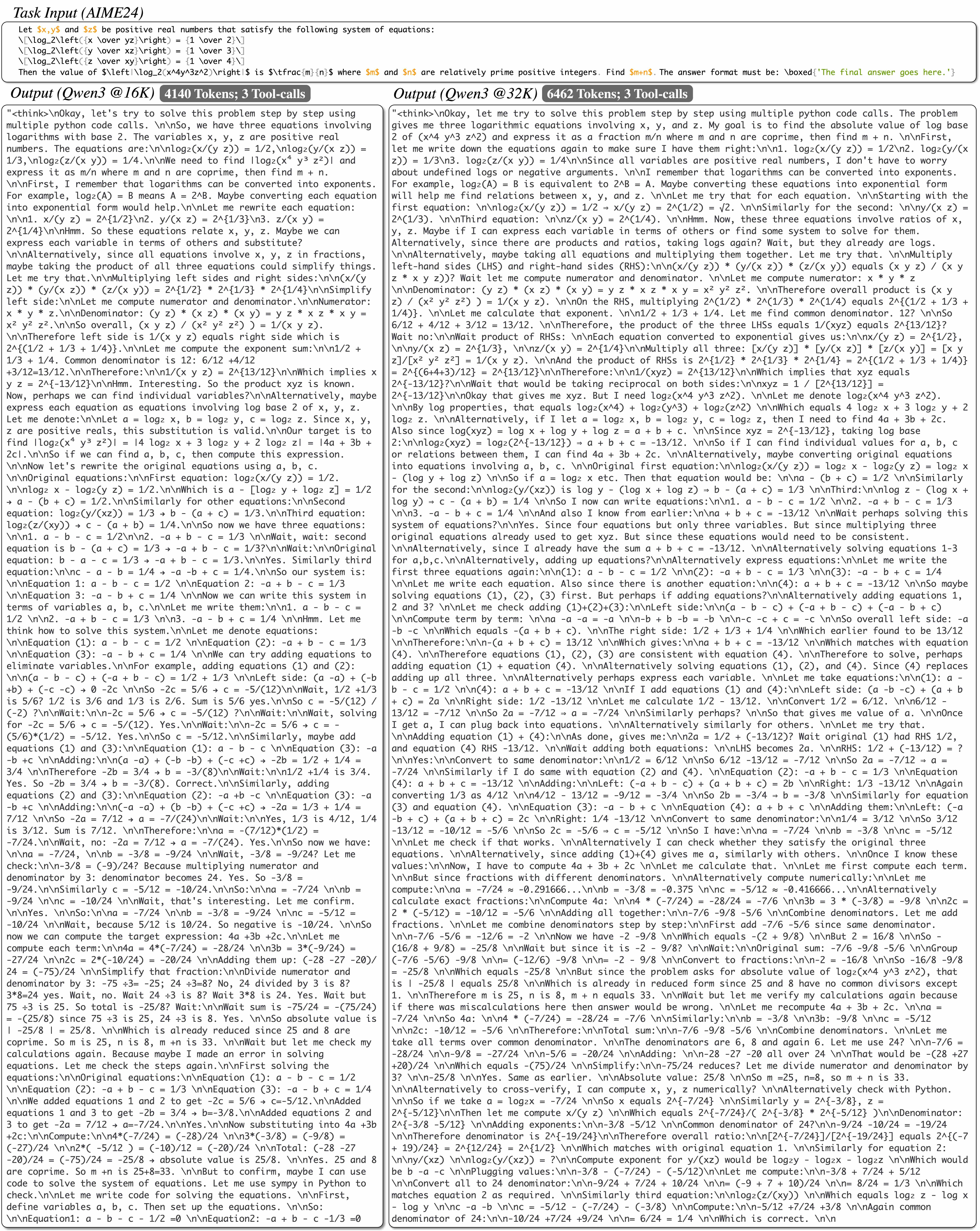}
\end{center}
\caption{{The qualitative results of Qwen3 reasoning between 16K and 32K context (first part).}}
\label{fig:qualitativeqwen3a}
\end{figure}

\begin{figure}[htbp]
\begin{center}
\centering
\includegraphics[width=\textwidth]{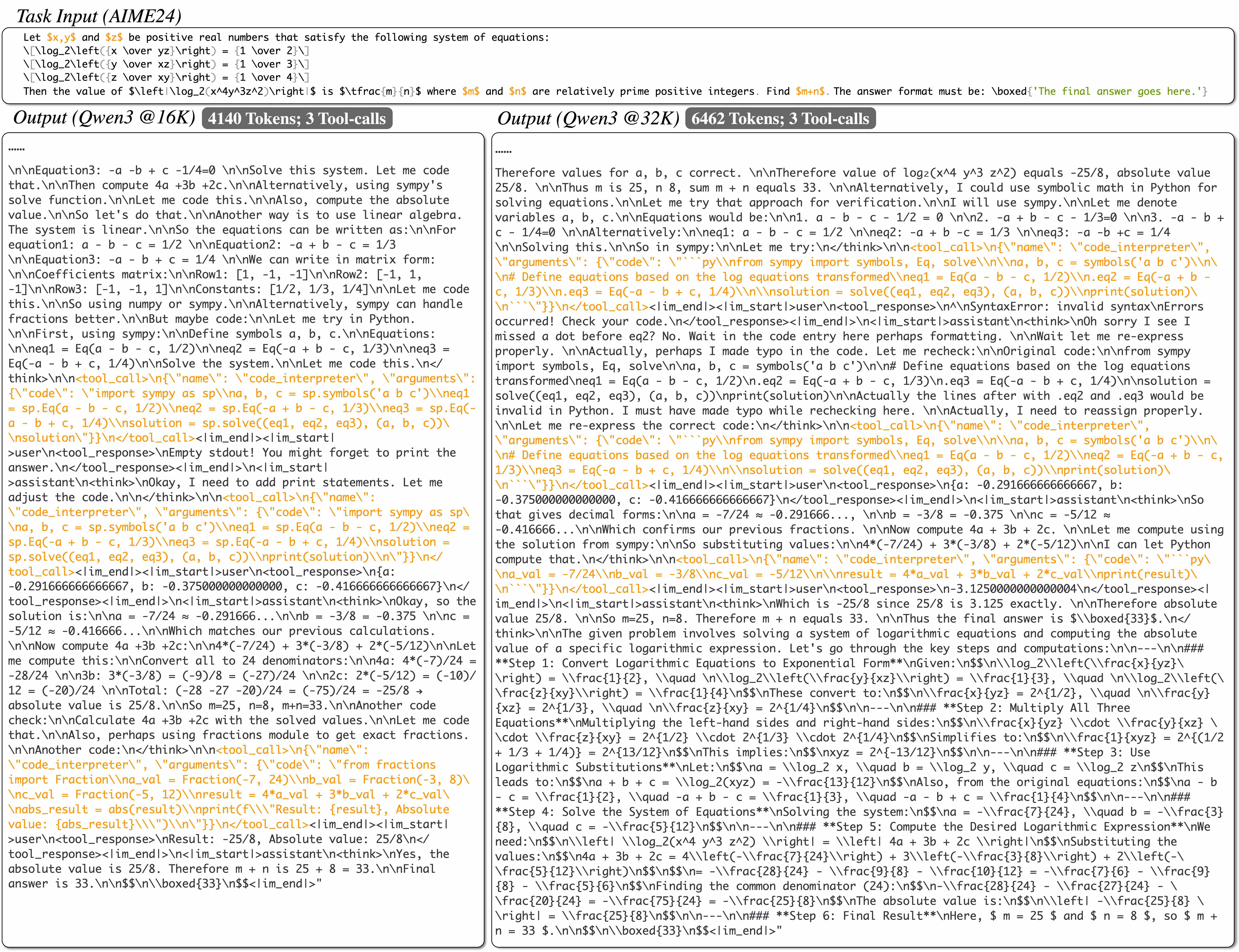}
\end{center}
\caption{{The qualitative results of Qwen3 reasoning between 16K and 32K context (second part).}}
\label{fig:qualitativeqwen3b}
\end{figure}

Examples on the reasoning patterns of Qwen2.5 and Qwen3 under 16K and 32K contexts are respectively provided in Figures~\ref{fig:qualitativeqwen2}~\ref{fig:qualitativeqwen3a}~\ref{fig:qualitativeqwen3b}.
We randomly choose one sample from the AIME 24 benchmark.
It shows that for both Qwen2.5 and Qwen3 models,
the number of tool calls does not increase drastically, which is consistent with the Table~\ref{tab:contextbudget}.
We believe the AIME benchmarks are of reasoning-heavy tasks which challenge the agent the most its complex reasoning capabilities.
In this case,
our SPEAR balances the tool call frequency and the final outcome by:
1) stimulating exploration at an early stage with a bounded tool call reward (maximum of 1),
and 2) guaranteeing dominance of the outcome reward via scheduled adjustment.
Such design prevents the agent from hacking reward simply by frequent tool calling.
Instead, the agent learns to reason deeply in text,
and uses the tool observation as feedback to cross-validate its previous deduction and computation.
The increased context budget allows longer thinking and reflection process,
leading to performance gains.
}

\begin{figure}[htbp]
\begin{center}
\begin{subfigure}{.4\textwidth}
  \centering
  \includegraphics[width=\textwidth]{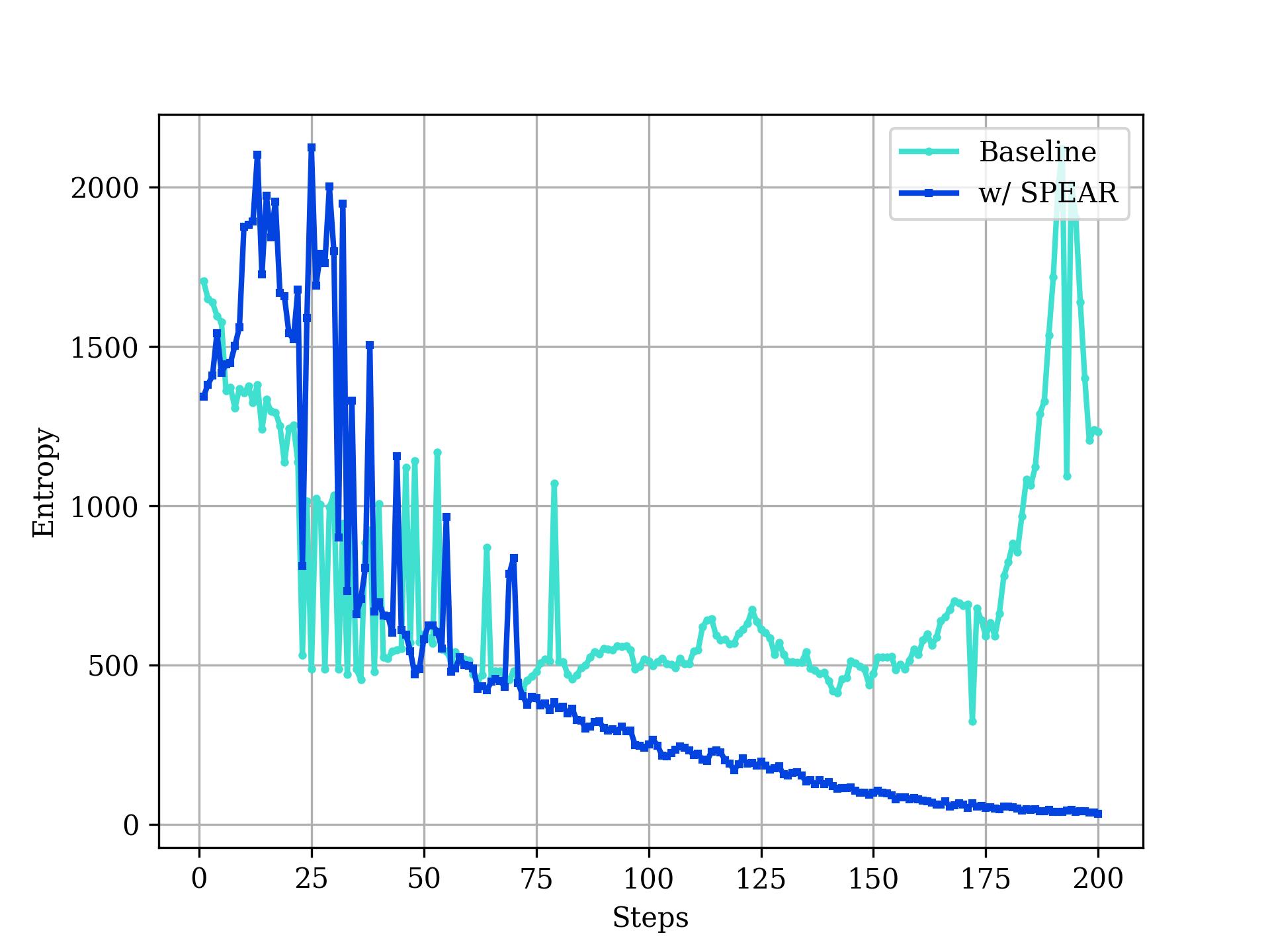}
  \caption{ALFWorld 1.5B \textit{Dr.BoT} (GRPO).}
\end{subfigure}
\begin{subfigure}{.4\textwidth}
  \centering
  \includegraphics[width=\textwidth]{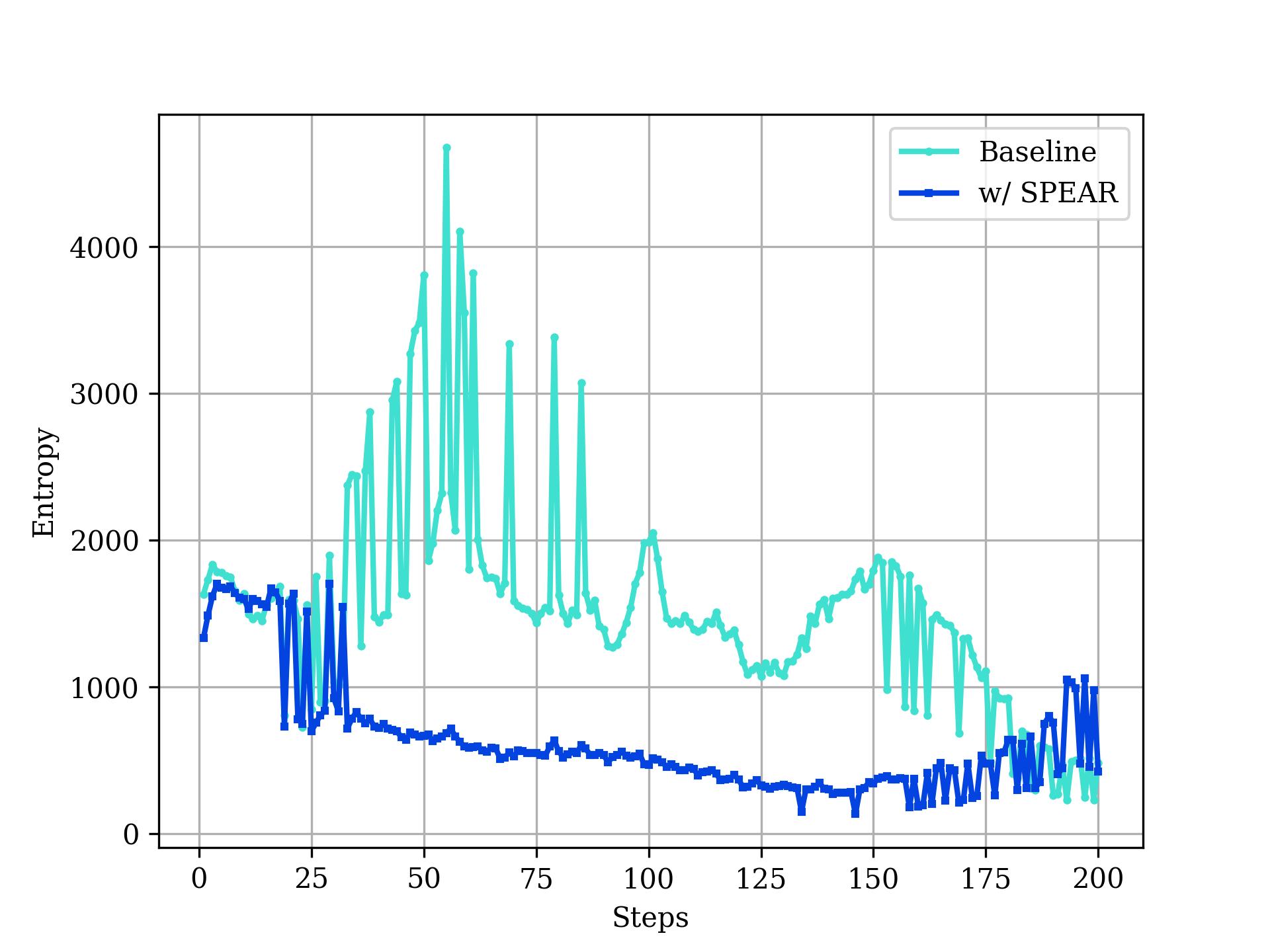}
  \caption{ALFWorld 1.5B \textit{Dr.BoT} (GiGPO).}
\end{subfigure}
\\
\begin{subfigure}{.4\textwidth}
  \centering
  \includegraphics[width=\textwidth]{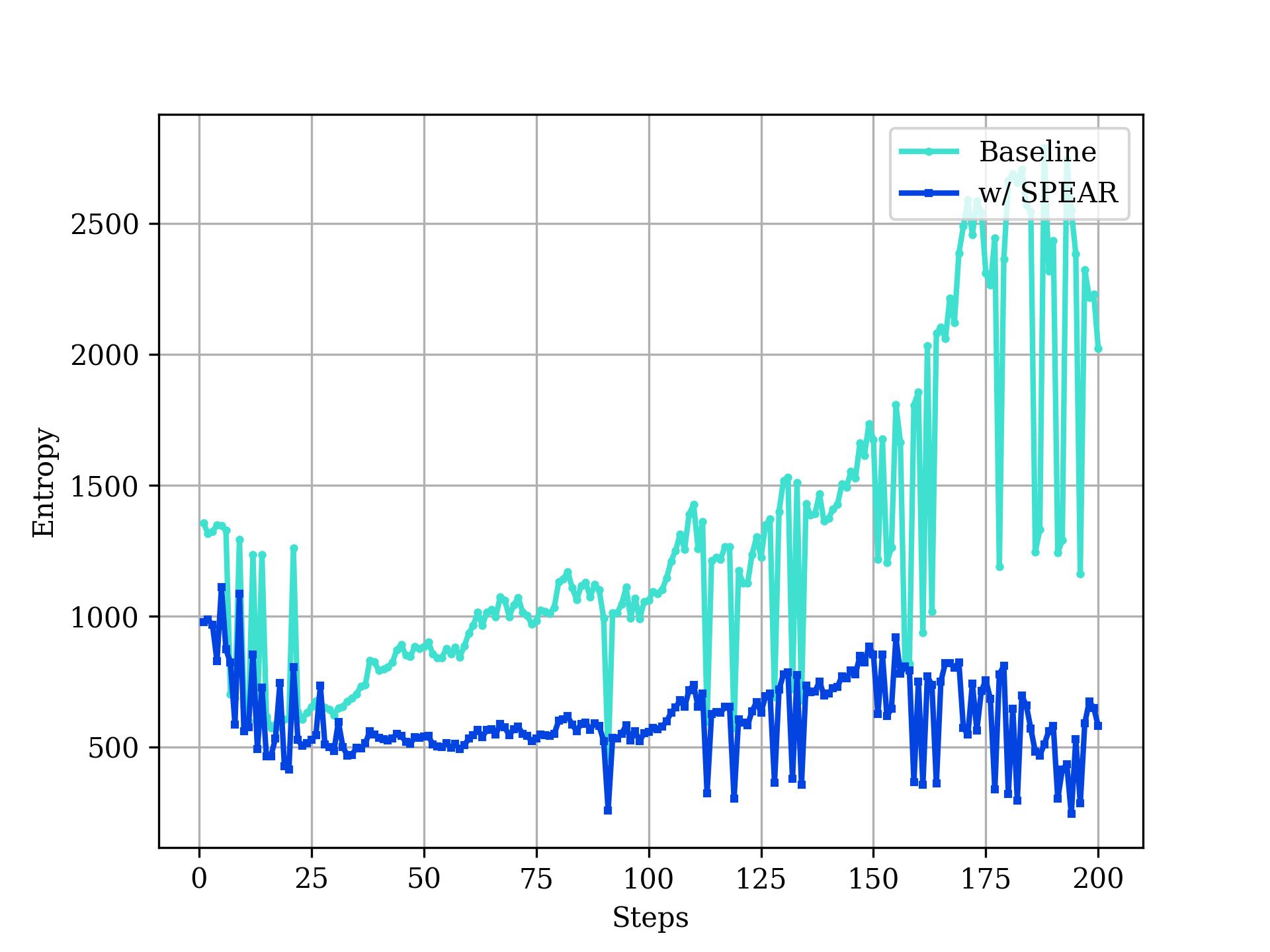}
  \caption{ALFWorld 7B \textit{Dr.BoT} (GRPO).}
\end{subfigure}
\begin{subfigure}{.4\textwidth}
  \centering
  \includegraphics[width=\textwidth]{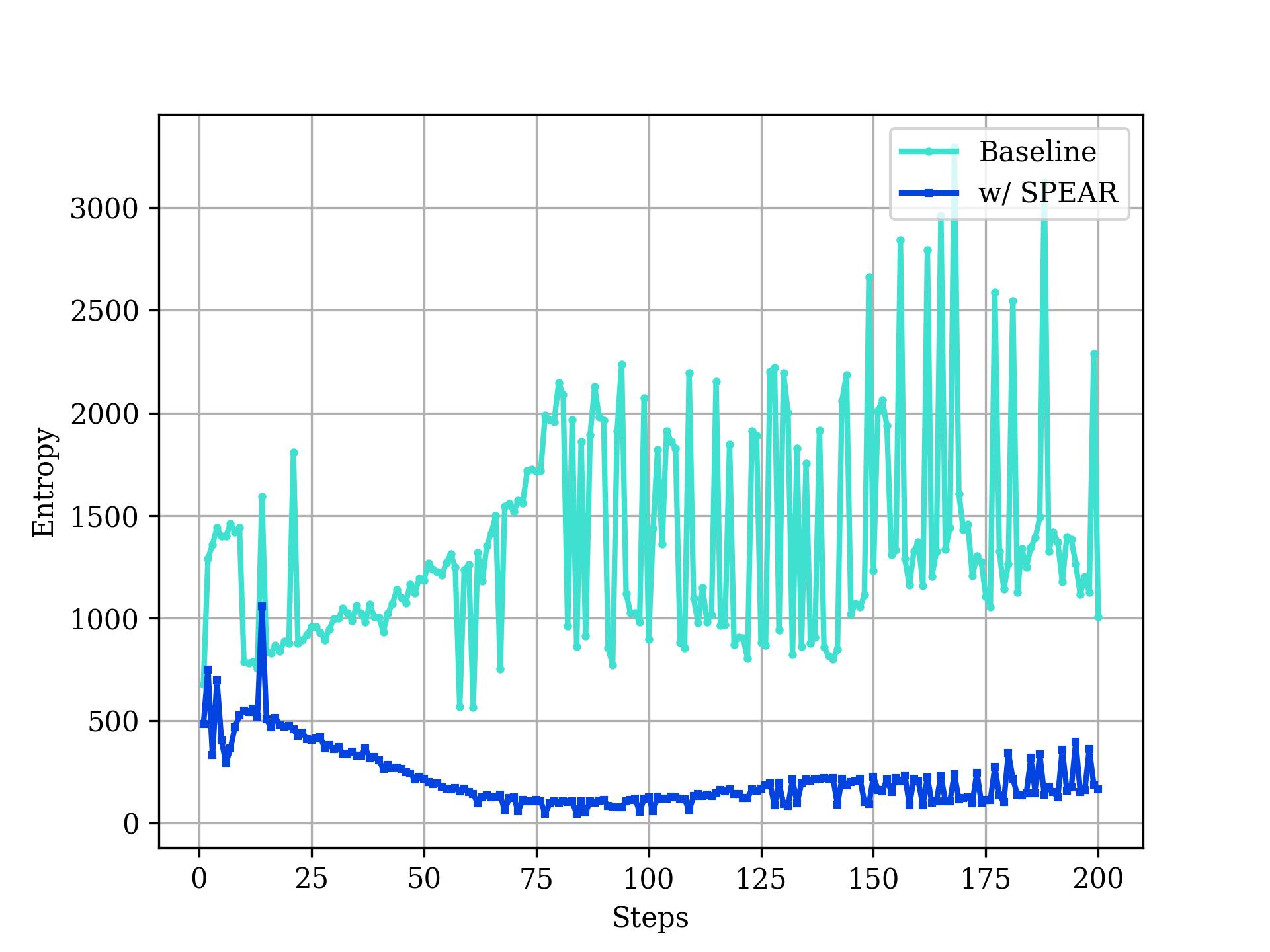}
  \caption{ALFWorld 7B \textit{Dr.BoT} (GiGPO).}
\end{subfigure}
\\
\begin{subfigure}{.4\textwidth}
  \centering
  \includegraphics[width=\textwidth]{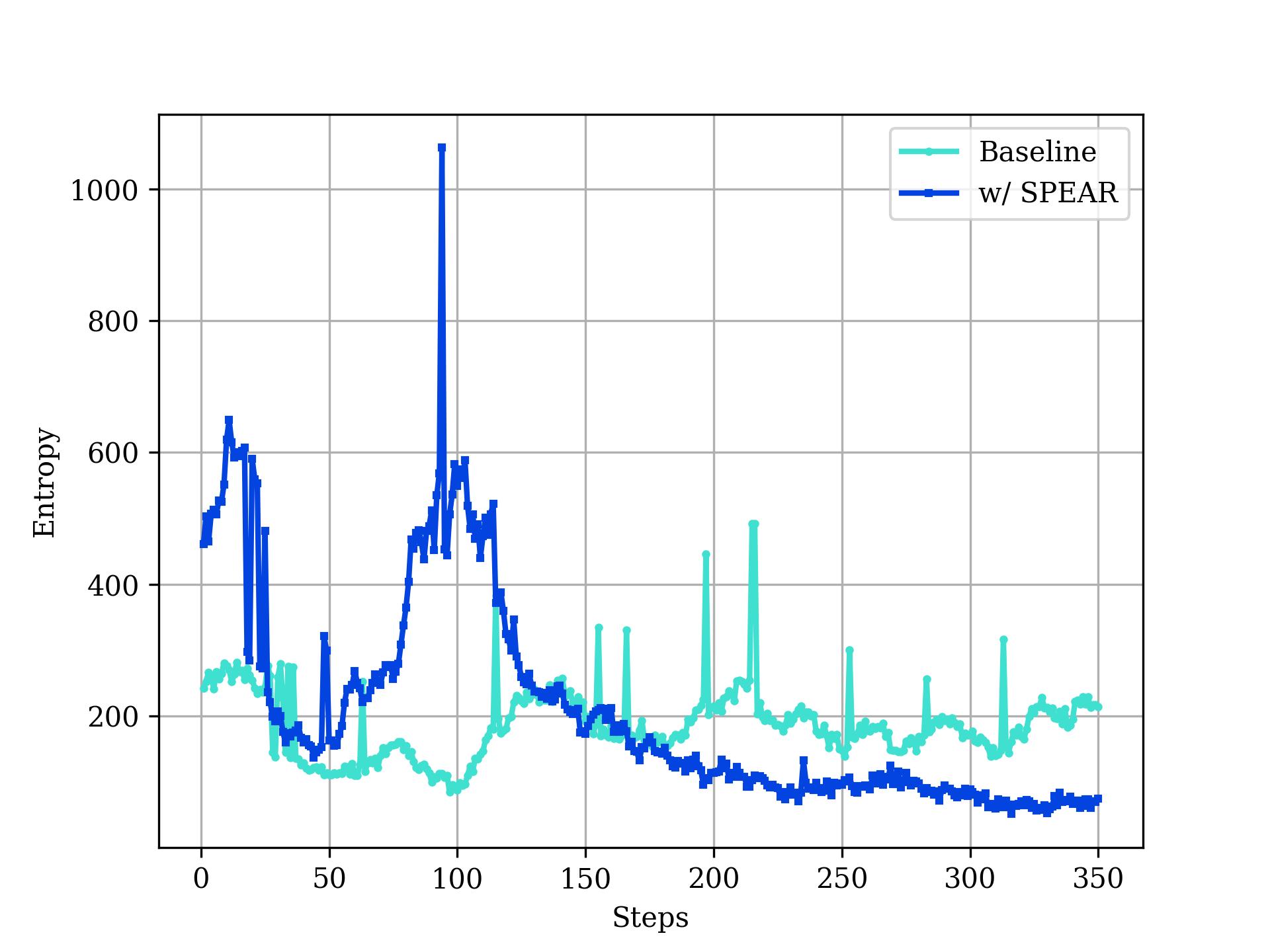}
  \caption{WebShop 1.5B \textit{Dr.BoT} (GRPO).}
\end{subfigure}
\begin{subfigure}{.4\textwidth}
  \centering
  \includegraphics[width=\textwidth]{figures/entropy_sil-alfworld-1.5b-gigpo.jpg}
  \caption{WebShop 1.5B \textit{Dr.BoT} (GiGPO).}
\end{subfigure}
\\
\begin{subfigure}{.4\textwidth}
  \centering
  \includegraphics[width=\textwidth]{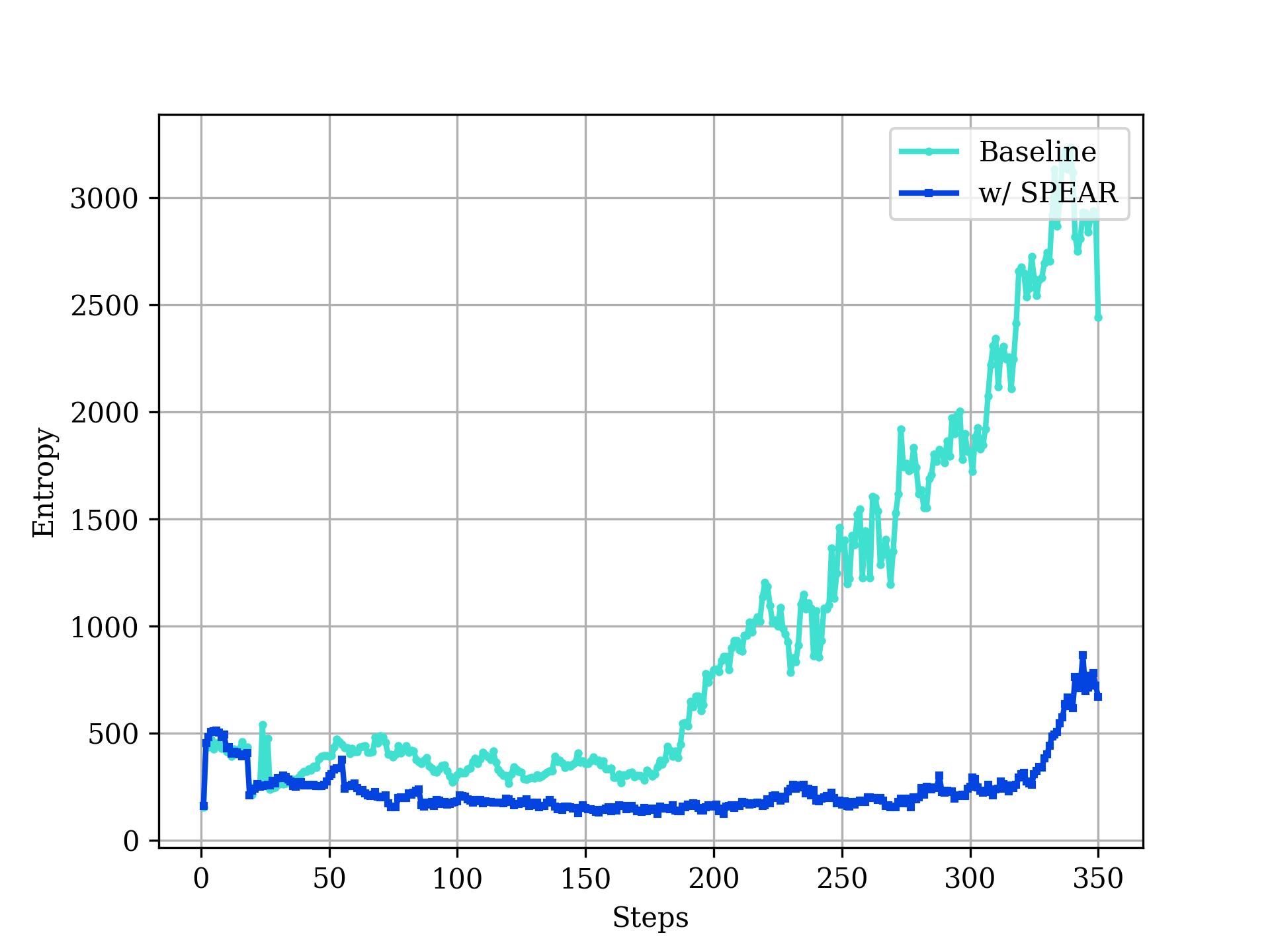}
  \caption{WebShop 7B \textit{Dr.BoT} (GRPO).}
\end{subfigure}
\begin{subfigure}{.4\textwidth}
  \centering
  \includegraphics[width=\textwidth]{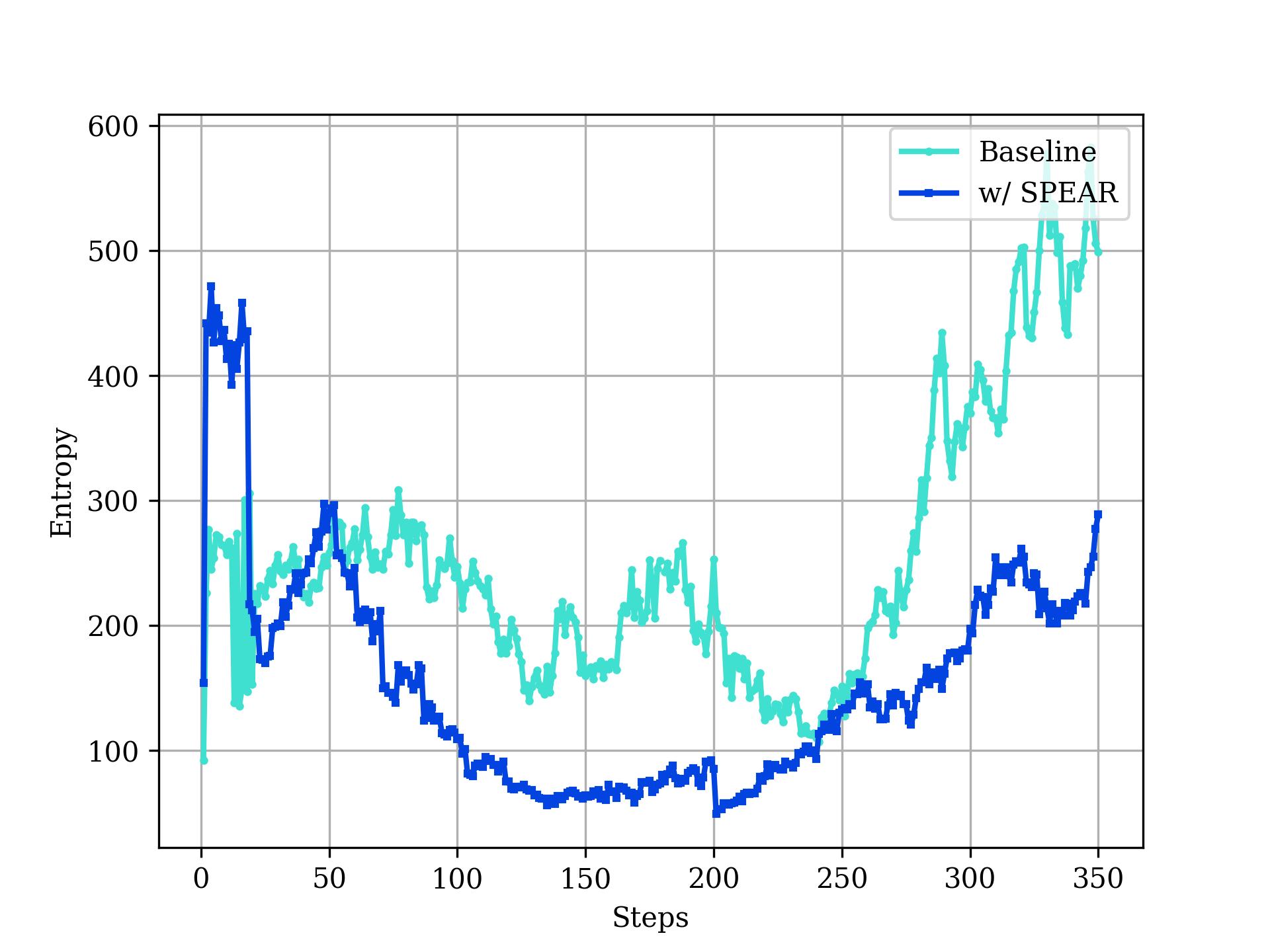}
  \caption{WebShop 7B \textit{Dr.BoT} (GiGPO).}
\end{subfigure}
\end{center}
\caption{{Entropy (\texttt{seq-mean-token-sum-norm}) across tasks, algorithms, and model scales.}
}
\label{fig:entropymeasure}
\end{figure}

\subsubsection{Additional Entropy Measurements}

Figure~\ref{fig:entropymeasure} illustrates the variance of entropy of the proposed \textit{Dr.BoT} with and without SPEAR. We can observe that:

1) For most tasks and model scales, the policy entropy of the vanilla \textit{Dr.BoT} does not converge. This is in line with our findings in Figure 3 where the consistent uncertainty about the environments and actions causes policy entropy divergence.

2) Due to the curriculum scheduling of self-imitation, the policy entropy maintains a steady trend across stages. The SPEAR allows sufficient exploration at the beginning and gradually strengthens imitation of self-generated promising experience. Therefore, the entropy varies smoothly during training.

3) Due to the curriculum scheduling of tool call reward, the interaction with the environments is encouraged and therefore the policy entropy of SPEAR can even surpass the baseline (e.g., ALFWorld 1.5B and WebShop 1.5B). However, such exploration about the environment does not necessarily correlate with entropy variation. We believe the distributional gap between task domains and the pretrained knowledge of LLMs plays a critical role. For larger models (7B), its internal parameterized knowledge is richer to handle the observation states properly.

\subsection{Training Cost and Complexity}
\label{sec:trainingcost}

The following contents are mentioned in Section~\ref{sec:discussionintotal} in the main text.

\begin{table}[htbp]
\begin{center}
\caption{Comparison on the complexity of the vanilla GRPO and the proposed \method.
PG, FW, BP, RB, and Adv respectively stand for the policy gradient loss computation, forward, back-propagation, replay buffer, and advantage.
Out of simplicity, we use the $\mathcal{O}(M)$ to denote the forward FLOPs which is positively associated with the model size and the input length.
$\mathcal{O}(P)$ denotes the BP operations proportional to the number of LLM parameters.
We use $n_{\text{SIL}}$ to refer to the equivalent number of off-policy update (by SIL) per on-policy update.
After filtering by $\hat{A}_{j}>0\ \&\ \tilde{A}_{j}>0$ (Equation~\ref{eq:silvanilla}),
the number of samples in SIL is represented as $K, K\leq N_{\mathcal{D}}$.
}
\label{tab:computationcomplexity}
\resizebox{0.8\linewidth}{!}{
\setlength{\tabcolsep}{1mm}{
\fontsize{9pt}{10pt}\selectfont{
\begin{tabular}{llll}
\toprule
\begin{tabular}[c]{@{}l@{}}Training\\ Stage\end{tabular} & \begin{tabular}[c]{@{}l@{}}Computation\\ of GRPO\\ (vanilla)\end{tabular} & \begin{tabular}[c]{@{}l@{}}Additional\\ Computation\\ by \method\end{tabular} & Description \\
\midrule
On-policy Rollout & $2 G T \mathcal{O}(M)$ & --  & FW \& sampling w/ $\pi_{\theta_{\text{old}}}$. \\
RB Update & -- &  $\mathcal{O}(G T)$  &  Copy operation (negligible).  \\
On-policy PG & $G T \mathcal{O}(M)$ & -- & FW w/ $\pi_{\theta}$ (w/o KL $\pi_{\theta_{\text{ref}}}$). \\
On-policy BP & $\mathcal{O}(P)$ & -- & BP w/ $\pi_{\theta}$. \\
RB Filtering & -- & $\mathcal{O}(N_{\mathcal{D}})$ & Look-up operation (negligible). \\
Adv Recalibration & -- & $\mathcal{O}(N_{\mathcal{D}})$+$\mathcal{O}(N_{\mathcal{D}_{R}})$ & Additive operation (negligible). \\
Replay PG & -- & \begin{tabular}[c]{@{}l@{}}$n_{\text{SIL}} K T \mathcal{O}(M)$\\ + $n_{\text{SIL}} \mathcal{O}(K T)$\end{tabular}  & 
\begin{tabular}[c]{@{}l@{}}FW w/ $\pi_{\theta}$, token-wise \\ $\text{clip} \& \min$ (negligible).\end{tabular} \\
Replay BP & -- & $n_{\text{SIL}} \mathcal{O}(P)$ & BP w/ $\pi_{\theta}$. \\
\midrule
In Total & $3 G T \mathcal{O}(M)+\mathcal{O}(P)$ & $n_{\text{SIL}}(K T \mathcal{O}(M)+\mathcal{O}(P))$ & Dominance by FW \& BP \\
\bottomrule
\end{tabular}
}}}
\end{center}
\end{table}

\begin{figure}[htbp]
\begin{center}
\begin{subfigure}{.33\textwidth}
  \centering
  \includegraphics[width=\textwidth]{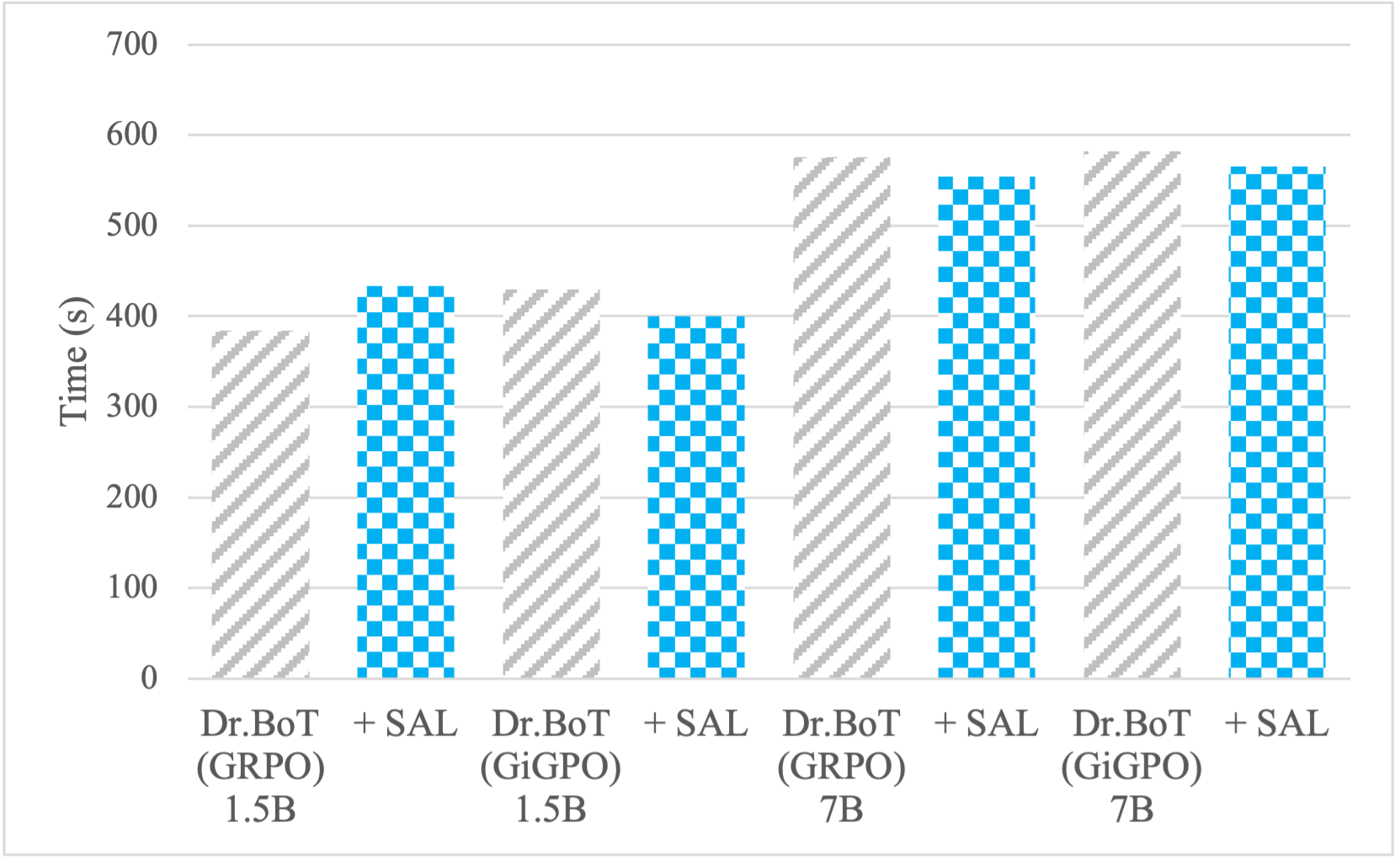}
  \caption{ALFWorld.}
\end{subfigure}%
\begin{subfigure}{.32\textwidth}
  \centering
  \includegraphics[width=\textwidth]{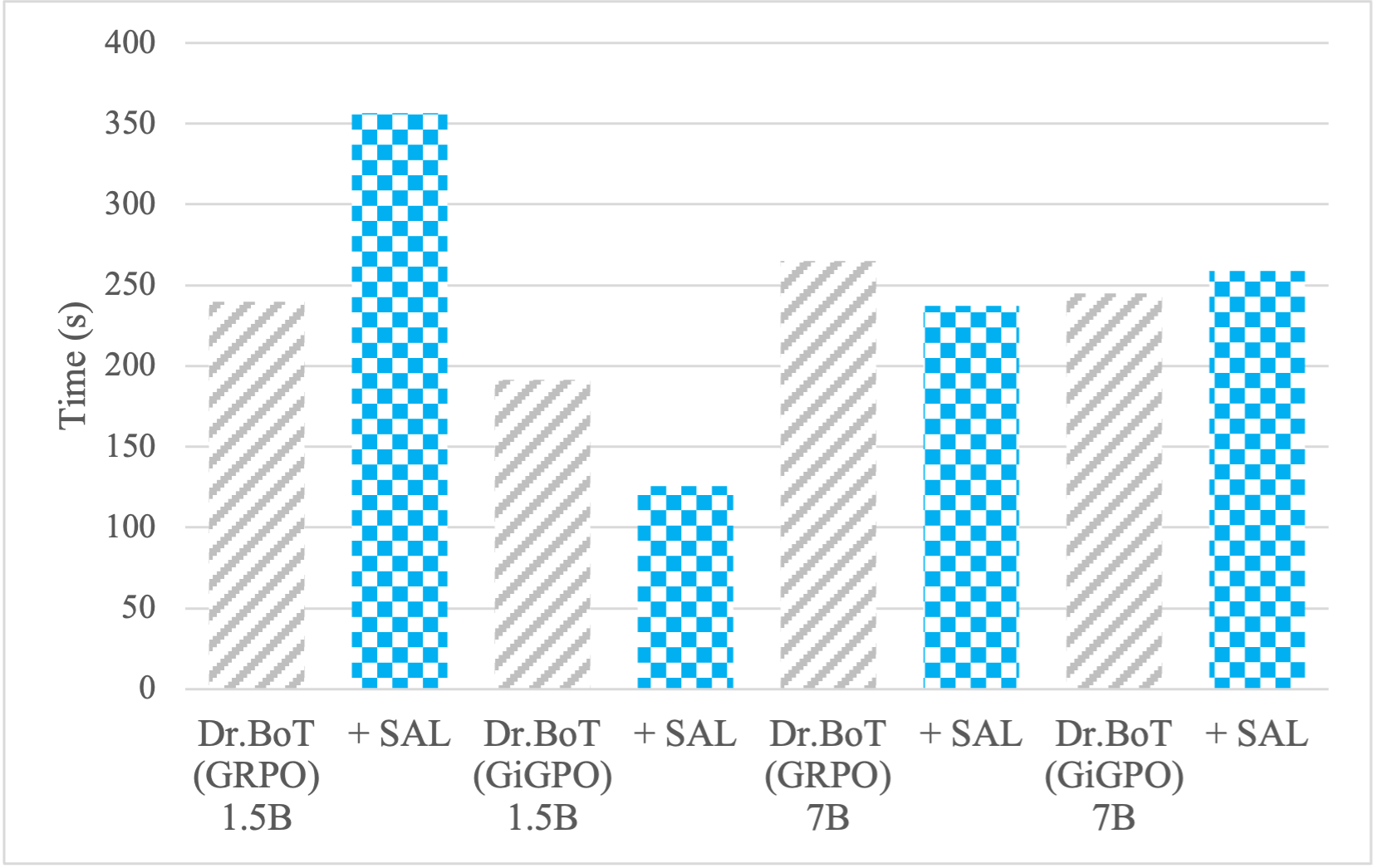}
  \caption{WebShop.}
\end{subfigure}%
\begin{subfigure}{.28\textwidth}
  \centering
  \includegraphics[width=\textwidth]{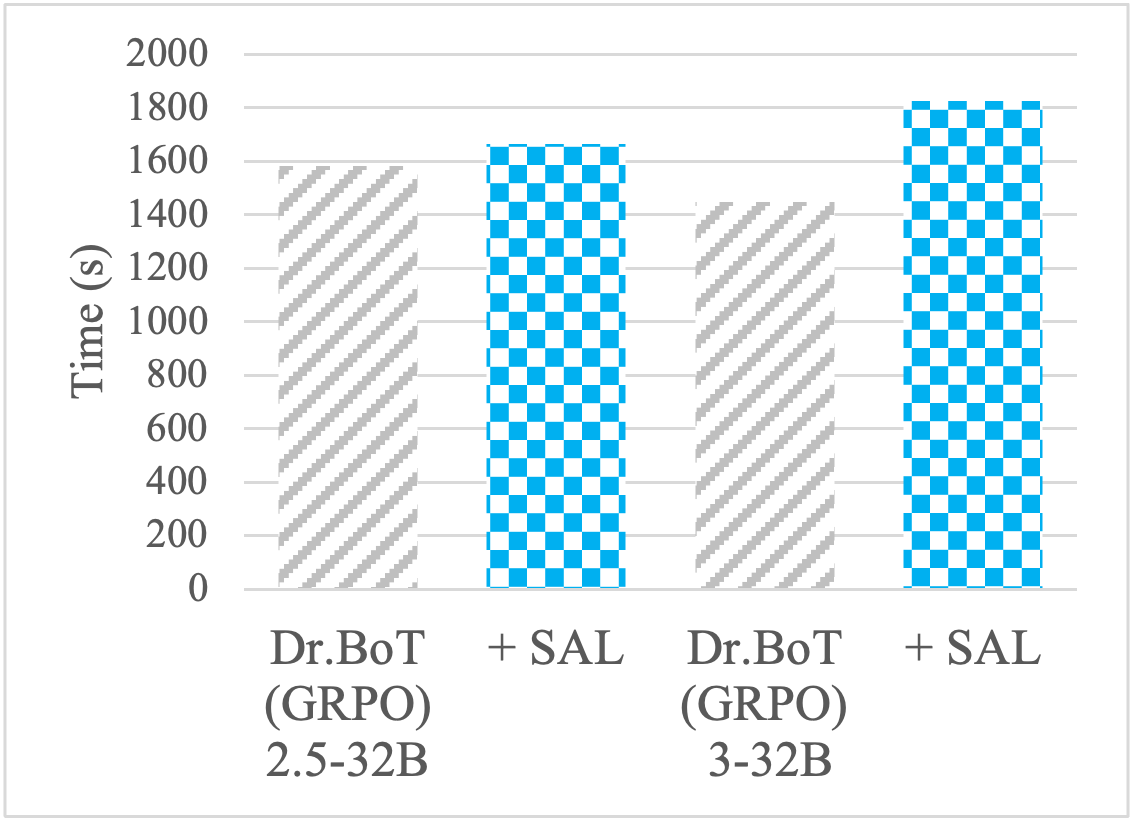}
  \caption{DAPO-MATH-17K.}
\end{subfigure}%
\end{center}
\caption{The averaged policy training time (s) per step with and without the proposed \method.
}
\label{fig:trainingtime}
\end{figure}

We compare the computational complexity of our \method with the vanilla GRPO algorithm in Table~\ref{tab:computationcomplexity}.
Most of the computation comes from the forward and back-propagation of the filtered samples in the replay buffer.
The memory operations such as the update and filtering of the buffer are light-weight and can be simply ignored.
Given the current experimental settings (see Table~\ref{tab:hyperparams_value}),
we observe that $n_{\text{SIL}}\approx0.5$ for ALFWorld and WebShop, and $n_{\text{SIL}}\approx0.33$ for DAPO-MATH-17K.
In this case,
our \method additionally introduces around $10\%\sim 25\%$ computation overhead with $K\leq G$.
Such computation complexity is acceptable in practice as the time of each training iteration is dominated by that of on-policy rollout generation.

Figure~\ref{fig:trainingtime} shows the runtime per iteration step with and without the proposed \method across different tasks and model scales.
the total optimization procedure (including the rollout generation, advantage computation, log-probability inference, reward computation, and the actor update) is quite similar on average for ALFWorld, WebShop, and their \method counterparts.
For ALFWorld and WebShop,
the 1.5B models exhibit larger variance than 7B models in training time.
We believe such variance is associated with findings of the previous study~\cite{havrilla2024teaching} that the size of LLMs matters to the exploration diversity.
Smaller LLMs are less diverse in exploring strategies due to their shallower reasoning nature,
and are therefore prone to suboptimal policies with relatively increased stochasticity in training dynamics.
For DAPO-MATH-17K,
an increase around 5\% and 26\% is observed respectively on Qwen2.5 and Qwen3 models.
Since the time per step is dominated by the rollout generation and actor update,
we believe such increase in time is caused by the longer reasoning traces, more tool call interactions, and the additional iterations from the replay buffer.
Such encouraged exploration by \method is exaggerated on the reasoning model Qwen3 and leads to longer training time.

{

It is noted that the proposed SPEAR does not increase GPU memory usage.
The introduction of the experience replay buffer is equivalent to increasing the training batch size per step.
Due to the current sequential implementation that uses gradient accumulation with a fixed mini training batch size,
we can achieve policy optimization on batches of any size without OOM issues.
}

{

\subsection{Future Work}
\label{sec:future}
\subsubsection{Dynamic Scheduling}
In the future, one of the promising research direction is to model and adjust the scheduling parameters dynamically.
It is noted that there exists no clear-cut line between exploitation and exploration during training~\cite{snoek2012practical,wang2018exploration}.
The exploitation and exploration are intertwined and optimized together, which is often context-dependent~\cite{bellemare2016unifying} or guided by the policy itself~\cite{pathak2017curiosity}.
Therefore, the scheduling should be progressive and smooth.
We believe three kinds of techniques can be utilized for guiding the exploration:
\paragraph{Entropy as the medium.}
Following ARPO~\cite{dong2025agentic}, we could schedule the self-imitation and intrinsic reward with monitoring of the entropy itself. It is direct and intuitive, and it allows flexible and frequent adjustments. However, the modeling of the relationship between policy entropy and scheduling itself is often task-dependent and parameter-involved, introducing additional computation. In addition, the entropy is prone to noise where outliers of certain tokens might interfere with the scheduling negatively.
\paragraph{Performance as the medium.}
One could also adjust the scheduling by the performance-related metrics~\cite{agrawal2012analysis} such as the task completion rate and the number of tool-calls. The association between exploration and success rate can be utilized. Furthermore, the number of tool-calls often indicates the degree of exploration with the environment. Nevertheless, the metrics might be deceptive as an early stop of exploration stimulation could lead to suboptimal convergence.
\paragraph{Curiosity or Self-confidence as the medium.}
One could intensify the exploration when the policy exhibits uncertainty~\cite{pathak2017curiosity,ladosz2022exploration} about its actions or confusion about the transition of environment states.
The policy's familiarity of the environment and its action reflects the exploration status.
But it often requires parameterized learning of the curiosity or confidence via quantification of the inconsistency between the expected state transition and the real one.

\subsubsection{Stepwise Credit Assignment}
In a extremely noisy tool ecosystem, the discrimination between good and bad experience is rather challenging merely with the outcome reward~\cite{deng2025effect,zeng2025reinforcing}.
Under such circumstance,
a process reward model (PRM) would be beneficial to provide fine-grained, stepwise supervision.
However, it remains prohibitive to conduct manual evaluation and preference annotation for training online PRMs.
Very recent studies highlight a few potential directions:
\paragraph{The usage of meta-reward via LLM-as-a-Judge.} Instead of  training a process reward from scratch, one could directly use an off-the-shelf LLM to assess each step not from the accuracy but from  the aspect of meta-reasoning~\cite{zhang2025rlvmr} behaviors (e.g., planning, exploration, and reflection).
\paragraph{The employment of implicit PRMs.} One could derive an implicit PRM~\cite{cui2025process} by reparameterization of the outcome reward as a sum of log-likelihood ratios of two LLMs over steps.  Therefore, the step-wise reward can be approximated as the differences between two adjacent steps (agent actions)~\cite{liu2025agentic}
\paragraph{The introduction of world models.} The noise from real-world tool ecosystem might be inevitable and therefore it is reasonable to perform a model-based sim2real RL~\cite{moerland2023model}. One could prepare an internal world model to deliver reliable state transition~\cite{gu2024your} for tool-based interaction, which help the agentic LLMs develop strategies via RL. Then, the trained LLM further adapts to real environment after a few more steps of training to gain robustness against noise.
}

\subsection{The Use of Large Language Models}
In the present study,
{we use the LLMs for the polishing of the manuscript writing and the discussions for analysis.}


\begin{thebibliography}{124}
\providecommand{\natexlab}[1]{#1}
\providecommand{\url}[1]{\texttt{#1}}
\expandafter\ifx\csname urlstyle\endcsname\relax
  \providecommand{\doi}[1]{doi: #1}\else
  \providecommand{\doi}{doi: \begingroup \urlstyle{rm}\Url}\fi

\bibitem[Lambert et~al.(2024)Lambert, Morrison, Pyatkin, Huang, Ivison, Brahman, Miranda, Liu, Dziri, Lyu, et~al.]{lambert2024tulu}
Nathan Lambert, Jacob Morrison, Valentina Pyatkin, Shengyi Huang, Hamish Ivison, Faeze Brahman, Lester James~V Miranda, Alisa Liu, Nouha Dziri, Shane Lyu, et~al.
\newblock Tulu 3: Pushing frontiers in open language model post-training.
\newblock \emph{arXiv preprint arXiv:2411.15124}, 2024.

\bibitem[Guo et~al.(2025)Guo, Yang, Zhang, Song, Zhang, Xu, Zhu, Ma, Wang, Bi, et~al.]{guo2025deepseek}
Daya Guo, Dejian Yang, Haowei Zhang, Junxiao Song, Ruoyu Zhang, Runxin Xu, Qihao Zhu, Shirong Ma, Peiyi Wang, Xiao Bi, et~al.
\newblock Deepseek-r1: Incentivizing reasoning capability in llms via reinforcement learning.
\newblock \emph{arXiv preprint arXiv:2501.12948}, 2025.

\bibitem[Qin et~al.(2025{\natexlab{a}})Qin, Li, Li, Xu, Shi, Lin, Cui, Li, and Sun]{qin2025incentivizing}
Yulei Qin, Gang Li, Zongyi Li, Zihan Xu, Yuchen Shi, Zhekai Lin, Xiao Cui, Ke~Li, and Xing Sun.
\newblock Incentivizing reasoning for advanced instruction-following of large language models.
\newblock \emph{arXiv preprint arXiv:2506.01413}, 2025{\natexlab{a}}.

\bibitem[Yao et~al.(2023)Yao, Zhao, Yu, Du, Shafran, Narasimhan, and Cao]{yao2023react}
Shunyu Yao, Jeffrey Zhao, Dian Yu, Nan Du, Izhak Shafran, Karthik Narasimhan, and Yuan Cao.
\newblock React: Synergizing reasoning and acting in language models.
\newblock In \emph{International Conference on Learning Representations (ICLR)}, 2023.

\bibitem[Shridhar et~al.(2020)Shridhar, Yuan, C{\^o}t{\'e}, Bisk, Trischler, and Hausknecht]{shridhar2020alfworld}
Mohit Shridhar, Xingdi Yuan, Marc-Alexandre C{\^o}t{\'e}, Yonatan Bisk, Adam Trischler, and Matthew Hausknecht.
\newblock Alfworld: Aligning text and embodied environments for interactive learning.
\newblock \emph{arXiv preprint arXiv:2010.03768}, 2020.

\bibitem[Li et~al.(2024)Li, Zhao, Wang, Wang, Zhou, Srivastava, Gokmen, Lee, Li, Zhang, et~al.]{li2024embodied}
Manling Li, Shiyu Zhao, Qineng Wang, Kangrui Wang, Yu~Zhou, Sanjana Srivastava, Cem Gokmen, Tony Lee, Erran~Li Li, Ruohan Zhang, et~al.
\newblock Embodied agent interface: Benchmarking llms for embodied decision making.
\newblock \emph{Advances in Neural Information Processing Systems}, 37:\penalty0 100428--100534, 2024.

\bibitem[Wang et~al.(2024)Wang, Xu, Jia, Zhang, Yan, Shen, Zhang, Huang, and Sang]{wang2024mobile}
Junyang Wang, Haiyang Xu, Haitao Jia, Xi~Zhang, Ming Yan, Weizhou Shen, Ji~Zhang, Fei Huang, and Jitao Sang.
\newblock Mobile-agent-v2: Mobile device operation assistant with effective navigation via multi-agent collaboration.
\newblock \emph{Advances in Neural Information Processing Systems}, 37:\penalty0 2686--2710, 2024.

\bibitem[Li et~al.(2025{\natexlab{a}})Li, Qu, Zhou, Wang, Wen, Du, Lou, Peng, and Zhang]{li2025mobileuse}
Ning Li, Xiangmou Qu, Jiamu Zhou, Jun Wang, Muning Wen, Kounianhua Du, Xingyu Lou, Qiuying Peng, and Weinan Zhang.
\newblock Mobileuse: A gui agent with hierarchical reflection for autonomous mobile operation.
\newblock \emph{arXiv preprint arXiv:2507.16853}, 2025{\natexlab{a}}.

\bibitem[Furuta et~al.(2023)Furuta, Lee, Nachum, Matsuo, Faust, Gu, and Gur]{furuta2023multimodal}
Hiroki Furuta, Kuang-Huei Lee, Ofir Nachum, Yutaka Matsuo, Aleksandra Faust, Shixiang~Shane Gu, and Izzeddin Gur.
\newblock Multimodal web navigation with instruction-finetuned foundation models.
\newblock \emph{arXiv preprint arXiv:2305.11854}, 2023.

\bibitem[He et~al.(2024)He, Yao, Ma, Yu, Dai, Zhang, Lan, and Yu]{he2024webvoyager}
Hongliang He, Wenlin Yao, Kaixin Ma, Wenhao Yu, Yong Dai, Hongming Zhang, Zhenzhong Lan, and Dong Yu.
\newblock Webvoyager: Building an end-to-end web agent with large multimodal models.
\newblock \emph{arXiv preprint arXiv:2401.13919}, 2024.

\bibitem[Jin et~al.(2025{\natexlab{a}})Jin, Zeng, Yue, Yoon, Arik, Wang, Zamani, and Han]{jin2025search}
Bowen Jin, Hansi Zeng, Zhenrui Yue, Jinsung Yoon, Sercan Arik, Dong Wang, Hamed Zamani, and Jiawei Han.
\newblock Search-r1: Training llms to reason and leverage search engines with reinforcement learning.
\newblock \emph{arXiv preprint arXiv:2503.09516}, 2025{\natexlab{a}}.

\bibitem[Li et~al.(2025{\natexlab{b}})Li, Jin, Dong, Qian, Zhu, Wu, Wen, and Dou]{li2025webthinker}
Xiaoxi Li, Jiajie Jin, Guanting Dong, Hongjin Qian, Yutao Zhu, Yongkang Wu, Ji-Rong Wen, and Zhicheng Dou.
\newblock Webthinker: Empowering large reasoning models with deep research capability.
\newblock \emph{arXiv preprint arXiv:2504.21776}, 2025{\natexlab{b}}.

\bibitem[Tao et~al.(2025)Tao, Wu, Yin, Zhang, Li, Shen, Li, Zhang, Wang, Jiang, et~al.]{tao2025webshaper}
Zhengwei Tao, Jialong Wu, Wenbiao Yin, Junkai Zhang, Baixuan Li, Haiyang Shen, Kuan Li, Liwen Zhang, Xinyu Wang, Yong Jiang, et~al.
\newblock Webshaper: Agentically data synthesizing via information-seeking formalization.
\newblock \emph{arXiv preprint arXiv:2507.15061}, 2025.

\bibitem[Qin et~al.(2025{\natexlab{b}})Qin, Ye, Fang, Wang, Liang, Tian, Zhang, Li, Li, Huang, et~al.]{qin2025ui}
Yujia Qin, Yining Ye, Junjie Fang, Haoming Wang, Shihao Liang, Shizuo Tian, Junda Zhang, Jiahao Li, Yunxin Li, Shijue Huang, et~al.
\newblock Ui-tars: Pioneering automated gui interaction with native agents.
\newblock \emph{arXiv preprint arXiv:2501.12326}, 2025{\natexlab{b}}.

\bibitem[Hong et~al.(2024)Hong, Wang, Lv, Xu, Yu, Ji, Wang, Wang, Dong, Ding, et~al.]{hong2024cogagent}
Wenyi Hong, Weihan Wang, Qingsong Lv, Jiazheng Xu, Wenmeng Yu, Junhui Ji, Yan Wang, Zihan Wang, Yuxiao Dong, Ming Ding, et~al.
\newblock Cogagent: A visual language model for gui agents.
\newblock In \emph{Proceedings of the IEEE/CVF Conference on Computer Vision and Pattern Recognition}, pages 14281--14290, 2024.

\bibitem[Sutton(1988)]{sutton1988learning}
Richard~S Sutton.
\newblock Learning to predict by the methods of temporal differences.
\newblock \emph{Machine learning}, 3\penalty0 (1):\penalty0 9--44, 1988.

\bibitem[Williams and Peng(1991)]{williams1991function}
Ronald~J Williams and Jing Peng.
\newblock Function optimization using connectionist reinforcement learning algorithms.
\newblock \emph{Connection Science}, 3\penalty0 (3):\penalty0 241--268, 1991.

\bibitem[Cui et~al.(2025{\natexlab{a}})Cui, Zhang, Chen, Yuan, Wang, Zuo, Li, Fan, Chen, Chen, et~al.]{cui2025entropy}
Ganqu Cui, Yuchen Zhang, Jiacheng Chen, Lifan Yuan, Zhi Wang, Yuxin Zuo, Haozhan Li, Yuchen Fan, Huayu Chen, Weize Chen, et~al.
\newblock The entropy mechanism of reinforcement learning for reasoning language models.
\newblock \emph{arXiv preprint arXiv:2505.22617}, 2025{\natexlab{a}}.

\bibitem[Xue et~al.(2025)Xue, Zheng, Liu, Li, Zheng, Ma, and An]{xue2025simpletir}
Zhenghai Xue, Longtao Zheng, Qian Liu, Yingru Li, Xiaosen Zheng, Zejun Ma, and Bo~An.
\newblock Simpletir: End-to-end reinforcement learning for multi-turn tool-integrated reasoning, 2025.
\newblock URL \url{https://arxiv.org/abs/2509.02479}.

\bibitem[Ziebart et~al.(2008)Ziebart, Maas, Bagnell, Dey, et~al.]{ziebart2008maximum}
Brian~D Ziebart, Andrew~L Maas, J~Andrew Bagnell, Anind~K Dey, et~al.
\newblock Maximum entropy inverse reinforcement learning.
\newblock In \emph{Aaai}, volume~8, pages 1433--1438. Chicago, IL, USA, 2008.

\bibitem[Schulman et~al.(2017{\natexlab{a}})Schulman, Wolski, Dhariwal, Radford, and Klimov]{schulman2017proximal}
John Schulman, Filip Wolski, Prafulla Dhariwal, Alec Radford, and Oleg Klimov.
\newblock Proximal policy optimization algorithms.
\newblock \emph{arXiv preprint arXiv:1707.06347}, 2017{\natexlab{a}}.

\bibitem[Haarnoja et~al.(2018)Haarnoja, Zhou, Abbeel, and Levine]{haarnoja2018soft}
Tuomas Haarnoja, Aurick Zhou, Pieter Abbeel, and Sergey Levine.
\newblock Soft actor-critic: Off-policy maximum entropy deep reinforcement learning with a stochastic actor.
\newblock In \emph{International conference on machine learning}, pages 1861--1870. Pmlr, 2018.

\bibitem[Haarnoja et~al.(2017)Haarnoja, Tang, Abbeel, and Levine]{haarnoja2017reinforcement}
Tuomas Haarnoja, Haoran Tang, Pieter Abbeel, and Sergey Levine.
\newblock Reinforcement learning with deep energy-based policies.
\newblock In \emph{International conference on machine learning}, pages 1352--1361. PMLR, 2017.

\bibitem[Zhao et~al.(2019)Zhao, Sun, and Tresp]{zhao2019maximum}
Rui Zhao, Xudong Sun, and Volker Tresp.
\newblock Maximum entropy-regularized multi-goal reinforcement learning.
\newblock In \emph{International Conference on Machine Learning}, pages 7553--7562. PMLR, 2019.

\bibitem[Xin et~al.(2020)Xin, Yu, Qin, Tang, and Zhu]{xin2020exploration}
Bo~Xin, Haixu Yu, You Qin, Qing Tang, and Zhangqing Zhu.
\newblock Exploration entropy for reinforcement learning.
\newblock \emph{Mathematical Problems in Engineering}, 2020\penalty0 (1):\penalty0 2672537, 2020.

\bibitem[Zhang et~al.(2021)Zhang, Cai, Huang, and Li]{zhang2021exploration}
Chuheng Zhang, Yuanying Cai, Longbo Huang, and Jian Li.
\newblock Exploration by maximizing r{\'e}nyi entropy for reward-free rl framework.
\newblock In \emph{Proceedings of the AAAI Conference on Artificial Intelligence}, volume~35, pages 10859--10867, 2021.

\bibitem[Seo et~al.(2021)Seo, Chen, Shin, Lee, Abbeel, and Lee]{seo2021state}
Younggyo Seo, Lili Chen, Jinwoo Shin, Honglak Lee, Pieter Abbeel, and Kimin Lee.
\newblock State entropy maximization with random encoders for efficient exploration.
\newblock In \emph{International conference on machine learning}, pages 9443--9454. PMLR, 2021.

\bibitem[Mehr et~al.(2023)Mehr, Wang, Bhatt, and Schwager]{mehr2023maximum}
Negar Mehr, Mingyu Wang, Maulik Bhatt, and Mac Schwager.
\newblock Maximum-entropy multi-agent dynamic games: Forward and inverse solutions.
\newblock \emph{IEEE transactions on robotics}, 39\penalty0 (3):\penalty0 1801--1815, 2023.

\bibitem[Kim and Sung(2023)]{kim2023adaptive}
Woojun Kim and Youngchul Sung.
\newblock An adaptive entropy-regularization framework for multi-agent reinforcement learning.
\newblock In \emph{International Conference on Machine Learning}, pages 16829--16852. PMLR, 2023.

\bibitem[Hao et~al.(2023)Hao, Yang, Tang, Bai, Liu, Meng, Liu, and Wang]{hao2023exploration}
Jianye Hao, Tianpei Yang, Hongyao Tang, Chenjia Bai, Jinyi Liu, Zhaopeng Meng, Peng Liu, and Zhen Wang.
\newblock Exploration in deep reinforcement learning: From single-agent to multiagent domain.
\newblock \emph{IEEE Transactions on Neural Networks and Learning Systems}, 35\penalty0 (7):\penalty0 8762--8782, 2023.

\bibitem[Dong et~al.(2025{\natexlab{a}})Dong, Mao, Ma, Bao, Chen, Wang, Chen, Du, Wang, Zhang, et~al.]{dong2025agentic}
Guanting Dong, Hangyu Mao, Kai Ma, Licheng Bao, Yifei Chen, Zhongyuan Wang, Zhongxia Chen, Jiazhen Du, Huiyang Wang, Fuzheng Zhang, et~al.
\newblock Agentic reinforced policy optimization.
\newblock \emph{arXiv preprint arXiv:2507.19849}, 2025{\natexlab{a}}.

\bibitem[Mai et~al.(2025)Mai, Xu, Wang, Hu, Zhang, Zhang, et~al.]{mai2025agent}
Xinji Mai, Haotian Xu, Weinong Wang, Jian Hu, Yingying Zhang, Wenqiang Zhang, et~al.
\newblock Agent rl scaling law: Agent rl with spontaneous code execution for mathematical problem solving.
\newblock \emph{arXiv preprint arXiv:2505.07773}, 2025.

\bibitem[Yao et~al.(2025)Yao, Liu, Zhang, Dong, Shang, and Gao]{yao2025offpolicy}
Feng Yao, Liyuan Liu, Dinghuai Zhang, Chengyu Dong, Jingbo Shang, and Jianfeng Gao.
\newblock Your efficient rl framework secretly brings you off-policy rl training, August 2025.
\newblock URL \url{https://fengyao.notion.site/off-policy-rl}.

\bibitem[Wang et~al.(2025{\natexlab{a}})Wang, Wang, Wang, Zhang, Li, Yang, Jin, Yu, Nguyen, Liu, et~al.]{wang2025ragen}
Zihan Wang, Kangrui Wang, Qineng Wang, Pingyue Zhang, Linjie Li, Zhengyuan Yang, Xing Jin, Kefan Yu, Minh~Nhat Nguyen, Licheng Liu, et~al.
\newblock Ragen: Understanding self-evolution in llm agents via multi-turn reinforcement learning.
\newblock \emph{arXiv preprint arXiv:2504.20073}, 2025{\natexlab{a}}.

\bibitem[Feng et~al.(2025{\natexlab{a}})Feng, Huang, Qu, Zhang, Qin, Zhong, Jiang, Chi, and Zhong]{feng2025retool}
Jiazhan Feng, Shijue Huang, Xingwei Qu, Ge~Zhang, Yujia Qin, Baoquan Zhong, Chengquan Jiang, Jinxin Chi, and Wanjun Zhong.
\newblock Retool: Reinforcement learning for strategic tool use in llms.
\newblock \emph{arXiv preprint arXiv:2504.11536}, 2025{\natexlab{a}}.

\bibitem[Qin et~al.(2025{\natexlab{c}})Qin, Yang, Guo, Li, Shao, Shi, Xu, Gu, Li, and Sun]{qin2025unleashing}
Yulei Qin, Yuncheng Yang, Pengcheng Guo, Gang Li, Hang Shao, Yuchen Shi, Zihan Xu, Yun Gu, Ke~Li, and Xing Sun.
\newblock Unleashing the power of data tsunami: A comprehensive survey on data assessment and selection for instruction tuning of language models.
\newblock \emph{Transactions on Machine Learning Research}, 2025{\natexlab{c}}.

\bibitem[Zhang et~al.(2025{\natexlab{a}})Zhang, Xie, Sun, Chen, Wang, Li, Ding, and Zhou]{zhang2025policy}
Wenhao Zhang, Yuexiang Xie, Yuchang Sun, Yanxi Chen, Guoyin Wang, Yaliang Li, Bolin Ding, and Jingren Zhou.
\newblock On-policy rl meets off-policy experts: Harmonizing supervised fine-tuning and reinforcement learning via dynamic weighting.
\newblock \emph{arXiv preprint arXiv:2508.11408}, 2025{\natexlab{a}}.

\bibitem[Oh et~al.(2018)Oh, Guo, Singh, and Lee]{oh2018self}
Junhyuk Oh, Yijie Guo, Satinder Singh, and Honglak Lee.
\newblock Self-imitation learning.
\newblock In \emph{International conference on machine learning}, pages 3878--3887. PMLR, 2018.

\bibitem[Ferret et~al.(2020)Ferret, Pietquin, and Geist]{ferret2020self}
Johan Ferret, Olivier Pietquin, and Matthieu Geist.
\newblock Self-imitation advantage learning.
\newblock \emph{arXiv preprint arXiv:2012.11989}, 2020.

\bibitem[Feng et~al.(2025{\natexlab{b}})Feng, Xue, Liu, and An]{feng2025group}
Lang Feng, Zhenghai Xue, Tingcong Liu, and Bo~An.
\newblock Group-in-group policy optimization for llm agent training.
\newblock \emph{arXiv preprint arXiv:2505.10978}, 2025{\natexlab{b}}.

\bibitem[Yao et~al.(2022)Yao, Chen, Yang, and Narasimhan]{yao2022webshop}
Shunyu Yao, Howard Chen, John Yang, and Karthik Narasimhan.
\newblock Webshop: Towards scalable real-world web interaction with grounded language agents.
\newblock \emph{Advances in Neural Information Processing Systems}, 35:\penalty0 20744--20757, 2022.

\bibitem[AIME(2025)]{aime}
AIME.
\newblock Aime problems and solutions.
\newblock \url{https://artofproblemsolving.com/wiki/index.php/AIME_Problems_and_Solutions}, 2025.

\bibitem[Jaech et~al.(2024)Jaech, Kalai, Lerer, Richardson, El-Kishky, Low, Helyar, Madry, Beutel, Carney, et~al.]{jaech2024openai}
Aaron Jaech, Adam Kalai, Adam Lerer, Adam Richardson, Ahmed El-Kishky, Aiden Low, Alec Helyar, Aleksander Madry, Alex Beutel, Alex Carney, et~al.
\newblock Openai o1 system card.
\newblock \emph{arXiv preprint arXiv:2412.16720}, 2024.

\bibitem[Ouyang et~al.(2022)Ouyang, Wu, Jiang, Almeida, Wainwright, Mishkin, Zhang, Agarwal, Slama, Ray, et~al.]{ouyang2022training}
Long Ouyang, Jeffrey Wu, Xu~Jiang, Diogo Almeida, Carroll Wainwright, Pamela Mishkin, Chong Zhang, Sandhini Agarwal, Katarina Slama, Alex Ray, et~al.
\newblock Training language models to follow instructions with human feedback.
\newblock \emph{Advances in neural information processing systems}, 35:\penalty0 27730--27744, 2022.

\bibitem[Shao et~al.(2024)Shao, Wang, Zhu, Xu, Song, Bi, Zhang, Zhang, Li, Wu, et~al.]{shao2024deepseekmath}
Zhihong Shao, Peiyi Wang, Qihao Zhu, Runxin Xu, Junxiao Song, Xiao Bi, Haowei Zhang, Mingchuan Zhang, YK~Li, Yang Wu, et~al.
\newblock Deepseekmath: Pushing the limits of mathematical reasoning in open language models.
\newblock \emph{arXiv preprint arXiv:2402.03300}, 2024.

\bibitem[Yu et~al.(2025)Yu, Zhang, Zhu, Yuan, Zuo, Yue, Dai, Fan, Liu, Liu, et~al.]{yu2025dapo}
Qiying Yu, Zheng Zhang, Ruofei Zhu, Yufeng Yuan, Xiaochen Zuo, Yu~Yue, Weinan Dai, Tiantian Fan, Gaohong Liu, Lingjun Liu, et~al.
\newblock Dapo: An open-source llm reinforcement learning system at scale.
\newblock \emph{arXiv preprint arXiv:2503.14476}, 2025.

\bibitem[Liu et~al.(2025{\natexlab{a}})Liu, Chen, Li, Qi, Pang, Du, Lee, and Lin]{liu2025understanding}
Zichen Liu, Changyu Chen, Wenjun Li, Penghui Qi, Tianyu Pang, Chao Du, Wee~Sun Lee, and Min Lin.
\newblock Understanding r1-zero-like training: A critical perspective.
\newblock \emph{arXiv preprint arXiv:2503.20783}, 2025{\natexlab{a}}.

\bibitem[Li et~al.(2025{\natexlab{c}})Li, Zou, and Liu]{li2025torl}
Xuefeng Li, Haoyang Zou, and Pengfei Liu.
\newblock Torl: Scaling tool-integrated rl.
\newblock \emph{arXiv preprint arXiv:2503.23383}, 2025{\natexlab{c}}.

\bibitem[Gao et~al.(2025)Gao, Fu, Xie, Xu, He, Mei, Zhu, and Wu]{gao2025beyond}
Jiaxuan Gao, Wei Fu, Minyang Xie, Shusheng Xu, Chuyi He, Zhiyu Mei, Banghua Zhu, and Yi~Wu.
\newblock Beyond ten turns: Unlocking long-horizon agentic search with large-scale asynchronous rl.
\newblock \emph{arXiv preprint arXiv:2508.07976}, 2025.

\bibitem[Pathak et~al.(2017)Pathak, Agrawal, Efros, and Darrell]{pathak2017curiosity}
Deepak Pathak, Pulkit Agrawal, Alexei~A Efros, and Trevor Darrell.
\newblock Curiosity-driven exploration by self-supervised prediction.
\newblock In \emph{International conference on machine learning}, pages 2778--2787. PMLR, 2017.

\bibitem[Houthooft et~al.(2016)Houthooft, Chen, Duan, Schulman, De~Turck, and Abbeel]{houthooft2016vime}
Rein Houthooft, Xi~Chen, Yan Duan, John Schulman, Filip De~Turck, and Pieter Abbeel.
\newblock Vime: Variational information maximizing exploration.
\newblock \emph{Advances in neural information processing systems}, 29, 2016.

\bibitem[Bellemare et~al.(2016)Bellemare, Srinivasan, Ostrovski, Schaul, Saxton, and Munos]{bellemare2016unifying}
Marc Bellemare, Sriram Srinivasan, Georg Ostrovski, Tom Schaul, David Saxton, and Remi Munos.
\newblock Unifying count-based exploration and intrinsic motivation.
\newblock \emph{Advances in neural information processing systems}, 29, 2016.

\bibitem[Tang et~al.(2017)Tang, Houthooft, Foote, Stooke, Xi~Chen, Duan, Schulman, DeTurck, and Abbeel]{tang2017exploration}
Haoran Tang, Rein Houthooft, Davis Foote, Adam Stooke, OpenAI Xi~Chen, Yan Duan, John Schulman, Filip DeTurck, and Pieter Abbeel.
\newblock \# exploration: A study of count-based exploration for deep reinforcement learning.
\newblock \emph{Advances in neural information processing systems}, 30, 2017.

\bibitem[Gregor et~al.(2016)Gregor, Rezende, and Wierstra]{gregor2016variational}
Karol Gregor, Danilo~Jimenez Rezende, and Daan Wierstra.
\newblock Variational intrinsic control.
\newblock \emph{arXiv preprint arXiv:1611.07507}, 2016.

\bibitem[Eysenbach et~al.(2018)Eysenbach, Gupta, Ibarz, and Levine]{eysenbach2018diversity}
Benjamin Eysenbach, Abhishek Gupta, Julian Ibarz, and Sergey Levine.
\newblock Diversity is all you need: Learning skills without a reward function.
\newblock \emph{arXiv preprint arXiv:1802.06070}, 2018.

\bibitem[Schaul et~al.(2015)Schaul, Quan, Antonoglou, and Silver]{schaul2015prioritized}
Tom Schaul, John Quan, Ioannis Antonoglou, and David Silver.
\newblock Prioritized experience replay.
\newblock \emph{arXiv preprint arXiv:1511.05952}, 2015.

\bibitem[Horgan et~al.(2018)Horgan, Quan, Budden, Barth-Maron, Hessel, Van~Hasselt, and Silver]{horgan2018distributed}
Dan Horgan, John Quan, David Budden, Gabriel Barth-Maron, Matteo Hessel, Hado Van~Hasselt, and David Silver.
\newblock Distributed prioritized experience replay.
\newblock \emph{arXiv preprint arXiv:1803.00933}, 2018.

\bibitem[Gangwani et~al.(2018)Gangwani, Liu, and Peng]{gangwani2018learning}
Tanmay Gangwani, Qiang Liu, and Jian Peng.
\newblock Learning self-imitating diverse policies.
\newblock \emph{arXiv preprint arXiv:1805.10309}, 2018.

\bibitem[Pan et~al.(2022)Pan, Mei, Farahmand, White, Yao, Rohani, and Luo]{pan2022understanding}
Yangchen Pan, Jincheng Mei, Amir-massoud Farahmand, Martha White, Hengshuai Yao, Mohsen Rohani, and Jun Luo.
\newblock Understanding and mitigating the limitations of prioritized experience replay.
\newblock In \emph{Uncertainty in Artificial Intelligence}, pages 1561--1571. PMLR, 2022.

\bibitem[Saglam et~al.(2023)Saglam, Mutlu, Cicek, and Kozat]{saglam2023actor}
Baturay Saglam, Furkan~B Mutlu, Dogan~C Cicek, and Suleyman~S Kozat.
\newblock Actor prioritized experience replay.
\newblock \emph{Journal of Artificial Intelligence Research}, 78:\penalty0 639--672, 2023.

\bibitem[Tang(2020)]{tang2020self}
Yunhao Tang.
\newblock Self-imitation learning via generalized lower bound q-learning.
\newblock \emph{Advances in neural information processing systems}, 33:\penalty0 13964--13975, 2020.

\bibitem[Pshikhachev et~al.(2022)Pshikhachev, Ivanov, Egorov, and Shpilman]{pshikhachev2022self}
Georgiy Pshikhachev, Dmitry Ivanov, Vladimir Egorov, and Aleksei Shpilman.
\newblock Self-imitation learning from demonstrations.
\newblock \emph{arXiv preprint arXiv:2203.10905}, 2022.

\bibitem[Xiao et~al.(2024)Xiao, Li, Yuan, Zhu, Cui, and Honavar]{xiao2024leverage}
Teng Xiao, Mingxiao Li, Yige Yuan, Huaisheng Zhu, Chao Cui, and Vasant~G Honavar.
\newblock How to leverage demonstration data in alignment for large language model? a self-imitation learning perspective.
\newblock \emph{arXiv preprint arXiv:2410.10093}, 2024.

\bibitem[Dong et~al.(2025{\natexlab{b}})Dong, Chen, Li, Jin, Qian, Zhu, Mao, Zhou, Dou, and Wen]{dong2025tool}
Guanting Dong, Yifei Chen, Xiaoxi Li, Jiajie Jin, Hongjin Qian, Yutao Zhu, Hangyu Mao, Guorui Zhou, Zhicheng Dou, and Ji-Rong Wen.
\newblock Tool-star: Empowering llm-brained multi-tool reasoner via reinforcement learning.
\newblock \emph{arXiv preprint arXiv:2505.16410}, 2025{\natexlab{b}}.

\bibitem[Schulman et~al.(2017{\natexlab{b}})Schulman, Chen, and Abbeel]{schulman2017equivalence}
John Schulman, Xi~Chen, and Pieter Abbeel.
\newblock Equivalence between policy gradients and soft q-learning.
\newblock \emph{arXiv preprint arXiv:1704.06440}, 2017{\natexlab{b}}.

\bibitem[Schulman et~al.(2015)Schulman, Moritz, Levine, Jordan, and Abbeel]{schulman2015high}
John Schulman, Philipp Moritz, Sergey Levine, Michael Jordan, and Pieter Abbeel.
\newblock High-dimensional continuous control using generalized advantage estimation.
\newblock \emph{arXiv preprint arXiv:1506.02438}, 2015.

\bibitem[Luo et~al.(2021)Luo, Kasaei, and Schomaker]{luo2021self}
Sha Luo, Hamidreza Kasaei, and Lambert Schomaker.
\newblock Self-imitation learning by planning.
\newblock In \emph{2021 IEEE International Conference on Robotics and Automation (ICRA)}, pages 4823--4829. IEEE, 2021.

\bibitem[Qian et~al.(2025)Qian, Acikgoz, He, Wang, Chen, Hakkani-T{\"u}r, Tur, and Ji]{qian2025toolrl}
Cheng Qian, Emre~Can Acikgoz, Qi~He, Hongru Wang, Xiusi Chen, Dilek Hakkani-T{\"u}r, Gokhan Tur, and Heng Ji.
\newblock Toolrl: Reward is all tool learning needs.
\newblock \emph{arXiv preprint arXiv:2504.13958}, 2025.

\bibitem[Li et~al.(2023)Li, Chai, Wang, Sun, Tian, Zhang, and Wu]{li2023tool}
Lei Li, Yekun Chai, Shuohuan Wang, Yu~Sun, Hao Tian, Ningyu Zhang, and Hua Wu.
\newblock Tool-augmented reward modeling.
\newblock \emph{arXiv preprint arXiv:2310.01045}, 2023.

\bibitem[Da et~al.(2025)Da, Wang, Deng, Ma, Barhate, and Hendryx]{da2025agent}
Jeff Da, Clinton Wang, Xiang Deng, Yuntao Ma, Nikhil Barhate, and Sean Hendryx.
\newblock Agent-rlvr: Training software engineering agents via guidance and environment rewards.
\newblock \emph{arXiv preprint arXiv:2506.11425}, 2025.

\bibitem[Xia et~al.(2025)Xia, Fan, Chen, Yan, Cong, Zhang, Lu, Lin, Liu, and Sun]{xia2025agentrm}
Yu~Xia, Jingru Fan, Weize Chen, Siyu Yan, Xin Cong, Zhong Zhang, Yaxi Lu, Yankai Lin, Zhiyuan Liu, and Maosong Sun.
\newblock Agentrm: Enhancing agent generalization with reward modeling.
\newblock \emph{arXiv preprint arXiv:2502.18407}, 2025.

\bibitem[Singh et~al.(2025)Singh, Magazine, Pandya, and Nambi]{singh2025agentic}
Joykirat Singh, Raghav Magazine, Yash Pandya, and Akshay Nambi.
\newblock Agentic reasoning and tool integration for llms via reinforcement learning.
\newblock \emph{arXiv preprint arXiv:2505.01441}, 2025.

\bibitem[Wei et~al.(2025)Wei, Yu, Weng, Pan, Li, and Du]{wei2025autotir}
Yifan Wei, Xiaoyan Yu, Yixuan Weng, Tengfei Pan, Angsheng Li, and Li~Du.
\newblock Autotir: Autonomous tools integrated reasoning via reinforcement learning.
\newblock \emph{arXiv preprint arXiv:2507.21836}, 2025.

\bibitem[Gou et~al.(2023)Gou, Shao, Gong, Shen, Yang, Huang, Duan, and Chen]{gou2023tora}
Zhibin Gou, Zhihong Shao, Yeyun Gong, Yelong Shen, Yujiu Yang, Minlie Huang, Nan Duan, and Weizhu Chen.
\newblock Tora: A tool-integrated reasoning agent for mathematical problem solving.
\newblock \emph{arXiv preprint arXiv:2309.17452}, 2023.

\bibitem[Lin and Xu(2025)]{lin2025understanding}
Heng Lin and Zhongwen Xu.
\newblock Understanding tool-integrated reasoning.
\newblock \emph{arXiv preprint arXiv:2508.19201}, 2025.

\bibitem[Liu et~al.(2025{\natexlab{b}})Liu, Liu, He, Wang, Liu, Pan, Hu, Xiong, Huang, Hu, et~al.]{liu2025part}
Zihe Liu, Jiashun Liu, Yancheng He, Weixun Wang, Jiaheng Liu, Ling Pan, Xinyu Hu, Shaopan Xiong, Ju~Huang, Jian Hu, et~al.
\newblock Part i: Tricks or traps? a deep dive into rl for llm reasoning.
\newblock \emph{arXiv preprint arXiv:2508.08221}, 2025{\natexlab{b}}.

\bibitem[Sun et~al.(2025)Sun, Li, Chen, Qin, and Hu]{sun2025stabilizing}
Zetian Sun, Dongfang Li, Zhuoen Chen, Yuhuai Qin, and Baotian Hu.
\newblock Stabilizing long-term multi-turn reinforcement learning with gated rewards.
\newblock \emph{arXiv preprint arXiv:2508.10548}, 2025.

\bibitem[Bai et~al.(2025{\natexlab{a}})Bai, Min, Zhang, Chen, Zhao, Fang, Liu, Wang, and Wen]{bai2025towards}
Fei Bai, Yingqian Min, Beichen Zhang, Zhipeng Chen, Wayne~Xin Zhao, Lei Fang, Zheng Liu, Zhongyuan Wang, and Ji-Rong Wen.
\newblock Towards effective code-integrated reasoning.
\newblock \emph{arXiv preprint arXiv:2505.24480}, 2025{\natexlab{a}}.

\bibitem[Yang et~al.(2025)Yang, Li, Yang, Zhang, Hui, Zheng, Yu, Gao, Huang, Lv, et~al.]{yang2025qwen3}
An~Yang, Anfeng Li, Baosong Yang, Beichen Zhang, Binyuan Hui, Bo~Zheng, Bowen Yu, Chang Gao, Chengen Huang, Chenxu Lv, et~al.
\newblock Qwen3 technical report.
\newblock \emph{arXiv preprint arXiv:2505.09388}, 2025.

\bibitem[Schrader(2018)]{SchraderSokoban2018}
Max-Philipp~B. Schrader.
\newblock gym-sokoban.
\newblock \url{https://github.com/mpSchrader/gym-sokoban}, 2018.

\bibitem[Bai et~al.(2025{\natexlab{b}})Bai, Chen, Liu, Wang, Ge, Song, Dang, Wang, Wang, Tang, et~al.]{bai2025qwen2}
Shuai Bai, Keqin Chen, Xuejing Liu, Jialin Wang, Wenbin Ge, Sibo Song, Kai Dang, Peng Wang, Shijie Wang, Jun Tang, et~al.
\newblock Qwen2. 5-vl technical report.
\newblock \emph{arXiv preprint arXiv:2502.13923}, 2025{\natexlab{b}}.

\bibitem[Kwiatkowski et~al.(2019)Kwiatkowski, Palomaki, Redfield, Collins, Parikh, Alberti, Epstein, Polosukhin, Devlin, Lee, et~al.]{kwiatkowski2019natural}
Tom Kwiatkowski, Jennimaria Palomaki, Olivia Redfield, Michael Collins, Ankur Parikh, Chris Alberti, Danielle Epstein, Illia Polosukhin, Jacob Devlin, Kenton Lee, et~al.
\newblock Natural questions: a benchmark for question answering research.
\newblock \emph{Transactions of the Association for Computational Linguistics}, 7:\penalty0 453--466, 2019.

\bibitem[Joshi et~al.(2017)Joshi, Choi, Weld, and Zettlemoyer]{joshi2017triviaqa}
Mandar Joshi, Eunsol Choi, Daniel~S Weld, and Luke Zettlemoyer.
\newblock Triviaqa: A large scale distantly supervised challenge dataset for reading comprehension.
\newblock \emph{arXiv preprint arXiv:1705.03551}, 2017.

\bibitem[Mallen et~al.(2023)Mallen, Asai, Zhong, Das, Khashabi, and Hajishirzi]{mallen2023not}
Alex Mallen, Akari Asai, Victor Zhong, Rajarshi Das, Daniel Khashabi, and Hannaneh Hajishirzi.
\newblock When not to trust language models: Investigating effectiveness of parametric and non-parametric memories.
\newblock In \emph{Proceedings of the 61st Annual Meeting of the Association for Computational Linguistics (Volume 1: Long Papers)}, pages 9802--9822, 2023.

\bibitem[Yang et~al.(2018)Yang, Qi, Zhang, Bengio, Cohen, Salakhutdinov, and Manning]{yang2018hotpotqa}
Zhilin Yang, Peng Qi, Saizheng Zhang, Yoshua Bengio, William Cohen, Ruslan Salakhutdinov, and Christopher~D Manning.
\newblock Hotpotqa: A dataset for diverse, explainable multi-hop question answering.
\newblock In \emph{Proceedings of the 2018 conference on empirical methods in natural language processing}, pages 2369--2380, 2018.

\bibitem[Ho et~al.(2020)Ho, Nguyen, Sugawara, and Aizawa]{ho2020constructing}
Xanh Ho, Anh-Khoa~Duong Nguyen, Saku Sugawara, and Akiko Aizawa.
\newblock Constructing a multi-hop qa dataset for comprehensive evaluation of reasoning steps.
\newblock \emph{arXiv preprint arXiv:2011.01060}, 2020.

\bibitem[Trivedi et~al.(2022)Trivedi, Balasubramanian, Khot, and Sabharwal]{trivedi2022musique}
Harsh Trivedi, Niranjan Balasubramanian, Tushar Khot, and Ashish Sabharwal.
\newblock Musique: Multihop questions via single-hop question composition.
\newblock \emph{Transactions of the Association for Computational Linguistics}, 10:\penalty0 539--554, 2022.

\bibitem[Press et~al.(2023)Press, Zhang, Min, Schmidt, Smith, and Lewis]{press2023measuring}
Ofir Press, Muru Zhang, Sewon Min, Ludwig Schmidt, Noah~A Smith, and Mike Lewis.
\newblock Measuring and narrowing the compositionality gap in language models.
\newblock In \emph{Findings of the Association for Computational Linguistics: EMNLP 2023}, pages 5687--5711, 2023.

\bibitem[Jin et~al.(2025{\natexlab{b}})Jin, Yoon, Kargupta, Arik, and Han]{jin2025empirical}
Bowen Jin, Jinsung Yoon, Priyanka Kargupta, Sercan~O Arik, and Jiawei Han.
\newblock An empirical study on reinforcement learning for reasoning-search interleaved llm agents.
\newblock \emph{arXiv preprint arXiv:2505.15117}, 2025{\natexlab{b}}.

\bibitem[Zhang et~al.(2025{\natexlab{b}})Zhang, Chen, Li, Tu, and Li]{zhang2025rlvmr}
Zijing Zhang, Ziyang Chen, Mingxiao Li, Zhaopeng Tu, and Xiaolong Li.
\newblock Rlvmr: Reinforcement learning with verifiable meta-reasoning rewards for robust long-horizon agents.
\newblock \emph{arXiv preprint arXiv:2507.22844}, 2025{\natexlab{b}}.

\bibitem[Wu et~al.(2025)Wu, Zhou, Ziheng, Peng, Ye, Hu, Zhu, Qi, Yang, and Yang]{wu2025generalization}
Yongliang Wu, Yizhou Zhou, Zhou Ziheng, Yingzhe Peng, Xinyu Ye, Xinting Hu, Wenbo Zhu, Lu~Qi, Ming-Hsuan Yang, and Xu~Yang.
\newblock On the generalization of sft: A reinforcement learning perspective with reward rectification.
\newblock \emph{arXiv preprint arXiv:2508.05629}, 2025.

\bibitem[C{\^o}t{\'e} et~al.(2018)C{\^o}t{\'e}, K{\'a}d{\'a}r, Yuan, Kybartas, Barnes, Fine, Moore, Hausknecht, El~Asri, Adada, et~al.]{cote2018textworld}
Marc-Alexandre C{\^o}t{\'e}, Akos K{\'a}d{\'a}r, Xingdi Yuan, Ben Kybartas, Tavian Barnes, Emery Fine, James Moore, Matthew Hausknecht, Layla El~Asri, Mahmoud Adada, et~al.
\newblock Textworld: A learning environment for text-based games.
\newblock In \emph{Workshop on Computer Games}, pages 41--75. Springer, 2018.

\bibitem[Bytedance-Seed-Foundation-Code-Team et~al.(2025)Bytedance-Seed-Foundation-Code-Team, :, Cheng, Chen, Chen, Chen, Chen, Chen, Chen, Geng, Li, Li, Li, Li, Liu, Liu, Liu, Liu, Liu, Liu, Liu, Liu, Liu, Long, Mai, Ning, Peng, Shen, Su, Su, Sun, Sun, Tao, Wang, Wang, Wang, Wang, Wang, Xia, Xiang, Xiao, Xiao, Xi, Xin, Xu, Xu, Yang, Yang, Yang, Yuan, Zhang, Zhang, Zhang, Zheng, Zhu, and Zhu]{bytedanceseedfoundationcodeteam2025fullstackbenchevaluatingllms}
Bytedance-Seed-Foundation-Code-Team, :, Yao Cheng, Jianfeng Chen, Jie Chen, Li~Chen, Liyu Chen, Wentao Chen, Zhengyu Chen, Shijie Geng, Aoyan Li, Bo~Li, Bowen Li, Linyi Li, Boyi Liu, Jiaheng Liu, Kaibo Liu, Qi~Liu, Shukai Liu, Siyao Liu, Tianyi Liu, Tingkai Liu, Yongfei Liu, Rui Long, Jing Mai, Guanghan Ning, Z.~Y. Peng, Kai Shen, Jiahao Su, Jing Su, Tao Sun, Yifan Sun, Yunzhe Tao, Guoyin Wang, Siwei Wang, Xuwu Wang, Yite Wang, Zihan Wang, Jinxiang Xia, Liang Xiang, Xia Xiao, Yongsheng Xiao, Chenguang Xi, Shulin Xin, Jingjing Xu, Shikun Xu, Hongxia Yang, Jack Yang, Yingxiang Yang, Jianbo Yuan, Jun Zhang, Yufeng Zhang, Yuyu Zhang, Shen Zheng, He~Zhu, and Ming Zhu.
\newblock Fullstack bench: Evaluating llms as full stack coders, 2025.
\newblock URL \url{https://arxiv.org/abs/2412.00535}.

\bibitem[Sutton et~al.(1998)Sutton, Barto, et~al.]{sutton1998reinforcement}
Richard~S Sutton, Andrew~G Barto, et~al.
\newblock \emph{Reinforcement learning: An introduction}, volume~1.
\newblock MIT press Cambridge, 1998.

\bibitem[Christiano et~al.(2017)Christiano, Leike, Brown, Martic, Legg, and Amodei]{christiano2017deep}
Paul~F Christiano, Jan Leike, Tom Brown, Miljan Martic, Shane Legg, and Dario Amodei.
\newblock Deep reinforcement learning from human preferences.
\newblock \emph{Advances in neural information processing systems}, 30, 2017.

\bibitem[Sheng et~al.(2024)Sheng, Zhang, Ye, Wu, Zhang, Zhang, Peng, Lin, and Wu]{sheng2024hybridflow}
Guangming Sheng, Chi Zhang, Zilingfeng Ye, Xibin Wu, Wang Zhang, Ru~Zhang, Yanghua Peng, Haibin Lin, and Chuan Wu.
\newblock Hybridflow: A flexible and efficient rlhf framework.
\newblock \emph{arXiv preprint arXiv: 2409.19256}, 2024.

\bibitem[Zhuang et~al.(2025)Zhuang, Vu, Dimakis, and Sathiamoorthy]{bespoke_improving_multi_turn_tool_use}
Richard Zhuang, Trung Vu, Alex Dimakis, and Maheswaran Sathiamoorthy.
\newblock Improving multi-turn tool use with reinforcement learning.
\newblock https://www.bespokelabs.ai/blog/improving-multi-turn-tool-use-with-reinforcement-learning, 2025.
\newblock Accessed: 2025-04-17.

\bibitem[Wang et~al.(2025{\natexlab{b}})Wang, Yu, Gao, Zheng, Liu, Lu, Dang, Chen, Yang, Zhang, et~al.]{wang2025beyond}
Shenzhi Wang, Le~Yu, Chang Gao, Chujie Zheng, Shixuan Liu, Rui Lu, Kai Dang, Xionghui Chen, Jianxin Yang, Zhenru Zhang, et~al.
\newblock Beyond the 80/20 rule: High-entropy minority tokens drive effective reinforcement learning for llm reasoning.
\newblock \emph{arXiv preprint arXiv:2506.01939}, 2025{\natexlab{b}}.

\bibitem[Huber(2011)]{huber2011robust}
Peter~J Huber.
\newblock Robust statistics.
\newblock In \emph{International encyclopedia of statistical science}, pages 1248--1251. Springer, 2011.

\bibitem[Law(1986)]{law1986robust}
John Law.
\newblock Robust statistics—the approach based on influence functions, 1986.

\bibitem[Degris et~al.(2012)Degris, White, and Sutton]{degris2012off}
Thomas Degris, Martha White, and Richard~S Sutton.
\newblock Off-policy actor-critic.
\newblock \emph{arXiv preprint arXiv:1205.4839}, 2012.

\bibitem[Achiam et~al.(2023)Achiam, Adler, Agarwal, Ahmad, Akkaya, Aleman, Almeida, Altenschmidt, Altman, Anadkat, et~al.]{achiam2023gpt}
Josh Achiam, Steven Adler, Sandhini Agarwal, Lama Ahmad, Ilge Akkaya, Florencia~Leoni Aleman, Diogo Almeida, Janko Altenschmidt, Sam Altman, Shyamal Anadkat, et~al.
\newblock Gpt-4 technical report.
\newblock \emph{arXiv preprint arXiv:2303.08774}, 2023.

\bibitem[Team et~al.(2023)Team, Anil, Borgeaud, Alayrac, Yu, Soricut, Schalkwyk, Dai, Hauth, Millican, et~al.]{team2023gemini}
Gemini Team, Rohan Anil, Sebastian Borgeaud, Jean-Baptiste Alayrac, Jiahui Yu, Radu Soricut, Johan Schalkwyk, Andrew~M Dai, Anja Hauth, Katie Millican, et~al.
\newblock Gemini: a family of highly capable multimodal models.
\newblock \emph{arXiv preprint arXiv:2312.11805}, 2023.

\bibitem[Shinn et~al.(2023)Shinn, Cassano, Gopinath, Narasimhan, and Yao]{shinn2023reflexion}
Noah Shinn, Federico Cassano, Ashwin Gopinath, Karthik Narasimhan, and Shunyu Yao.
\newblock Reflexion: Language agents with verbal reinforcement learning.
\newblock \emph{Advances in Neural Information Processing Systems}, 36:\penalty0 8634--8652, 2023.

\bibitem[Kool et~al.(2019)Kool, van Hoof, and Welling]{kool2019buy}
Wouter Kool, Herke van Hoof, and Max Welling.
\newblock Buy 4 reinforce samples, get a baseline for free!
\newblock 2019.

\bibitem[Ahmadian et~al.(2024)Ahmadian, Cremer, Gall{\'e}, Fadaee, Kreutzer, Pietquin, {\"U}st{\"u}n, and Hooker]{ahmadian2024back}
Arash Ahmadian, Chris Cremer, Matthias Gall{\'e}, Marzieh Fadaee, Julia Kreutzer, Olivier Pietquin, Ahmet {\"U}st{\"u}n, and Sara Hooker.
\newblock Back to basics: Revisiting reinforce style optimization for learning from human feedback in llms.
\newblock \emph{arXiv preprint arXiv:2402.14740}, 2024.

\bibitem[Yang et~al.(2024)Yang, Zhang, Hui, Gao, Yu, Li, Liu, Tu, Zhou, Lin, et~al.]{yang2024qwen2}
An~Yang, Beichen Zhang, Binyuan Hui, Bofei Gao, Bowen Yu, Chengpeng Li, Dayiheng Liu, Jianhong Tu, Jingren Zhou, Junyang Lin, et~al.
\newblock Qwen2. 5-math technical report: Toward mathematical expert model via self-improvement.
\newblock \emph{arXiv preprint arXiv:2409.12122}, 2024.

\bibitem[Team(2025{\natexlab{a}})]{team2025sky}
NovaSky Team.
\newblock Sky-t1: Train your own o1 preview model within \$450, 2025{\natexlab{a}}.

\bibitem[Team(2025{\natexlab{b}})]{team2025qwq}
Qwen Team.
\newblock Qwq-32b: Embracing the power of reinforcement learning, 2025{\natexlab{b}}.

\bibitem[Muennighoff et~al.(2025)Muennighoff, Yang, Shi, Li, Fei-Fei, Hajishirzi, Zettlemoyer, Liang, Cand{\`e}s, and Hashimoto]{muennighoff2025s1}
Niklas Muennighoff, Zitong Yang, Weijia Shi, Xiang~Lisa Li, Li~Fei-Fei, Hannaneh Hajishirzi, Luke Zettlemoyer, Percy Liang, Emmanuel Cand{\`e}s, and Tatsunori Hashimoto.
\newblock s1: Simple test-time scaling.
\newblock \emph{arXiv preprint arXiv:2501.19393}, 2025.

\bibitem[Li et~al.(2025{\natexlab{d}})Li, Lin, Jiang, Cao, Liu, Zhang, Huang, Chen, Sun, Wang, et~al.]{li2025chain}
Weizhen Li, Jianbo Lin, Zhuosong Jiang, Jingyi Cao, Xinpeng Liu, Jiayu Zhang, Zhenqiang Huang, Qianben Chen, Weichen Sun, Qiexiang Wang, et~al.
\newblock Chain-of-agents: End-to-end agent foundation models via multi-agent distillation and agentic rl.
\newblock \emph{arXiv preprint arXiv:2508.13167}, 2025{\natexlab{d}}.

\bibitem[Kwon et~al.(2023)Kwon, Li, Zhuang, Sheng, Zheng, Yu, Gonzalez, Zhang, and Stoica]{kwon2023efficient}
Woosuk Kwon, Zhuohan Li, Siyuan Zhuang, Ying Sheng, Lianmin Zheng, Cody~Hao Yu, Joseph~E. Gonzalez, Hao Zhang, and Ion Stoica.
\newblock Efficient memory management for large language model serving with pagedattention.
\newblock In \emph{Proceedings of the ACM SIGOPS 29th Symposium on Operating Systems Principles}, 2023.

\bibitem[Sun et~al.(2024)Sun, Chen, Huang, Xie, Zhu, Zhang, Li, Yang, Han, Shu, et~al.]{sun2024hunyuan}
Xingwu Sun, Yanfeng Chen, Yiqing Huang, Ruobing Xie, Jiaqi Zhu, Kai Zhang, Shuaipeng Li, Zhen Yang, Jonny Han, Xiaobo Shu, et~al.
\newblock Hunyuan-large: An open-source moe model with 52 billion activated parameters by tencent.
\newblock \emph{arXiv preprint arXiv:2411.02265}, 2024.

\bibitem[Havrilla et~al.(2024)Havrilla, Du, Raparthy, Nalmpantis, Dwivedi-Yu, Zhuravinskyi, Hambro, Sukhbaatar, and Raileanu]{havrilla2024teaching}
Alex Havrilla, Yuqing Du, Sharath~Chandra Raparthy, Christoforos Nalmpantis, Jane Dwivedi-Yu, Maksym Zhuravinskyi, Eric Hambro, Sainbayar Sukhbaatar, and Roberta Raileanu.
\newblock Teaching large language models to reason with reinforcement learning.
\newblock \emph{arXiv preprint arXiv:2403.04642}, 2024.

\bibitem[Snoek et~al.(2012)Snoek, Larochelle, and Adams]{snoek2012practical}
Jasper Snoek, Hugo Larochelle, and Ryan~P Adams.
\newblock Practical bayesian optimization of machine learning algorithms.
\newblock \emph{Advances in neural information processing systems}, 25, 2012.

\bibitem[Wang et~al.(2018)Wang, Zariphopoulou, and Zhou]{wang2018exploration}
Haoran Wang, Thaleia Zariphopoulou, and Xunyu Zhou.
\newblock Exploration versus exploitation in reinforcement learning: A stochastic control approach.
\newblock \emph{arXiv preprint arXiv:1812.01552}, 2018.

\bibitem[Agrawal and Goyal(2012)]{agrawal2012analysis}
Shipra Agrawal and Navin Goyal.
\newblock Analysis of thompson sampling for the multi-armed bandit problem.
\newblock In \emph{Conference on learning theory}, pages 39--1. JMLR Workshop and Conference Proceedings, 2012.

\bibitem[Ladosz et~al.(2022)Ladosz, Weng, Kim, and Oh]{ladosz2022exploration}
Pawel Ladosz, Lilian Weng, Minwoo Kim, and Hyondong Oh.
\newblock Exploration in deep reinforcement learning: A survey.
\newblock \emph{Information Fusion}, 85:\penalty0 1--22, 2022.

\bibitem[Deng et~al.(2025)Deng, Ren, Li, Sutherland, Li, and Thrampoulidis]{deng2025effect}
Wenlong Deng, Yi~Ren, Muchen Li, Danica~J Sutherland, Xiaoxiao Li, and Christos Thrampoulidis.
\newblock On the effect of negative gradient in group relative deep reinforcement optimization.
\newblock \emph{arXiv preprint arXiv:2505.18830}, 2025.

\bibitem[Zeng et~al.(2025)Zeng, Wei, Brown, Frunza, Nevmyvaka, and Hong]{zeng2025reinforcing}
Siliang Zeng, Quan Wei, William Brown, Oana Frunza, Yuriy Nevmyvaka, and Mingyi Hong.
\newblock Reinforcing multi-turn reasoning in llm agents via turn-level credit assignment.
\newblock \emph{arXiv preprint arXiv:2505.11821}, 2025.

\bibitem[Cui et~al.(2025{\natexlab{b}})Cui, Yuan, Wang, Wang, Zhang, Chen, Li, He, Fan, Yu, et~al.]{cui2025process}
Ganqu Cui, Lifan Yuan, Zefan Wang, Hanbin Wang, Yuchen Zhang, Jiacheng Chen, Wendi Li, Bingxiang He, Yuchen Fan, Tianyu Yu, et~al.
\newblock Process reinforcement through implicit rewards.
\newblock \emph{arXiv preprint arXiv:2502.01456}, 2025{\natexlab{b}}.

\bibitem[Liu et~al.(2025{\natexlab{c}})Liu, Wang, Wu, Huang, Li, Zhang, and Jiao]{liu2025agentic}
Xiaoqian Liu, Ke~Wang, Yuchuan Wu, Fei Huang, Yongbin Li, Junge Zhang, and Jianbin Jiao.
\newblock Agentic reinforcement learning with implicit step rewards.
\newblock \emph{arXiv preprint arXiv:2509.19199}, 2025{\natexlab{c}}.

\bibitem[Moerland et~al.(2023)Moerland, Broekens, Plaat, Jonker, et~al.]{moerland2023model}
Thomas~M Moerland, Joost Broekens, Aske Plaat, Catholijn~M Jonker, et~al.
\newblock Model-based reinforcement learning: A survey.
\newblock \emph{Foundations and Trends{\textregistered} in Machine Learning}, 16\penalty0 (1):\penalty0 1--118, 2023.

\bibitem[Gu et~al.(2024)Gu, Zhang, Ning, Zheng, Gou, Xue, Chang, Srivastava, Xie, Qi, et~al.]{gu2024your}
Yu~Gu, Kai Zhang, Yuting Ning, Boyuan Zheng, Boyu Gou, Tianci Xue, Cheng Chang, Sanjari Srivastava, Yanan Xie, Peng Qi, et~al.
\newblock Is your llm secretly a world model of the internet? model-based planning for web agents.
\newblock \emph{arXiv preprint arXiv:2411.06559}, 2024.

\end{thebibliography}
\end{document}